%% file: InfoLandscape.tex
\documentclass[utf8]{frontiersSCNS} 

\usepackage{url,hyperref,lineno,microtype,subcaption}
\usepackage[onehalfspacing]{setspace}

\usepackage[american]{babel}
\usepackage[autostyle=true]{csquotes}
\usepackage[american]{isodate}
\usepackage{todonotes}
\usepackage{longtable}

\usepackage{graphicx}
\usepackage{caption}
\usepackage{subcaption}

\def\keyFont{\fontsize{8}{11}\helveticabold }
\def\firstAuthorLast{Mehler {et~al.}} 
\def\Authors{Alexander Mehler\,$^{1*}$, Wahed Hemati\,$^{1}$, Pascal Welke\,$^{2}$, Maxim Konca\,$^{1}$ and Tolga Uslu\,$^{1}$}
\def\Address{$^{1}$Text-Technology Lab, Institute of Computer Science, Faculty of Computer Science and Mathematics, Goethe University Frankfurt, Frankfurt a.\,M., Germany \\
  $^{2}$ Machine Learning and Artificial Intelligence Lab, Institute for Computer Science, University of Bonn, Bonn, Germany}
\def\corrAuthor{Corresponding Author}

\def\corrEmail{mehler@uni.frankfurt.de}

\usepackage[autolanguage]{numprint}
\usepackage{comment}
\usepackage{tikz} 
\usetikzlibrary{matrix}
\usetikzlibrary{fit}
\usetikzlibrary{external}
\DeclareMathOperator{\ged}{\mathit{ged}}
\definecolor{SeminarBlau}{RGB}{0,154,224}
\definecolor{SeminarRot}{rgb}{0.75,0,0}
\definecolor{SeminarGruen}{RGB}{0,128,0}
\definecolor{SeminarOrange}{RGB}{237,167,45} 
\definecolor{SeminarGrau}{rgb}{0.32,0.3,0.38}
\colorlet{SeminarHellBlau}{SeminarBlau!25!white}
\colorlet{SeminarHellGruen}{SeminarGruen!25!white}
\colorlet{SeminarHellRot}{SeminarRot!25!white}
\colorlet{SeminarHellGrau}{SeminarGrau!35!white}
\colorlet{SeminarSehrHellRot}{SeminarRot!15!white}
\colorlet{SeminarSehrHellGrau}{SeminarGrau!15!white}
\colorlet{SeminarSehrHellGruen}{SeminarGruen!15!white}
\colorlet{SeminarSehrHellBlau}{SeminarBlau!15!white}
\colorlet{SeminarSehrSehrHellGrau}{SeminarGrau!5!white}

\newcommand{\abs}[1]{\left| #1 \right|}
\newcommand{\rrid}[1]{(#1)}

\begin{document}
\onecolumn
\firstpage{1}

\title[Multiple Texts as a Limiting Factor in Online Learning]{Multiple Texts as a Limiting Factor in Online Learning: Quantifying (Dis-)similarities of Knowledge Networks across Languages} 

\author[\firstAuthorLast ]{\Authors} 
\address{} 
\correspondance{} 

\extraAuth{}

\maketitle

\begin{abstract}
\noindent We test the hypothesis that the extent to which one obtains information on a given topic through Wikipedia depends on the language in which it is consulted. 
Controlling the size factor, we investigate this hypothesis for a number of 25 subject areas. 
%
%
Since Wikipedia is a central part of the web-based information landscape, this indicates a language-related,  linguistic bias. 
The article therefore deals with the question of whether Wikipedia exhibits this kind of linguistic relativity or not.
From the perspective of educational science, the article develops a computational model of the information landscape from which multiple texts are drawn as typical input of web-based reading.
For this purpose, it develops a hybrid model of intra- and intertextual similarity of different parts of the information landscape and tests this model on the example of 35 languages and corresponding Wikipedias.
In this way the article builds a bridge between reading research, educational science, Wikipedia research and computational linguistics.



\tiny
 \keyFont{ \section{Keywords:} Multiple Texts, Information Landscape, Knowledge Graphs, Intratextual Similarity, Intertextual Similarity, Three-level Topic Model, Network Similarity Measurement, Linguistic Relativity} 
\end{abstract}

\section{Introduction}\label{sec:Introduction}


Reading is increasingly carried out by means of online multiple texts, which can simultaneously consist of (segments of) texts of diverse genres, registers, authorships, credibilities etc.\ \citep{Barzilai:Zohar:2012,Britt:Rouet:Durik:2018,Goldman:Braasch:Wiley:Graesser:Brodowinska:2012}. 
That is, learning takes place, so to speak, on the basis of \enquote{document collages} whose components are gathered from a constantly growing, nowadays mostly web-based information landscape \citep{Zlatkin-Troitschanskaia:Schmidt:Molerov:Shavelson:Berliner:2017} or space \citep{Hartman:Hagerman:Leu:2018}.\footnote{The term information landscape \citep{Zlatkin-Troitschanskaia:Bruckner:Molerov:Bisang:2019,Zlatkin-Troitschanskaia:et:al:2019} includes but is not limited to the web as both a huge and highly diverse (multimodal) text data repository or source of online reading \citep{Cho:Afflerbach:2015,Wolf:2018}, which is substructured along countless web genres \citep{Mehler:Sharoff:Santini:2009:a}, registers and thematic domains.
Online reading resembles a traversal of this landscape, each of which involves numerous decisions about what to read, in what sequence and in what depth. 
The term `landscape' manifests this dual character: 
on the one hand, the information landscape is a repository that offers innumerable decision possibilities (intertext-as-product perspective), which on the other hand are to be decided by the reader \citep{Cho:Afflerbach:2015,Britt:Rouet:Durik:2018} in such a way that for each reading process a (usually different) multiple text is delimited in this landscape (intertext-as-process perspective).}
The multiplicity of the texts involved and the diversity of their genres and registers \citep{Halliday:Hasan:1989} are text-linguistic characteristics of online reading  \citep{Britt:Rouet:Durik:2018}.
A third, so to speak macroscopic aspect of this process is the starting point of this article. 
It is about the \textit{Information Landscape} (IL) from which innumerable readers in countless reading processes delineate ever new multiple texts and thus manifest a distributed process through which this landscape is opened up.
To introduce our research agenda regarding this IL, we start from the \emph{Documents Model} (DM) of \citet{Perfetti:Rouet:Britt:1999} and \citet{Britt:Rouet:Braasch:2012}.
While multiple texts are studied by a wide range of approaches\footnote{See \citet{Primor:Katzir:2018} for a current overview and \citet{List:Alexander:2019} for an integrated view of this model landscape.}, the reason for choosing the DM as a starting point is due to its text-linguistic heritage -- based on the Construction-Integration Model (CIM) of \citet{Kintsch:1998} -- and its context model, which facilitates modular extensions.
As far as the text-linguistic orientation of the DM is concerned, its notion of the so-called intertext model is of particular interest for our study of the IL.

Generally speaking, the DM distinguishes two outcomes of multiple text comprehension:
the \textit{Intertext Model} (IM), which comprises representations of the constituents of multiple texts and their links, and the \textit{Mental Model} (MM)\footnote{Instead of situation models, \citet{Britt:Rouet:Durik:2018} speak of integrated models as mental models of the semantic content of  multiple texts, to emphasize that this content is not limited to descriptions of situations (as provided, e.g., by narrative texts). 
In this way, depending on the task, texts of different types get into the focus of their reading model. 
We prefer to speak of mental models to emphasize the openness of our approach to the DM.} as a result of comprehension processes that operate within and beyond the boundaries of these constituents.
This includes the process of integration, a term borrowed from the CIM, which in the DM also concerns information from different texts.
In contrast to text linguistics, which predicts that cohesion and coherence relations should be resolvable within the boundaries of a text to facilitate its understanding \citep{Kintsch:1998}, this condition does not usually apply to multiple texts:
they induce additional intrinsic cognitive loads \citep{Sweller:1994} as a result of interacting elements of separate texts (such as conflicting, contradicting or otherwise incoherent statements about the same event; \citealt{Barzilai:Zohar:2012}) and increased efforts in decision making as a result of hyperlinkage \citep{DeStefano:LeFevre:2007}.\footnote{
\textcolor{black}{For a review of aspects of cognitive load of Internet-based online reading see \citet{Loh:Kanai:2016}.}}
\textcolor{black}{Consequently, \citet[p.357]{Goldman:Braasch:Wiley:Graesser:Brodowinska:2012} speak of (as we may add: online) reading as an \textit{intertextual process}.}

\begin{figure}[t]
	\centering
	\includegraphics[width=0.85\linewidth]{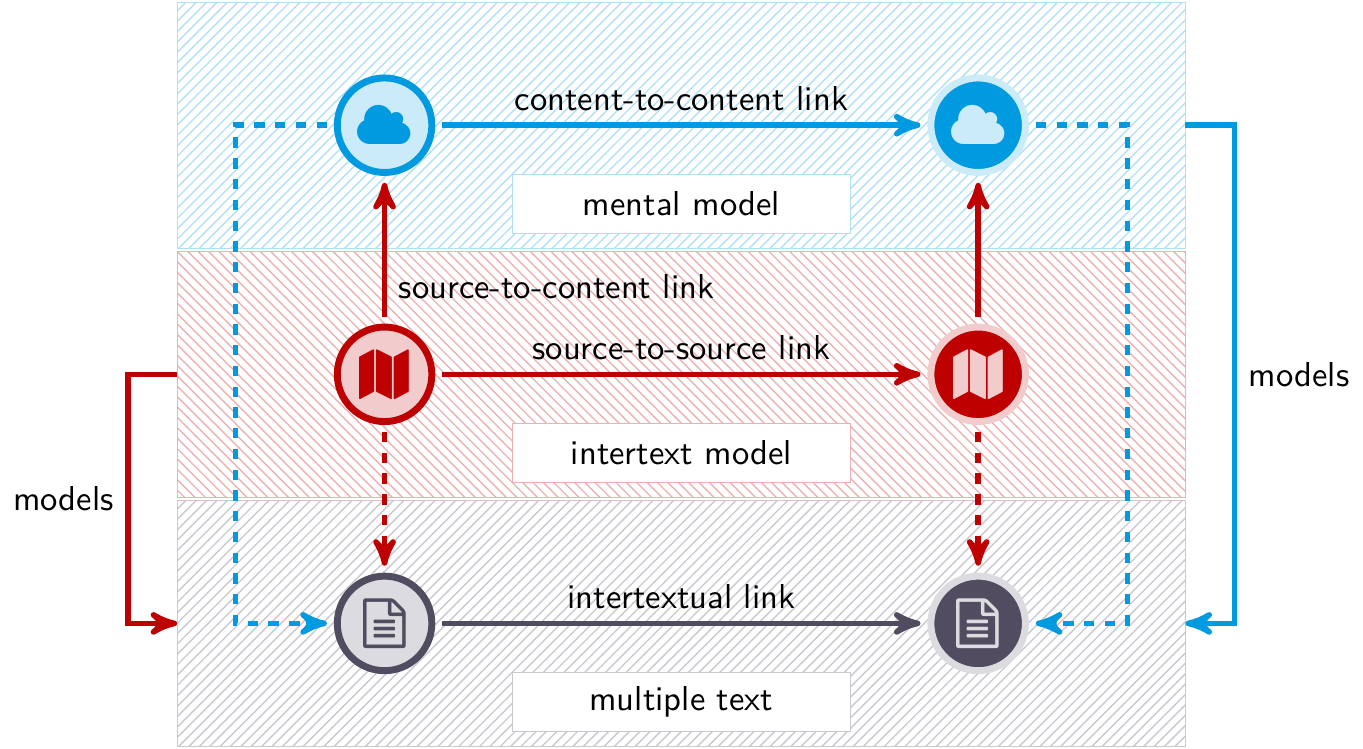}
	\caption[IM and MM]{Intertext model and mental model in relation to multiple texts: units and relations.}
	\label{fig:Diagram}
\end{figure}

\citet{Britt:Rouet:Braasch:2012} assume that intertext models represent selected constituents of multiple texts as \enquote{document entities} together with entity-related information (e.g.\ on authorship). 
This is supplemented by three types of links:
IM-related source-to-source (e.g.\ $x$ \textit{supports} or \textit{contradicts} $y$), MM-related content-to-content and source-to-content links (see Figure \ref{fig:Diagram}).
A prediction of the DM, which is crucial for our work, is that the probability of generating an intertext model as a result of reading a multiple text is a function of the number of the texts involved, their authors, the perspectives they provide on the described situation \citep[p.\,171]{Britt:Rouet:Braasch:2012}, the tasks to be accomplished and other contextual factors \citep{Britt:Rouet:Durik:2018}.
This suggests to speak of the intertext model as a kind of 
cognitive map \citep{Downs:Stea:1977} of the underlying multiple text, where the MM abstracts from this textbase (e.g.\ by applying macro operations; \citealp{Dijk:1980:a,Dijk:Kintsch:1983}):
that is, readers produce intertext models as cognitive maps of multiple texts as parts of the underlying IL, while groups or communities of readers produce distributed cognitive maps \citep{Mehler:Gleim:Gaitsch:Uslu:Hemati:2019} of larger sections of the IL or the IL as a whole.
This duality of small- and large-scale reading processes leads to the object of this article. 
That is, we can ask how the IL looks like from the perspective of these distributed cognitive maps or vice versa, how it presents itself to different reader communities.
The latter question will be in the focus of this article.

Although the DM takes the necessary step of generalizing the CIM towards modeling multiple texts, it is largely single reader-oriented.
To broaden this focus, we generalize the DM conceptually in two steps:

\begin{figure}[t]
	\centering
	\includegraphics[width=1.00\linewidth]{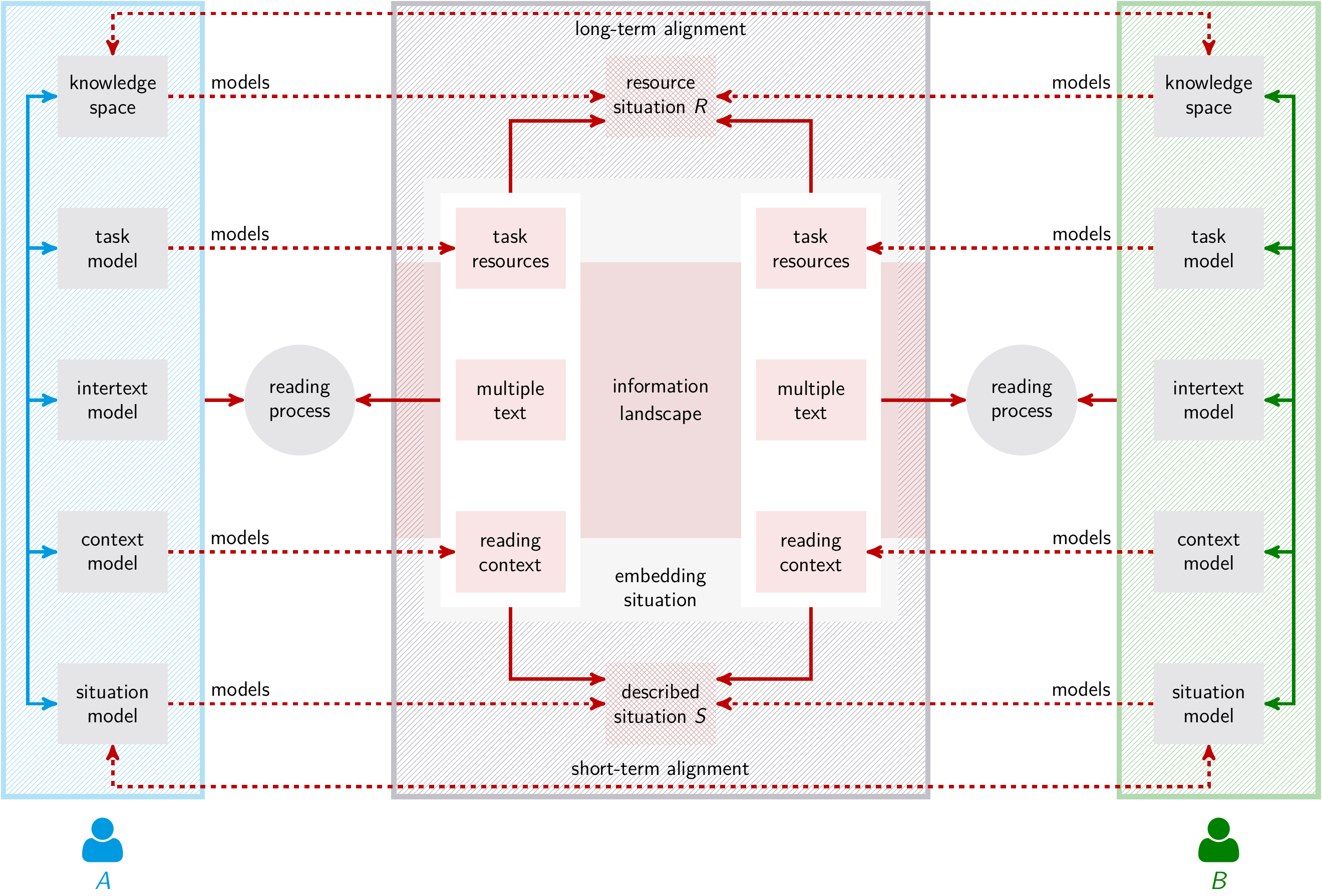}
	\caption[Documents model extended]{Extending the DM in terms of alignment research.}
	\label{fig:documentsmodelextended}
\end{figure}

\begin{enumerate}
\item \textit{Microscopic alignment:} 
In a first step, which is still within the boundaries of the DM, we consider intertext models as reader-centered approximations of parts of the IL with varying degrees of explicitness (reflecting the probability mentioned above).
This predicts that different readers can approximate different parts (i.e.\ multiple texts) of the IL just as they can align \citep{Pickering:Garrod:2004} their intertext models of overlapping parts depending on their interaction\textcolor{black}{, which according to \citet[p.\,286]{Salmeron:Kammerer:Delgado:2018} is a characteristic of non-academic online reading} and also regards online collaborative learning \citep[p.\,487]{Coiro:Sparks:Kulikowich:2018}.
Starting from the context model of \citet[p.\,53]{Britt:Rouet:Durik:2018}, 
Figure \ref{fig:documentsmodelextended} illustrates this alignment scenario in terms of a situation semantic adaptation \citep{Mehler:Ramesh:2019}:
two readers $A$ and $B$ read not necessarily different multiple texts in related contexts (related to space, time, etc.) to solve the same or related tasks (manifested by task instructions etc.) in order to achieve the same or related goals.
We assume that the texts originate from an IL in whose context they describe a situation $S$ and that $A$ and $B$ refer to the same resource situation $R$ to understand which situation their multiple texts actually describe, where all references to contextual units are indirect: 
they are mediated through mental representations \citep[cf.][p.\,45]{Britt:Rouet:Durik:2018} of multiple texts (IM), described situations (MM), reading contexts (context model), task contexts (task model) and resource situations (long-term memory (LTM) or knowledge space).
As a result of collaborative, cooperative or simply parallel reading processes, the representations of the readers may align with each other, in the short-term (concerning, e.g., their MMs) or long-term (regarding their LTMs).
That is, as illustrated in Figure \ref{fig:documentsmodelextended}, $A$ and $B$ have the possibility to align their mental representations so that they understand the same or similar multiple texts as descriptions of the same or similar situations.\footnote{Adding the notion of a described situation to the context model of \citet{Britt:Rouet:Durik:2018} allows for distinguishing communication scenarios such as misunderstandings ($A$ and $B$ represent the same situation $S$ in mutually incompatible ways), disinformation ($A$ pretends to describe a situation he or she knows to be unreal, whereas $B$ considers it to be factual) or misinformation ($A$ and $B$ represent the same unreal situation $S$, which both consider real) \citep{Kendeou:Robinson:McCrudden:2019} and related phenomena.}
Evidently, such an alignment requires many things, but at least the chance that both readers have access to the same or semantically sufficiently similar texts from which they can extract the same or sufficiently similar multiple texts. 
\textit{But do they?}
This question brings us to the second step of generalizing the DM:
\item \textit{Macroscopic alignment:} 
From the perspective of reader communities, reading is a distributed process that approximates a multifaceted IL, so that both the IL and its distributed representation by innumerable intertext models jointly develop.
Obviously, parallel to the diversity of the IL, communities of readers are also diverse as a result of a wide range of factors \citep{Braasch:McCrudden:Braten:2018,Hsieh:2012,Hargittai:Dobransky:2017}:
for example, membership in different language communities (as focused by this article), ethnicities, cultures, age groups, social groups (e.g.\ families), residency in different places (or geographies; \citealt{Graham:Straumann:Hogan:2015}) or practice of different social roles.
In any event, in analogy to Step 1, we may expect that different communities dealing with the same or semantically similar parts of the IL should be able to align their corresponding intertext models among each other.
In shorter terms: 
different groups should be able to represent similar parts of the IL in a similar way.
\textit{But do they actually have access to the same or at least sufficiently similar parts of the IL -- especially under the condition that they deal with the same topic?}

\begin{figure}[t]
	\centering
	\includegraphics[width=1.00\linewidth]{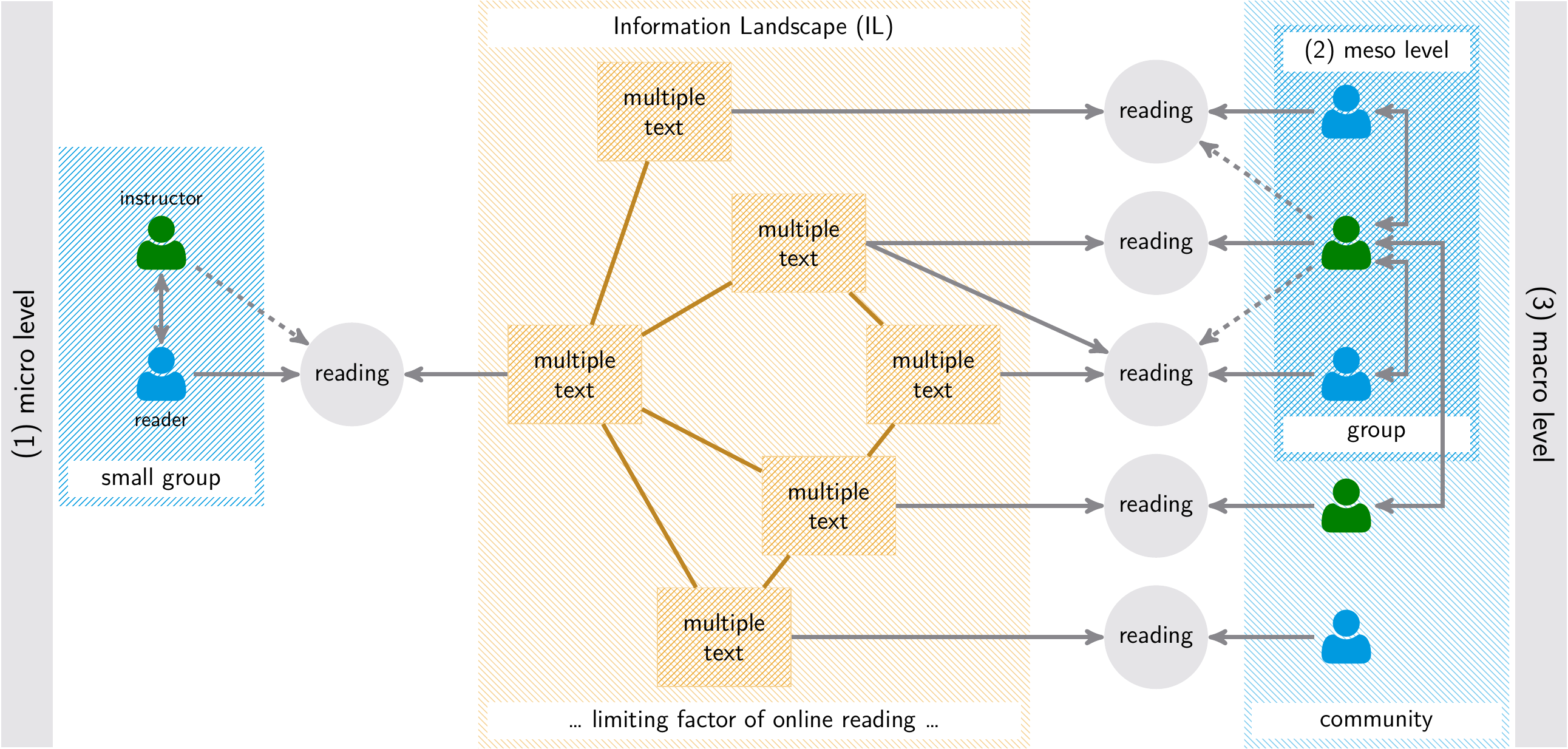}
	\caption[Research perspectives]{Three research perspectives on multiple texts.}
	\label{fig:Research perspectives}
\end{figure}

\item Step 1 concerns the microscopic alignment of intertext models as a result of reading situations that are paired together via shared, overlapping or otherwise related multiple texts, tasks, etc.
Step 2 refers to the macroscopic alignment of reader communities as a result of countless such pairings, whereby these communities and consequently their alignments are subdivided according to their social structure.
In the range of these extremes there are meso-level alignment processes manifested by smaller groups of agents (such as learning groups) (see Figure \ref{fig:Research perspectives}). 
The three levels have in common that the underlying IL is temporally on a different scale:
it is subject to a lower dynamic than the multiple texts that are extracted from it as a result of reading processes 
-- even in the case of algorithmically (generated and) linked documents (where the underlying algorithms may reflect user profiles).
But how uniformly does the IL present itself to its (communities or groups of) readers? 
Obviously this question is currently outside the scope of the framework of the DM and its relatives.
\end{enumerate}

Step 2 concerns precisely the viewpoint of this article.
That is, we are concerned with a central prerequisite for alignable intertext models among readers as members of large communities. 
This refers to the intertextual shape of the IL from the perspective of different communities who may have different accesses to it or \enquote{see} different landscapes, even in situations where the opposite would be assumed.
The DM and related approaches do not model what the multiple texts are extracted from and what countless intertext models in their distributed totality ultimately represent, that is, an underlying multifaceted, highly dynamic IL, its numerous document nodes and their relational, intertextual embeddings.

According to \citet [p.\,56]{Hartman:Hagerman:Leu:2018}, reading research mostly considers small amounts of offline texts pre-selected by the experimenter rather than open ILs in which users decide what to read.
But if reading is a kind of problem solving that involves multiple search and decision processes \citep[p.\,43]{Britt:Rouet:Durik:2018} (e.g.\ about what to search for and where to find it), then the question arises as to the limits of these processes as imposed by the IL and how they differ for which reader communities.
Apparently, approaches to multiple texts focus on micro-models that leave the corresponding macro-models, which inform about the shape of the IL and its organizational laws, under-specified.
The present paper takes a step in the direction of filling this gap:
it develops a macroscopic model of the IL and examines how its shape appears from the perspective of certain large-scale reader communities.
Our aim is, so to speak, to impart knowledge about the \enquote{wild} in which the sort of reading takes place which according to \citet[p.\,535]{Braasch:McCrudden:Braten:2018} is to become the subject of reading research.
Thus, our approach is complementary to current research on the intertext model:
we study the IL underlying the construction of intertext models from a macroscopic perspective, in contrast to reading research, which starts from a microscopic perspective of small groups or individual readers (see Figure \ref{fig:Research perspectives}).
\textcolor{black}{In terms of the integrated framework of multiple texts \citep{List:Alexander:2019} we are concerned with the intertextuality of those information units to which the cognitive strategies and behavioral skills of  readers are related.}
That is, in modification of the fourth goal of future research on the use of multiple sources according to \citet{Braasch:McCrudden:Braten:2018}, we deal with the phenomenon that different communities are offered different information, especially in the context of the same topic.
The extent to which this phenomenon applies to different language communities will be examined using the example of the most frequently used knowledge resource on the Web, that is Wikipedia \rrid{RRID:SCR\_004897}.


The article is organized as follows:
Section \ref{sec:Wikipedia} explains the relevance of Wikipedia for educational science and gives an overview of related research.
Section \ref{sec:Rationale and Method} explains our research questions and describes in detail the methods we have developed to answer them. 
In Section \ref{sec:Experiment}, we describe our experiments and discuss their results.
Finally, in Section \ref{sec:Conclusion} we give a conclusion and an outlook on future work.

\section{Wikipedia: Educational Relevance and Bias}\label{sec:Wikipedia}

\begin{figure}[t]
	\centering
	\includegraphics[width=1.00\linewidth]{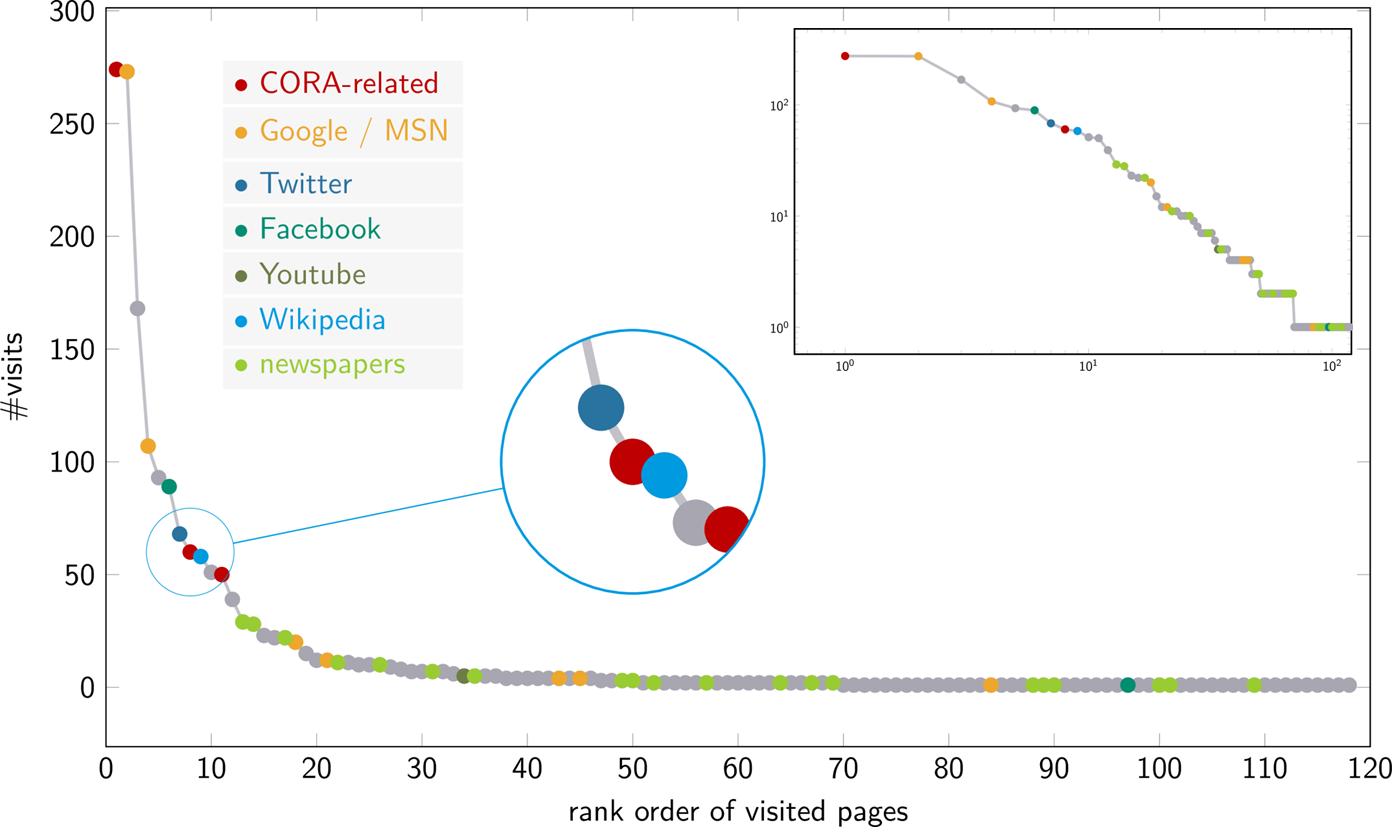}
	\caption[CORA]{Rank-frequency distribution of accesses to web resources in the CORA project.}
	\label{fig:CORA}
\end{figure}

Wikipedia is a primary multilingual source for web-based knowledge acquisition and online learning \citep{Head:2013,Konieczny:2016,Lemmerich:et:al:2019,Lucassen:Schraagen:2010,Mesgari:Okoli:Mehdi:Nielsen:Lanamaki:2015,Okoli:et:al:2012,Okoli:Mehdi:Mesgari:Nielsen:Lanamaki:2014,Singer:et:al:2017}.
It not only offers releases for numerous languages, but is changing and growing to a degree that makes it a first reference for an open topic universe, a resource which -- besides the web as a whole -- reflects the breadth, depth and dynamics of human knowledge generation to an outstanding degree: 
whenever a new topic comes up, little time passes before Wikipedia provides information about how it is classified and related to already established topics.\footnote{\textcolor{black}{Given its relevance, it is astonishing how rarely Wikipedia is mentioned as a source for reading in 
\citet{Braasch:Braaten:McCrudden:2018}.}}
Its dynamics and multilingualism even make it a source for the automatic extraction of 
learning content \citep{Pentzold:et:al:2017,Conde:et:al:2020}.
The educational relevance refers to Wikipedia as a whole as well as to specific domains such as law or health \citep{Okoli:Mehdi:Mesgari:Nielsen:Lanamaki:2014,Smith:2020}.
\citet{Singer:et:al:2017} 
show that 16\% of English Wikipedia uses are work/school-related tending to involve long page views in thematically coherent sessions.
%
\citet{Lemmerich:et:al:2019} additionally show that these uses vary considerably across languages, especially in the case of work/school-related uses. 
%
Independent of academic reservations \citep[cf.][]{Konieczny:2016,Mesgari:Okoli:Mehdi:Nielsen:Lanamaki:2015}, Wikipedia establishes itself as a source for reading by students \citep{Okoli:Mehdi:Mesgari:Nielsen:Lanamaki:2014}, as an additional learning resource \citep{Konieczny:2016}, partly with advantages over textbooks \citep{Scaffidi:et:al:2017}, a resource to which students themselves contribute \citep{Okoli:et:al:2012}.
Figure \ref{fig:CORA} illustrates the relevance of Wikipedia for online academic reading: 
it shows the Zipfian distribution of online sources frequented by students in the CORA study \citep{Molerov:et:al:2020}. 
In this context, Wikipedia appears as an outstanding reference. 
Even more so: taking into account the \enquote{mutual beneficial relationship} of Google and Wikipedia \citep{McMahon:Johnson:Hecht:2017} (according to which traffic on Wikipedia originates mainly from Google), both resources largely dominate this usage scenario.

Given the exceptional position of Wikipedia, the research community has investigated numerous biases regarding its content and use.
\citet{Denning:Horning:Parnas:Weinstein:2005} speak of accuracy, motive, expertise, volatility, coverage and source as risks of Wikipedia, which are potential reference points for biases.
In terms of coverage, this concerns, for example, the \enquote*{self-focus bias} \citep{Hecht:Gergle:2009} or the tendency that topic selections reflect author preferences (cf.\ \citealt{Holloway:Bozicevic:Boerner:2007} and \citealt{Halavais:Lackaff:2008}).
In this context, \citet{Hecht:Gergle:2010:a} 
show for 25 language editions that they differ enormously in the coverage of topics, that this diversity is not explained by their size and that English Wikipedia does not cover its sister editions.
A similar approach is taken by \citet{Warncke-Wang:et:al:2012}, who investigate a variant of Tobler's first law \citep{Tobler:1970} according to which geographically nearby Wikipedias are more similar in terms of \textit{Inter-Language Link}s (ILL).
What distinguishes these studies from ours is that, with few exceptions, conceptual alignments of Wikipedias concern only paired articles identified by ILLs, so that hypertext structure, which is crucial for online learning in terms of the DM, is ignored.
Furthermore, similarities of articles are quantified by degree statistics, so that one can hardly speak of a content-based comparison.
Comparable to the latter studies, however, we also refer to aligned articles to map shared topics (cf.\ \citealt{Bao:et:al:2012}), but do so via Wikidata \rrid{RRID:SCR\_018492} to identify commonly referred concepts \textit{and retain their network structure as manifested by Wikipedia's article graph}.

Biased topic coverage is related to what \citet{Massa:Scrinzi:2012} call the linguistic point of view (contradicting Wikipedia's NPOV), which predicts that different (e.g.\,cultural) communities tend to present the same topics differently.
In line with this view, thematic biases reflect cultural differences.
This perspective is further developed by \citet{Miquel-Ribe:Laniado:2016}, who speak of cultural identities, according to which editors tend to write about topics related to their culture. 
Using geo-referenced data and keyword-related heuristics, they identify cultural identity related articles (on average 25\% of articles in 40 languages) to diagnose language-specific thematic preferences (regarding 15 languages) and translation-related associations of editions that are  dominated by certain topics (e.g.\ \textit{geography}).
Thematic preferences are modeled using 18 topics derived with the help of Wikipedia's category system.
Based on this analysis, they distinguish types of culture-related articles:
language-specific articles shared by a few editors and articles appearing in many languages (in terms of ILLs).
Another example is \citet{Miz:Hanna:Aspert:Ricaud:Vandergheynst:2020}, who examine English, French and Russian Wikipedia by exploring clusters of trending articles using topic modelling based on 8 topics.
In contrast to these studies, we not only consider a larger number of languages and many more topics, but especially semantically coherent article subnetworks, which are examined for their differences along intra- and intertextual dimensions. 

Biases of Wikipedia were also analysed regarding selected areas:
%
%
\citet{Lorini:et:al:2020} observe a variant of Tobler's Law according to which authors tend to write about geographically close events. 
Similarly, \citet{Samoilenko:et:al:2017} describe a preference for recency.
%
%
\citet{Oeberst:vonderBeck:Matschke:Ihme:Cress:2019} investigate a bias of groups who present their views more positively (cf.\ \citealt{Alvarez:et:al:2020}).
A related example regarding biographical articles that includes linguistic analyses is given by \citet{Callahan:Herring:2011}. 
\citet{Wagner:et:al:2016} also present a multidimensional content analysis, now of a gender bias. 
%
Given the importance of Wikipedia as a knowledge repository and taking into account its various biases, the question of its influence on knowledge formation on the part of readers comes up \citep{Oeberst:et:al:2018}. 
%

The research considered so far shares the observation of a biased topic coverage, which relativizes
Wikipedia's domain independence \citep{Jiang:et:al:2017}, since certain topics \citep{Kittur:Chi:Suh:2009} or views dominate, be it due to cultural preferences \citep{Massa:Scrinzi:2012,Laufer:Wagner:Floeck:Strohmaier:2015,Miquel-Ribe:Laniado:2016}, language differences \citep{Hecht:Gergle:2010:a,Massa:Scrinzi:2012,Warncke-Wang:et:al:2012,Samoilenko:et:al:2016}, geographical factors \citep{Hecht:Gergle:2010:b,Karimi:et:al:2015,Laufer:Wagner:Floeck:Strohmaier:2015,Lorini:et:al:2020,Samoilenko:et:al:2016}, or the fact that group membership influences POV
\citep{Oeberst:vonderBeck:Matschke:Ihme:Cress:2019}.
However, though these observations should be based on content-related analyses of large amounts of data, they often concern rather non-content related features (e.g.\ degree statistics) taking into account a maximum of 20 topics, so that topic resolution is kept low while hypertext structure is underrepresented.
On the other hand, concentrating on selected areas allows accurate linguistic analyses to be carried out, but these are difficult to automate and thus difficult to apply across languages.
What is needed, therefore, is a procedure that allows for more precise thematic and article network-related analyses and which can be automatically applied to many languages. 
Exactly such a procedure is presented here: 
it uses Wikidata to identify subjects of articles of whatever subject areas, and about 100 topic categories to model their diversity.
%
%
This will allow us to investigate biased topic coverage 
intra- and intertextually.

\section{Rationale and Method}\label{sec:Rationale and Method}

To investigate the topic coverage of Wikipedia in educationally relevant areas, we investigate how the descriptions of the same entities or knowledge objects (from the fields of economics, physics, chemistry, biology, etc.) are distributed across its language editions.
That is, we investigate how Wikipedia presents itself to its readers as part of the IL in the area of education-related reading.
We test the hypothesis that the extent to which one obtains information on a given topic depends on the language in which Wikipedia is consulted. 
Given the skewed size distribution of Wikipedia's releases for different languages, this may sound obvious at first.
But we will control the size factor and examine this hypothesis for individual subject areas and topics.
%
%
Since Wikipedia is a highly frequented part of the IL, this would indicate a \textit{language-related bias}, that is, a sort of \textit{linguistic relativity}.
Thus, our article is ultimately concerned with the question of whether Wikipedia exhibits this kind of relativity or not. 
We test this hypothesis on the example of 35  Wikipedias.
To this end, we focus on three research questions:

\begin{figure}[t]
	\centering
	\includegraphics[width=0.37\linewidth]{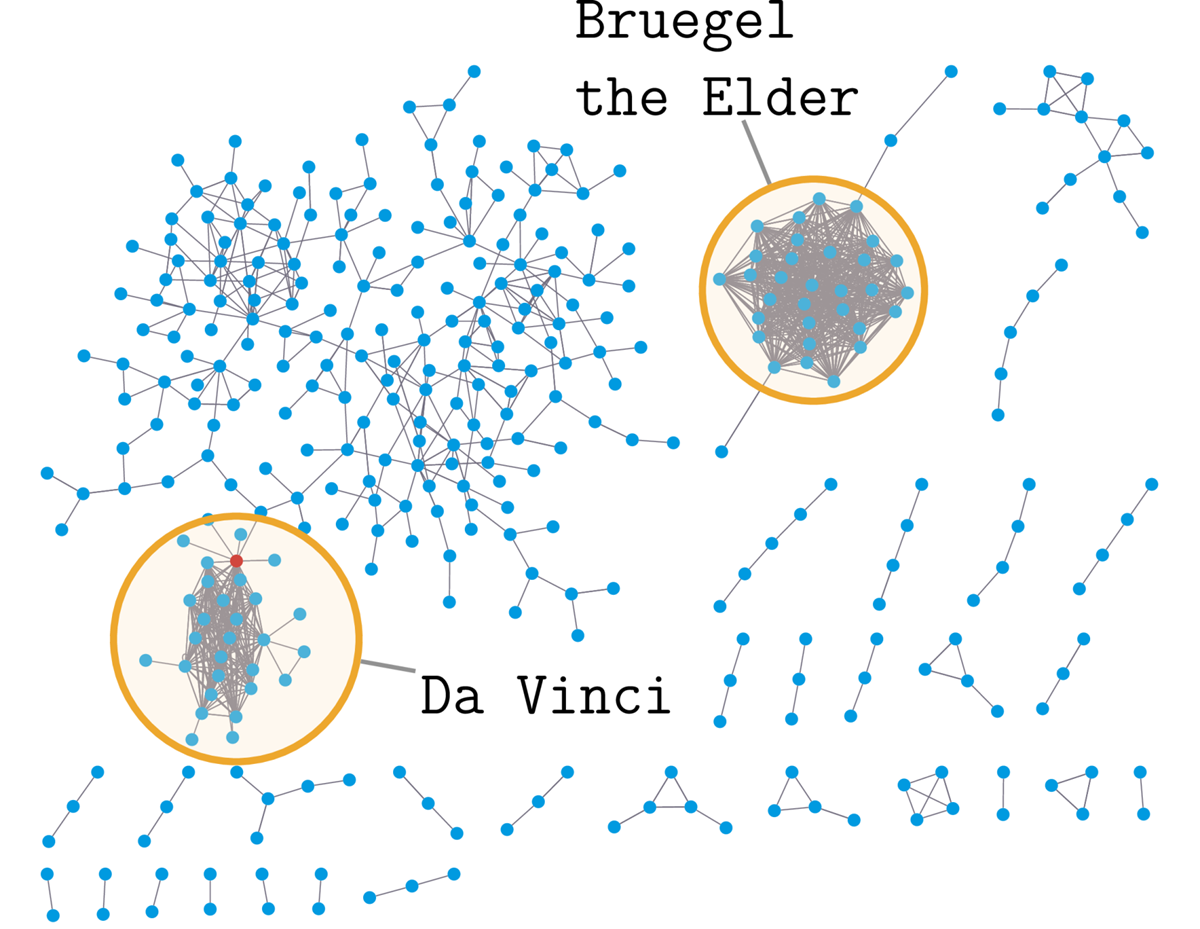}
	\hfill
	\includegraphics[width=0.60\linewidth]{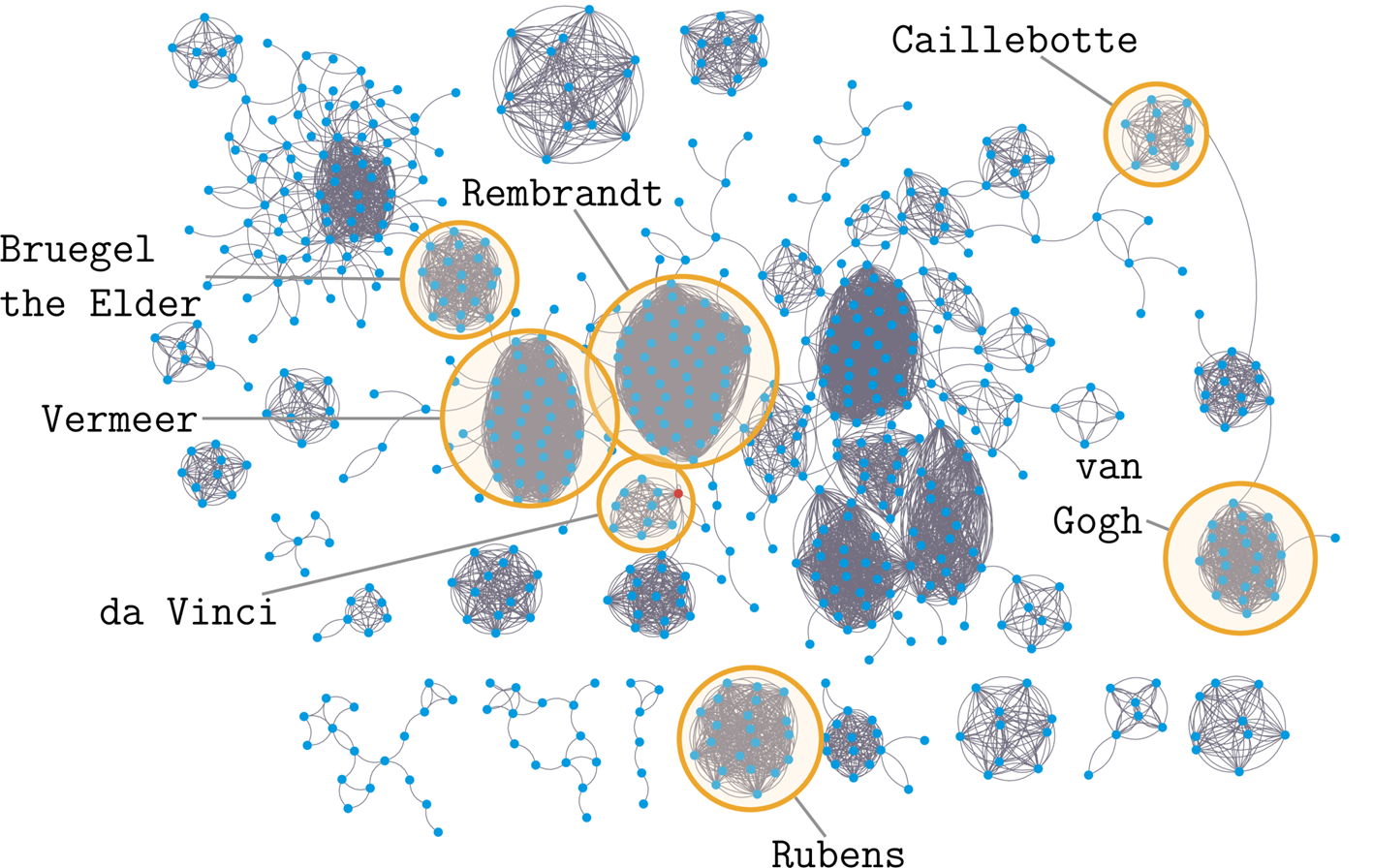}
	\caption{Extracts of two item networks of articles each describing a separate painting: 
	extracted from German (left) and Dutch Wikipedia (right). Download: June 2017.}
	\label{fig:Example}
\end{figure}

\begin{enumerate}
	\item[Q1] How do languages resemble each other in terms of the knowledge networks that manifest themselves in the associated Wikipedias?
\end{enumerate}

Obviously, high dissimilarities in the latter sense mean that students who consult the associated Wikipedias are informed very differently about the same field of knowledge. 
Differences in knowledge between learners of different languages may then be consolidated or even expanded as a result of such a bias.
An example of such a scenario is shown in Figure \ref{fig:Example}, which contrasts networks of articles about paintings from German and Dutch Wikipedia.
The extracted networks are obviously very different; they show very different parts of the information landscape, although on the same subject area.
%
%

\begin{enumerate}
\item[Q2] How do the similarity ratings after Q1 differ depending on the underlying knowledge domain?
\end{enumerate}

The various fields of knowledge and scientific disciplines that contribute to their development have been developed in different ways.
Domain-specific learning can therefore benefit considerably from discipline-specific strategies of knowledge acquisition and processing \citep{List:Alexander:2019}.
If this is true, then we might expect a shadowing of these differences in Wikipedia: it is then likely that different fields of knowledge manifest themselves differently in Wikipedia, while the similarities of languages as manifested in Wikipedia are strongly conditioned by the reference to these fields.
By answering Q2, we inform educational research about the manifestation density of certain knowledge domains in certain language editions.
\textcolor{black}{This research can help to avoid wrong conclusions from generalizations for knowledge domains or languages: 
different outcomes of learners of different languages, for example, may be the result of differences in such resources and not necessarily the result of different linguistic structures of the underlying task descriptions \citep{Mehler:Hemati:Uslu:Luecking:2018}.}

\begin{enumerate}
	\item[Q3] Regardless of such differences, is there a knowledge-related \enquote{lingua franca} which, by its Wikipedia, makes the dissemination of knowledge in other languages predictable and thus serves as a reference for knowledge dissemination? 
	%
\end{enumerate}

English Wikipedia could play the role of such a reference due to its size and status as the primary source of translation between Wikipedias \citep[cf.][]{Warncke-Wang:et:al:2012}.
However, several studies question this role \citep{Hecht:Gergle:2010:a,Samoilenko:et:al:2016}. 
Thus, the question arises whether these results also apply to our combined intra- and intertextual model.

%

\begin{figure}[t]
	\centering
	\includegraphics[width=1.00\linewidth]{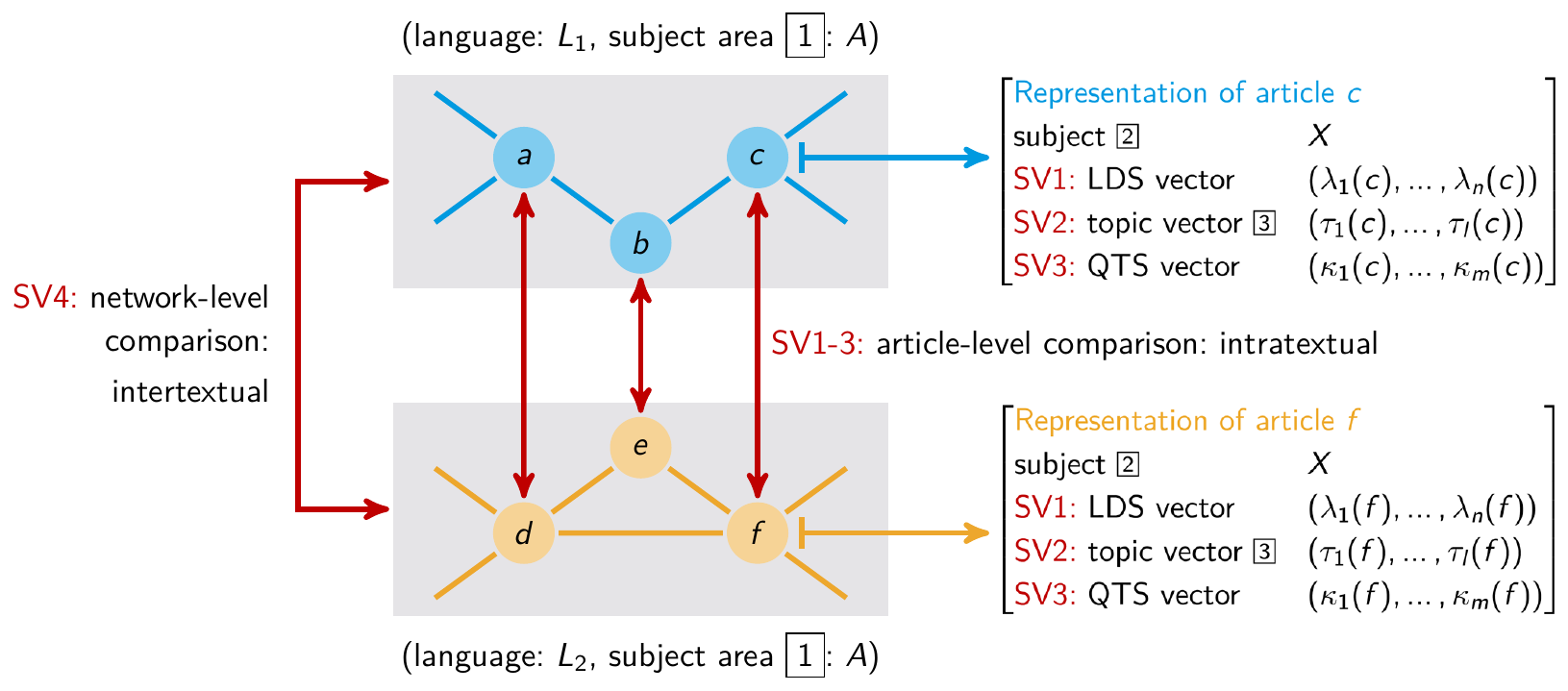}
	\caption[Topic model]{Three-level topic model of subject areas [1], subject instances [2] and topic distributions [3] extended by \textit{Quantitative Text Structure} (QTS) and \textit{Logical Document Structure} (LDS) vectors (of $m$ and $n$ dimensions, respectively). SV stands for \textit{Similarity View}.}
	\label{fig:ThreeLevelTopicModel}
\end{figure}

To answer Q1--3, we develop a method to extract and compare Wikipedia article graphs on the same topic, henceforth called \textit{item networks}.
Since we have to apply this method across languages, it must be both easy to implement and systematically reproducible, in such a way as to ensure that we are dealing with the same subject(s) regardless of the languages under consideration.
To realize this measurement operation with the help of Wikidata we compare item networks intra- and intertextually by means of a three-level topic model as depicted in Figure \ref{fig:ThreeLevelTopicModel}:
starting from any subject area (e.g.\ \textit{painting}) (1st level), we identify all its instances in Wikidata, which we use to extract all the articles within Wikipedia's language editions that address these instances. 
In this way, we identify the 2nd level of our topic model, whose elements we refer to here as subjects or subject instances (see Figure \ref{fig:ThreeLevelTopicModel}).
That is, a Wikipedia article is assigned a unique subject (e.g.\ \textit{Mona Lisa}) based on its Wikidata mapping, which specifies the entity (i.e.\ a \textit{Wikidata item} in the role of the \textit{definiendum}) that the article describes.
At this stage we get two topic assignments for each article:
the corresponding subject area (e.g.\ painting) and its subject instance (e.g.\ \textit{Mona Lisa}).
Using Wikipedia's article graph, where connections between articles are given by hyperlinks, we then obtain one article network per subject area and language, with the semantic coherence of these item networks resulting from the reference to the underlying subject area common to their articles.
In the third step we characterize each node of these item networks by three vectors (see Figure \ref{fig:ThreeLevelTopicModel}).
This applies in particular to the thematic perspectives under which subjects are treated.
In this way we reach the 3rd level of our topic model:
articles as manifestations of the same subject area are characterized according to the thematic perspectives (contributing to the \textit{definiens}) under which they describe their subject where these perspectives are modelled as topic vectors.

\begin{figure}[t]
	\centering
	\includegraphics[width=1.00\linewidth]{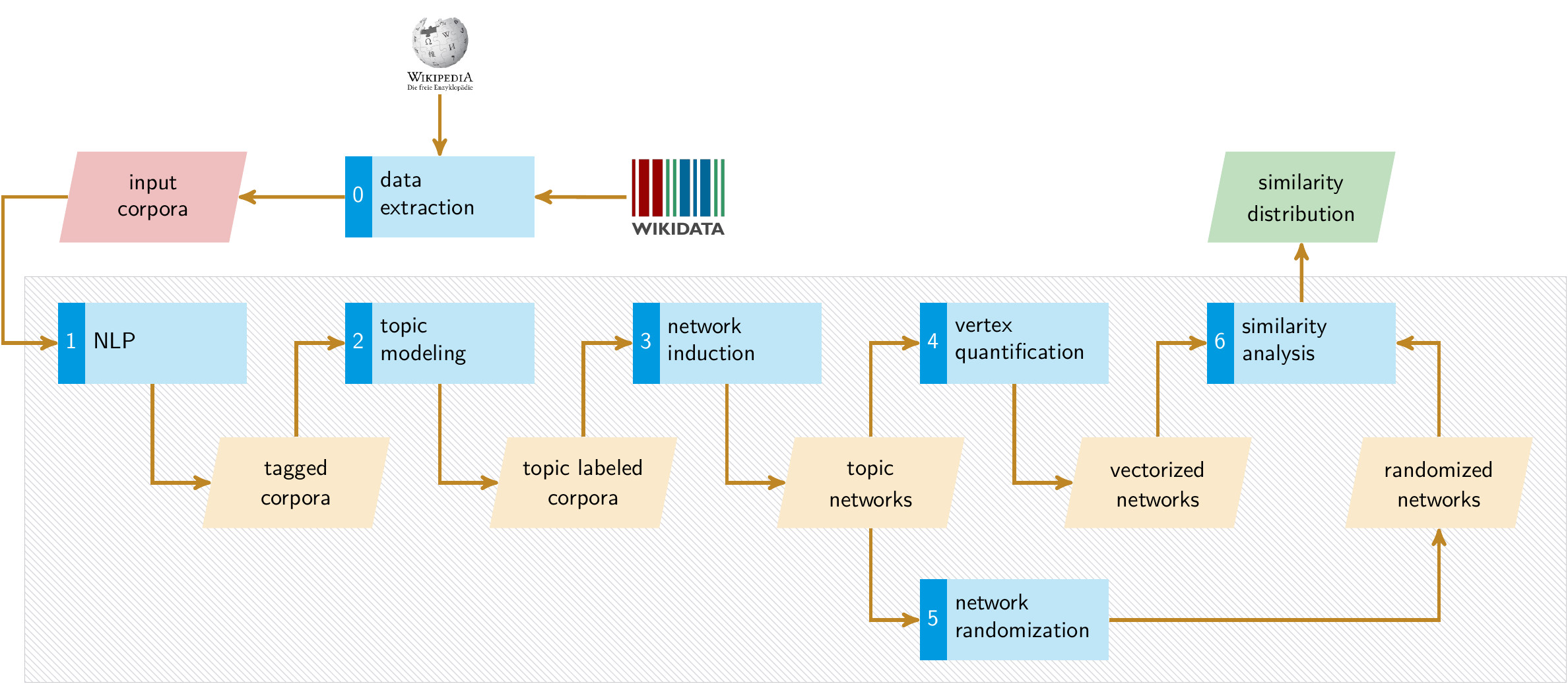}
	\caption[Procedure]{A procedure for measuring the similarity of thematically homogeneous article networks.}
	\label{fig:Procedure}
\end{figure}

Suppose the subject of a given article in a certain language is the painting \textit{Bal du moulin de la Galette} (by Pierre-Auguste Renoir) which instantiates the subject area \textit{painting}. 
Further, assume that our algorithm detects that this article deals with the topic economy (since it reports that this painting is one of the most expensive ever auctioned).
Then we get three topic-related assignments (see Figure \ref{fig:ThreeLevelTopicModel}): 
regarding the article's (abstract) subject area, its (concrete) subject and the thematic perspective of their descriptions.
Having done this for whole article networks of different languages, we can then compare these networks at the subject level by asking whether articles about the same subject thematize it in similar ways -- this concerns what we call \textit{intratextual similarity} -- and at the subject area level by asking whether the article networks are structurally similar in terms of what we call \textit{intertextual similarity}.
In this way, we implement the procedure to answer Q1--3, as shown in Figure \ref{fig:Procedure}.
Under the null hypothesis, different language article networks for the same subject area are very similar, both in terms of the articles' descriptions of the same subjects (intratextual similarity) and in terms of their hypertext structure (intertextual similarity).
We will essentially falsify this hypothesis and answer Q1--3 accordingly.
We now turn to explaining Step 0--6 of Figure \ref{fig:Procedure} in detail.

\subsection{Data Extraction}\label{sec:Data Extraction}

\begin{table}[t]
	\centering
	\begin{tikzpicture}[
	Zelle/.style={rectangle, fill=gray!20, draw=white, ultra thick, text width=4.0cm, inner sep=0.1cm, align=center, minimum height=1.25cm},
	Inhalt/.style={rectangle, fill=SeminarSehrHellBlau, draw=white, ultra thick, text width=4.0cm, inner sep=0.1cm, align=center, minimum height=1.25cm},
	Inhalt2/.style={rectangle, fill=SeminarSehrHellGruen, draw=white, ultra thick, text width=3.5cm, inner sep=0.1cm, align=center, minimum height=1.25cm},
	Left/.style={rectangle, text width=2.33cm, inner sep=0.1cm, draw=white, ultra thick, fill=SeminarSehrHellGruen, align=center, minimum height=0.7cm},
	Right/.style={rectangle, text width=8.2cm, inner sep=0.1cm, draw=white, ultra thick, fill=SeminarSehrHellBlau, align=center, minimum height=0.7cm},
	]
	
	\node [Zelle] at (0,0) {
		$
		\cup_{x\in\mathbb{O}} \{\check{B}_{\check{x}} \}, 
		\cup_{x\in\mathbb{O}} \check{\mathbb{B}}_{\check{x}}
		$
        {\footnotesize \numprint{1333421}}
	};
	\node [Zelle] at (4.2,0) {
		$
		\cup_{x\in\mathbb{O}} \{\hat{B}_{\check{x}} \}, 
		\cup_{x\in\mathbb{O}} \hat{\mathbb{B}}_{\check{x}}
		$
        {\footnotesize \numprint{24410338}}
	};
	\node [Zelle] at (0,-1.275) {
		$
		\cup_{x\in\mathbb{O}} \{\check{B}_{\hat{x}} \}, 
		\cup_{x\in\mathbb{O}} \check{\mathbb{B}}_{\hat{x}}
		$
		{\footnotesize \numprint{3190562}}
	};
	\node [Zelle] at (4.2,-1.275) {
		$
		\cup_{x\in\mathbb{O}} \{\hat{B}_{\hat{x}} \}, \cup_{x\in\mathbb{O}} \hat{\mathbb{B}}_{\hat{x}}
		$
        {\footnotesize $>$\numprint{32949992}}
	};
	
	\node [Inhalt2] at (-4.00,0) {unexpanded\\$({A}_{\check{x}})$} ;
	\node [Inhalt2] at (-4.0,-1.275) {expanded\\$({A}_{\hat{x}})$};
	
	\node [Inhalt, minimum height=0.7cm] at (0,0.985) {unexpanded} ;
	\node [Inhalt, minimum height=0.7cm] at (4.2,0.985) {expanded};
	
	\node [Left, rotate=90] at (-6.225,-0.645) {$A$-hierarchy};
	\node [Right] at (2.10, 1.705) {$B$-hierarchy};
	\end{tikzpicture}
	\caption{Variants of using Wikidata items to identify Wikipedia articles as vertices of INs.}\label{tab:Statistics of INs}
\end{table}

\begin{figure}[t]
	\centering
	\includegraphics[width=0.67\linewidth]{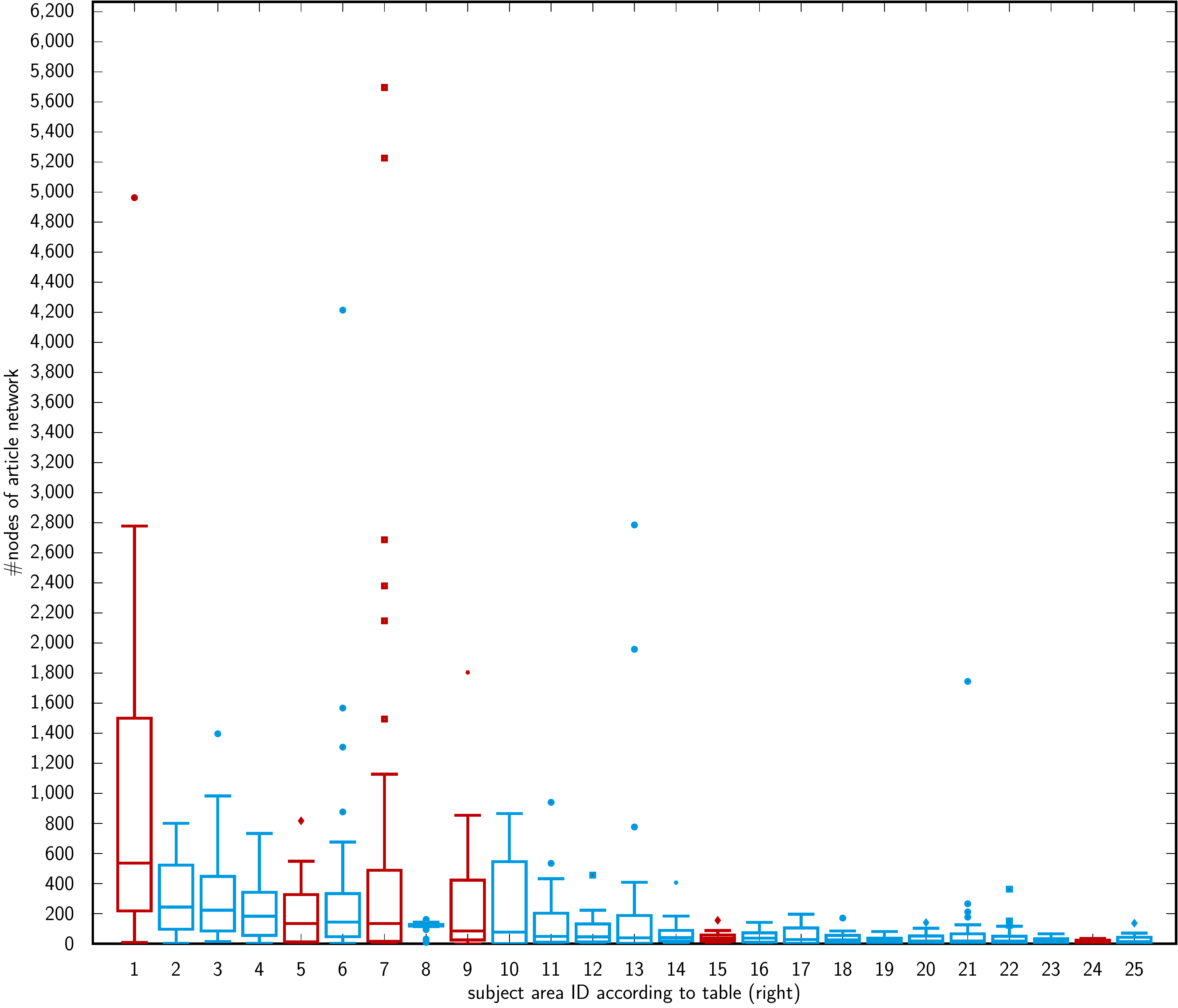}
	\hfill
	\scalebox{.54}{
		\input{./Tabellen/Subject-Areas.tex}
	}
	\caption{Boxplots of node distributions of article graphs extracted for the corresponding subject areas and language subsets as shown in the legend on the right. Red plots signal 6 additionally selected areas.
	}
	\label{fig:boxplotsNodes}
\end{figure}

In order to extract article networks on the same subject area relevant to educational science, that is, in order to perform Step 0 and 3 of Figure \ref{fig:Procedure}, we use the classification of the fields of science and technology of \citet{OECD:2007} and its correspondents in Wikidata.
The resulting networks, induced separately for each Wi\-ki\-pe\-dia, are called \textit{Item Networks} (IN) to emphasize the way they are induced by means of Wikidata items. 
In this way we guarantee three things: 
(i) the reference to subject areas such as those addressed by the PISA studies, (ii) a thematic breadth of the selected topics and (iii) their transferability between Wikipedias.
INs are generated as follows: 
for each of the OECD categories that can be assigned to Wikidata, we consult its \textit{studies}-statements to determine all Wikidata classes that are related to the respective OECD category in this way.
This is necessary because we need to move from OECD categories (e.g.\ \textit{art}) (which we refer to for ensuring thematic diversity of general topics) to subject areas (e.g.\ \textit{painting}) as likely destinations of searches in the context of the former:
%
while the OECD categories and their Wikidata subclasses induce \textit{subclass of}-hierarchies of fields of science and technology (henceforth called \textit{$A$-hierarchies}), the latter induce \textit{subclass of}-hierarchies of subject areas (cf.\ Figure \ref{fig:ThreeLevelTopicModel}) that are studied in these fields (called \textit{$B$-hierarchies}).
At this point we need an extension to ensure a higher coverage rate of OECD categories.
The reason for this is that some of them do not have \textit{studies}-statements in Wikidata, so that they would fall out of the selection process to the detriment of the targeted thematic breadth.
Therefore, we additionally examine the descendants of OECD categories in $A$-hierarchies to determine additional classes from $B$-hierarchies by means of these descendants' \textit{studies} statements.
This leads to a number of alternatives for sampling INs (see Table \ref{tab:Statistics of INs}), of which only a subset is feasible, as is now formally explained:
let $\mathcal{D} = (V_\mathcal{D}, A_\mathcal{D}, \lambda)$ be a representation of Wikidata as a directed graph with the set $V_\mathcal{D}$ of vertices (so-called Wikidata items), the  set $A_\mathcal{D}$ of arcs (links) between these items and the arc labeling function $\lambda$. 
Further, let
%
$\mathbb{O} = \{x, y, \ldots \}$
%
be the set of OECD categories. 
For a given OECD category $x\in \mathbb{O}$, we generate the set ${A}_{\hat{x}}$ of all items belonging to the $A$-hierarchy dominated by $x\in V_\mathcal{D}$:
\begin{equation}
{A}_{\hat{x}} = \{v\in V_\mathcal{D}\mid 0 < \ged_{sco}(x,v) < |V_\mathcal{D}| \}\cup\{x\}
\end{equation} 
where $\ged_{sco}(x,v)$ is the length of the shortest directed path from $x$ to $v$ in $\mathcal{D}$ crossing only \textit{subclass of}-links. 
Alternatively, we dispense with this expansion and get the set ${A}_{\check{x}} = \{x\}$. 
In the case of ${A}_{\hat{x}}$, we then explore \textit{studies}-links of $x$ \textit{and} of its subclasses, while in the case of ${A}_{\check{x}}$, only those of $x$ are explored.
The resulting sets of $B$-level items that are \enquote{studied} in this sense are denoted by 
\begin{align}
{B}_{\hat{x}} &= \{w\in V_\mathcal{D} \mid \exists v\in {A}_{\hat{x}}\!: (v,w)\in A_\mathcal{D} \wedge \lambda((v,w))=\textit{studies}\}\\
{B}_{\check{x}} &= \{w\in V_\mathcal{D} \mid 
(x,w)\in A_\mathcal{D}\wedge \lambda((x,w))=\textit{studies}\}
\end{align}
Obviously, ${A}_{\check{x}}\subseteq {A}_{\hat{x}}$ and 
${B}_{\check{x}}\subseteq {B}_{\hat{x}}$.
Now we have two alternatives again:
either, we recursively explore \textit{subclass of}-links to get more subject area-related Wikidata items for comparing Wikipedias (this leads to $\hat{B}_{\hat{x}}$ and $\hat{B}_{\check{x}}$) or not (generating  $\check{B}_{\hat{x}} = {B}_{\hat{x}}$ and $\check{B}_{\check{x}} = {B}_{\check{x}}$).
A test shows that recursive expansions on the side of $B$-hierarchies (generating $\hat{B}_{\hat{x}}, \hat{B}_{\check{x}}$) leads to overly large subject area representations that induce computationally hardly processable article networks.
Take the example of the OECD category \textit{Mathematics}: 
if we expand this category on the side of $B$-hierarchies, we get candidate items like \textit{set}, which are dominated by \textit{mathematical concept} as the target of a \textit{study}-statement starting from \textit{Mathematics}.
But \textit{set} has many instances in Wikidata, many of which are not mathematical concepts.
Such examples, which occur frequently in the variants $\hat{B}_{\hat{x}}$ and $\hat{B}_{\check{x}}$, realize unwanted changes of subject area, so that $B$-sided expansions are mentioned here only as a theoretical alternative.
To get processable networks, we alternatively represent the elements of item sets $B_x \in \{\hat{B}_{\hat{x}}, \hat{B}_{\check{x}}, \check{B}_{\hat{x}}, \check{B}_{\check{x}} \}$ as singletons, each containing a single Wikidata item that is finally used to extract Wikipedia article graphs:
\begin{equation}
\mathbb{B}_{{x}} = \{\{x\}\mid x\in B_{{x}} \}
\end{equation}
Take the example of 
$\check{\mathbb{B}}_{\hat{x}}$, that is, the set of singletons, each containing a topic studied either directly or indirectly under OECD category $x$ -- $\cup_{x\in\mathbb{O}}\check{\mathbb{B}}_{\hat{x}}$ currently contains 172 labels of corresponding subject areas (see Table \ref{tab:Statistics of INs}).
\textcolor{black}{Obviously, 
	$\check{\mathbb{B}}_{\check{x}} \subseteq \hat{\mathbb{B}}_{\hat{x}},
	\hat{\mathbb{B}}_{\check{x}} \subseteq \hat{\mathbb{B}}_{\hat{x}},
	\check{\mathbb{B}}_{\hat{x}} \subseteq 
	\hat{\mathbb{B}}_{\hat{x}}$.
}
Having identified all Wikidata \textit{item sets} for each OECD category $x$, we get an expression for the set of candidate subject areas:
\begin{equation}\label{math:Item Set}
\mathbb{B} = \bigcup_{x\in \mathbb{O}} \{\hat{B}_{\hat{x}}, \hat{B}_{\check{x}}, \check{B}_{\hat{x}}, \check{B}_{\check{x}} \} 
\cup \hat{\mathbb{B}}_{\hat{x}}
\cup \hat{\mathbb{B}}_{\check{x}}
\cup \check{\mathbb{B}}_{\hat{x}}
\cup \check{\mathbb{B}}_{\check{x}}
\end{equation}
An element $B_x\in \mathbb{B}$ is a set of Wikidata items that are (in-)directly accessible from OECD category $x$ via \textit{studies}-links.
Item sets like $B_x$ serve to identify Wikipedia articles on the same subject across languages.
This is achieved as follows:
each item in $B_x$ allows for identifying a corresponding set of instances (e.g.\ the painting \textit{Bal du moulin de la Galette} by Pierre-Auguste Renoir) by exploring \textit{instance of}-links in Wikidata. 
These instances are linked from Wikipedia articles and correspond to subjects in Figure \ref{fig:ThreeLevelTopicModel} as instances of subject area $B_x$, which in turn is derived from OECD category $x$ to ensure that we address educationally relevant topics.
At this point, an important difference to Section \ref{sec:Wikipedia} becomes clear:
the approaches mentioned there operate on small, closed lists of abstract topic categories (similar to the OECD categories used here), whereas we use OECD categories only to address the level of subject areas and their subject instances (cf.\ Figure \ref{fig:ThreeLevelTopicModel}).
The reference to ILL does not solve the problem of these approaches, since ILLs merely define groups of articles on the same subject, to put it in our terminology.
In contrast to this, we use Wikidata as a whole to extract arbitrary alignable article networks from different Wikipedias, differentiating between subject areas, subject instances and article-wise topic distributions -- in this sense, our approach is both thematically stratified and open.
In any event, we limit sampling to $\check{\mathbb{B}}_{\hat{x}}$ for extracting article networks. 
The reason is that $\check{\mathbb{B}}_{\check{x}}$ and $\hat{\mathbb{B}}_{\check{x}}$ explore to few \textit{studies}-links, while variants that expand $B$-level items induce unwanted topic changes (see above).
From the 172 candidate subject areas belonging to $\check{\mathbb{B}}_{\hat{x}}$ (see Table \ref{tab:Statistics of INs}), we select the 19 largest ones supplemented by 6 areas. 
Figure \ref{fig:boxplotsNodes} shows the boxplots of the article networks' orders (i.e.\ number of their nodes) for each of these subject areas, which we derived from 35 Wikipedia language editions (see Table \ref{tab:languages}). 
For these 35 languages we trained topic models to detect the topic distributions of their articles (see Section \ref{sec:A Hybrid Approach to Measuring the Similarity of INs}, Level 3 of Figure \ref{fig:ThreeLevelTopicModel} and Step 2 of Figure \ref{fig:Procedure}).

\begin{table}[t]
	\centering
	\scalebox{.85}{
		\input{./Tabellen/Languages1.tex}
	}
	\vspace*{1cm}
	\scalebox{.85}{
		\input{./Tabellen/Languages2.tex}
	}
	\caption{Wikipedia language editions, which were analyzed thematically: \enquote*{Topics} is the number of DDC-based topic classes trained for the corresponding language, \enquote*{Train} is the number of training examples and \enquote*{F-score} the harmonic mean of precision and recall of the corresponding test.}
	\label{tab:languages}
\end{table}

Now, let $\mathbb{W} = \{\mathcal{W}_1\ldots, \mathcal{W}_{l}\}$ be the set of all Wikipedias $\mathcal{W}_i = (V_i, A_i)\in \mathbb{W}$ each represented as a directed graph with the set of vertices (i.e.\ articles) $V_i$ and arc set $A_i$.
Further, let $a\in \mathbb{B}$ be a subject area, then we induce for each Wikipedia $\mathcal{W}_i$ the \textit{Item Network} (IN)
$I_a(\mathcal{W}_i)$
of all articles on subjects that are directly or indirectly studied under the OECD category corresponding to $a$: 
it is defined as the subgraph of $\mathcal{W}_i$'s article graph that consists only of articles which by their Wikidata links are mapped onto instances of elements of $a$.
To generate this graph, we explore \textit{instance of}-links from elements of $a$ to Wikidata items addressed by Wikipedia articles.
Let $\iota\!: \cup_{i=1}^{l} V_i\to V_{\mathcal{D}}$ denote the function that links articles as instances to Wikidata items in the latter sense, then we get the following expression for $I_a(i)$ (see Step 3 of Figure \ref{fig:Procedure}):
\begin{equation}\label{math:IN}
I_a(\mathcal{W}_i) = I_i^a = (V_{i}^a, A_{i}^a) = (\{v\in V_i\mid \iota(v)\in a\}, \{ (v,w)\in A_i\mid v,w\in V_{i}^a\})
\end{equation}
This allows us to finally define the so-called \textit{alignment set} of any pair of INs $I_i^x = (V_{i}^x, A_{i}^x),$ $I_j^y = (V_{j}^y, A_{j}^y)$ which contains all pairs of articles of these INs that are alignable because of treating the same subject:
\begin{align}
\mathcal{W}_{ij}^{xy} &= 
\{(v,w) \mid v\in V_{i}^x \wedge w\in V_{j}^y \wedge \iota(v) = \iota(w) \} = (\mathcal{W}_{ji}^{yx})^{-1} 
\end{align}

\subsection{A Hybrid Approach to Measuring the Similarity of INs}\label{sec:A Hybrid Approach to Measuring the Similarity of INs}

So far, we described the object of measuring thematic dissimilarities of Wikipedia's language editions.
Now we define the aspects under which we measure these dissimilarities.
To this end, we consider (1) \textit{syntactic} (text-structural), (2) \textit{semantic} (topic-related) and (3) {text statistical} aspects of the similarity of INs. 
While these \textit{intratextual} aspects are all vertex content-related, a fourth \textit{Similarity View} (SV) focuses on the network structure of INs and thus on \textit{intertextual} aspects of similarity.
%
Our starting point is the observation that each Wikidata item can be addressed by several Wikipedia articles, and vice versa, the same article can describe several items, even if it is injectively mapped to Wikidata.
The calculation of similarities of INs then requires the mapping of network similarities on the level of node content (intratextual) \textit{and} network structure (intertextual).
For this purpose, we vectorize the textual nodes of INs according to the similarity views SV1-4.
That is, given a SV $s$, we assume that each node $v\in V_{i}^x$ of each IN $I_i^x = (V_{i}^x,  A_{i}^x)$ is mapped onto a corresponding vector $\vec{v} \in \mathbb{R}^{k_s}$.
This is expressed by the function 
\begin{equation}
h_s\!: V_{i}^x \to \mathbb{R}^{k_s}
\end{equation}
where $k_s$ is the dimensionality associated with SV $s$.
For INs $I_i^x = (V_{i}^x, A_{i}^x), I_j^y = (V_{j}^y, A_{j}^y))$ we  calculate their similarity node- and network-wise using the alignment set $\mathcal{W}_{ij}^{xy}$ 
(with a few exceptions, elements of vectors $\vec{v}$ are non-negative): $\forall (v,w)\in V_{i}^x\times V_{j}^y\!$:
\begin{alignat}{2}
%
\sigma_{s_1}(I_i^x, I_j^y) 
&= 
\begin{cases}
\frac{
	\sum\limits_{(v,w)\in \mathcal{W}_{ij}^{xy}}
	\sum_{n=1}^{k_s} h_s(v)[n] h_s(w)[n] 
}{
	\sqrt{\sum\limits_{v\in V_{i}^x} \sum_{n=1}^{k_s} h_s(v)[n]^2}
	\sqrt{\sum\limits_{w\in V_{j}^y} \sum_{n=1}^{k_s} h_s(w)[n]^2} 
}
& V_{i}^x \not=\emptyset\not= V_{j}^x \\
0 & else
\end{cases}
&&\in [-1, 1] 
\label{math:Vertex set-related cosine} 
%
%
%
\end{alignat}

Computing Formula \ref{math:Vertex set-related cosine} is part of performing Step 6 of Figure \ref{fig:Procedure}.
%
%
It requires defining vectorization functions $h_s$ for SV $s = 1..3$ (cf.\ Step 4 of Figure \ref{fig:Procedure}) (SV4 is introduced below):

\begin{enumerate}
	\item \textit{SV1: \textit{Logical Document Structure} (LDS):}
	according to this SV, pairs of aligned articles are the more similar the more their LDS \citep{Power:Scott:BouayadAgha:2003} resemble each other, and the more such pairs of aligned articles of two INs resemble each other in this sense, the more similar these INs are.
	For measuring similarities of the LDS of aligned articles we vectorize articles along $k_1 = 11$ dimensions (cf.\ \citealt{Callahan:Herring:2011}): 
	\textit{
		number of characters, 
		number of sections, 
		breadth of the table of content tree, 
		depth of this tree, 
		number of outgoing links to pages inside Wikipedia, 
		number of outgoing links to pages outside Wikipedia, 
		number of pictures, 
		{number of tables}, 
		number of links to the \emph{Integrated Authority File} and related norm data, 
		number of references
	} and \textit{number of categories}.
	In this way, we identify text pairs that address the same Wikidata item by similar document structures, e.g.\ with regard to the length of the presentation, the use of images, tables or hyperlinks.
	Due to its orientation on surface structural features, this method can easily be calculated across languages.
	
	\item \textit{SV2: Thematic structure:}
	according to this SV, pairs of aligned articles are the more similar, the more similar the distributions of topics they address when describing their subjects (see Figure \ref{fig:ThreeLevelTopicModel}), and the more such pairs of aligned articles of two INs resemble each other in this sense, the more similar they are.
	SV2 is implemented with text2ddc \citep{Uslu:Mehler:Baumartz:2019} (Step 2 in Figure \ref{fig:Procedure}), a neural network based on fastText \citep{Joulin:Grave:Bojanowski:Mikolov:2016} which uses TextImager \citep{Hemati:Uslu:Mehler:2016} to preprocess texts (Step 1 in Figure \ref{fig:Procedure}).
	That is, topics are identified as elements of the 2nd level of the \textit{Dewey Decimal Classification} (DDC), a topic model widely used in the field of libraries.
	To this end, each article is mapped onto a 98 dimensional topic vector, whose 
	%
	%
	membership values encode the degree to which text2ddc estimates that the article deals with the topic corresponding to the respective dimension \citep[cf.][]{Uslu:Mehler:Baumartz:2019,Mehler:Gleim:Gaitsch:Uslu:Hemati:2019}.
	In this way, we identify text pairs that tend to describe the same subject of the same area in different languages under the perspective of similar topic distributions.
	%

%
Since we do not have \textit{Part of Speech} (POS) taggers for all target languages, we basically pursued a word-form-related approach to train text2ddc: 
%
%
we generated training and test corpora by retrieving information from Wikidata, Wikipedia and the Integrated Authority File of the German National Library. 
Since Wikipedia is offered for a variety of languages, such corpora can be created for many languages.
We optimized text2ddc with regard to selected linguistic features as a result of various pre-processing steps such as lemmatization and disambiguation.
%
%
In the last column of Table \ref{tab:languages} we show the F-values obtained for the corresponding tests.
The highest F-value (87\%) is achieved for German (where we also explored POS data), the lowest for Hindi (74\%).
Although this is a wider range of values, it is currently the only way to compare the content of texts in different languages in terms of the way their subjects are treated.
And although text2ddc was trained for a larger number of languages\footnote{See \url{https://textimager.hucompute.org/DDC/}.}, we concentrated on those for which it achieves an F-value of at least 74\%.

\item \textit{SV3: \textit{Quantitative Text Structure} (QTS):}
according to this SV, pairs of aligned articles are the more similar the more their QTSs resemble each other, and again, the more such pairs of aligned articles of two INs resemble each other in this sense, the more similar they are.
%
%
%
%
To get comparable vector representations of the QTS of articles, we use a subset of 17 dimensions of quantitative linguistics as evaluated by \citet{Konca:Mehler:Baumartz:Hemati:2020}: 
\textit{adjusted modulus}, \textit{alpha}, \textit{Gini coefficient}, \textit{$h$-point}, \textit{entropy}, \textit{hapax legomena percentage}, \textit{curve length}, \textit{lambda}, \textit{vocabulary richness}, \textit{repeat rate}, \textit{relative repeat rate}, \textit{thematic concentration}, \textit{secondary thematic concentration}, \textit{type-token-ratio}, \textit{unique trigrams}, \textit{average sentence length} and \textit{number of difficult words}. 
%
\textcolor{black}{Two text characteristics from \citealt{Konca:Mehler:Baumartz:Hemati:2020}, which require POS tagging, are excluded from SV3, since POS-tagging tools were not available for all languages considered here.}
We additionally compute autocorrelations (lag 1 to 10) of consecutive sentence-related association probabilities with BERT (\textit{Bidirectional Encoder Representations from Transformers}) \citep{Devlin:et:al:2018}. 
BERT is a language model based on a bidirectional transformer \citep{Vaswani:et:al:2017}, which uses encoder-decoder attention \citep{Bahdanau:et:al:2014} and self-attention \citep{Cheng:Dong:Lapata:2016} mechanisms to calculate token representations conditioned on both left and right context.
In total, we compute $k_3$ = 27 characteristics to vectorize the QTS of any article of the 35 languages considered here.
%
	
\end{enumerate}

While SV1-3 are article content-related, graph similarity measures consider intertextual structures (SV4). 
We can calculate SV4 independently of SV1-3, or so that the contributions of SV1-3 are embedded at the node level of SV4.
To elaborate the first variant,
%
%
we consider four measures (thereby implementing Step 6 of Figure \ref{fig:Procedure}):

\begin{enumerate}
	\item As a baseline, we compute the \emph{Graph Edit distance-based Similarity} (GES) using the graph edit distance for labeled directed graphs \citep[cf.][]{Mehler:Gleim:Gaitsch:Uslu:Hemati:2019}. 
	%
	In this way we arrive at a measure that calculates graph similarities as a function of overlaps of node and arc sets, respectively.
	


	\item \emph{Edge-Jaccard-Similarity (EJS):}
	As a second baseline, we compute the arc set-based Jaccard similarity:
	\[ 
	\text{EJS}(I_i^x, I_j^x) = \frac{\abs{E_{i}^x \cap E_{j}^x}}{\abs{E_{i}^x \cup E_{j}^x}} \in [0,1]
	\]
	$\text{EJS}(I_i^x, I_j^x) = 1$ if and only if $I_i^x$ and $I_j^x$ have identical nonempty arc sets (where the Jaccard similarity of empty sets is defined to be zero).
	EJS decreases with increasing symmetric difference in the arc sets.

	\item \emph{DeltaCon with Personalized PageRank:}
	\citet{Koutra:et:al:2016} propose DeltaCon, a family of similarity measures for node-aligned graphs. 
	It is based on the distance function 
	\[ 
	d(I_i^x, I_j^x) = 
	\sqrt{\sum_{v \in V_{i}^x \cup V_{j}^x} \sum_{w \in V_{i}^x \cup V_{j}^x} \left( \sqrt{s_{i}(v,w)\strut} - \sqrt{s_{j}(v,w)} \right)^2} 
	\]
	where the columns of $s_i, s_j \in \mathbb{R}^{\abs{V_{i}^x \cup V_{j}^x} \times \abs{V_{i}^x \cup V_{j}^x}}$ store \emph{affinity} values between vertices $v, w \in  V_{i}^x \cup V_{j}^x$ in the union graphs ${I''}_{i}^{x} = (V_i^{x} \cup V_j^{x}, E_i^x)$ and ${I''}_j^x = (V_i^{x} \cup V_j^{x}, E_j^x)$.
	As $d(I_i^x, I_j^x)$ is a metric, \citeauthor{Koutra:et:al:2016} propose to define the DeltaCon similarity as $sim(I_i^x, I_j^x) = \frac{1}{1+d(I_i^x, I_j^x)}$.
	DeltaCon can be parameterized by various node similarity measures, of which we consider a variant that is semantically meaningful in our context:
	%
		%
		%
		Personalized PageRank (for vertices $v,w \in V({I''}_{i}^{x})$) measures the probability of being in vertex $w$ in the stable distribution of a random walk on ${I''}_{i}^{x}$ that has a probability of $1-\alpha$ to be reset to $v$ in each step.
		This models the probability of landing at an article $w$ when starting at article $v$, randomly following links and going back to article $v$ when one is ``lost''.
		It thus models, considering all articles in turn as starting articles, a crowd of Wikipedia users following hyperlinks to navigate the information landscape.
		We proceed similarly to compute $s_j$ and set $\alpha=0.85$.

\item As both EJS and DeltaCon are highly sensitive to non-overlapping node sets, we restrict the input INs $I_i^x = (V_{i}^x, A_{i}^x)$ and $I_j^x = (V_{j}^x, A_{j}^x)$ for these two similarity measures to the subgraphs induced by the intersection of their vertex sets, that is, ${I'}_{i}^{x} = {I}_{i}^{x}[V_i^{x} \cap V_j^{x}]$ and ${I'}_j^x = {I}_j^x[V_i^x \cap V_j^x]$.
This approach yields high similarities if the subgraphs on the aligned nodes are similar and disregards dissimilarity induced by 
unaligned nodes in both graphs. 
Hence, the latter two measures can be seen as ``optimistic'' and may give high similarities even if the overlap of two vertex sets is rather small.
To provide an alternative to this view, we compute the arc set-based cosine graph similarity of \citet{Mehler:Gleim:Gaitsch:Uslu:Hemati:2019}.
For this purpose we start from the following \enquote{axioms} concerning the similarity of INs:
\begin{enumerate}
\item[A1] The higher the number of shared subject instances, the more similar the INs.
\item[A2] The lower the proportion of shared subject instances in the total number of instances of the underlying subject area, the less similar the INs.
\item[A3] The higher the number of shared paths, the more similar the INs.
\end{enumerate}

A1 also concerns isolated vertices: 
two INs can be identical even if all their vertices are isolated. 
A2 damps this depending on the total number of shareable subjects: 
the smaller the orders of the INs in relation to this number, the less similar they are.
A3 prefers pairs of INs that share many edges, but only in cases where equal paths start from aligned subjects.
A3 essentially states that two networks are the more similar, the more similar they look like from the perspective of the more aligned nodes.
A1 and A3 are measured with the apparatus of \citet{Mehler:Gleim:Gaitsch:Uslu:Hemati:2019}; 
to satisfy A2, we damp the resulting cosine similarity by the quotient of shared nodes and the total number of shareable items; we can do the same regarding the size of the networks, but refrain from this \enquote{pessimistic} variant in the present paper. 
We will refer to this variant by \textit{Cosine Graph Similarity} (CGS):
it rates pairs of networks as similar that link large proportions of candidate Wikidata items in a similar way.
\end{enumerate}

%
\textcolor{black}{We calculate four graph similarity measures ranging from set-based to spherical measures, where EJS and DeltaCon weight similarities more optimistically and CGS more pessimistically.}

\subsubsection{Assessing Observed Similarities}\label{sec:Assessing Observed Similarities}

Any similarity found between INs on the same subject area has to be evaluated according to how far it is higher than what is randomly expected and lower than what is ideally expected (see Step 5 of Figure \ref{fig:Procedure}).
For this purpose, we consider the following bounds:
\begin{enumerate}
	\item To get a \textit{lower bound} we compute the similarities of random counterparts of INs:
	we consider Erd\H{o}s-R{\'e}nyi \citep{Erdoes:Renyi:1959} graphs
	$R(I_i^x)$ chosen uniformly at random from all graphs of the same order and size as $I_i^x$.
	%
	This randomization concerns SV4 without node-related similarities, assuming a bijection between the nodes of $R(I_i^x)$ and $I_i^x$ based on their Wikidata items.
	Randomizations are performed 100 times per IN; similarity values are averaged accordingly.
	\item As an \textit{upper bound} we consider the maximum similarity of different language INs observed over all subject areas in our corpus of INs.
	Since we display these values as a function of the minimum $\min(|V_i^x|, |V_j^x|)$, we get an estimate of the maximum similarity observed for this minimum among all similarities observed in our experiment. 
	Though the theoretical maximum is always 1 for identical graphs, this maximum is unlikely to be observable in practice even under the condition of comparatively similar INs. 
	Therefore, by our re-estimation we achieve a more realistic upper bound that is actually observable. 
	%
	%
	%
	\item To obtain the lower bound for SV1-3, the similarity of the non-aligned, randomly chosen articles from each pair of INs were calculated (using Formula \ref{math:Vertex set-related cosine}). The articles in each pair were drawn from two independent random permutations, whereupon the number of pairs (of articles) was kept the same as in the aligned case. To reduce the impact of possible outliers, the results of each calculation were averaged over 100 independent runs.
\end{enumerate}

\section{Experiment}\label{sec:Experiment}

Applying the methods of Section \ref{sec:Rationale and Method} along the procedure of Figure \ref{fig:Procedure} to the Wikipedia editions of Table \ref{tab:languages} to generate language-specific INs for the subject areas of Figure \ref{fig:boxplotsNodes} produces the results of Table \ref{tab:Longtable}:
based on the number of INs per subject area listed in Table \ref{fig:boxplotsNodes} (whereby languages whose INs according to Formula \ref{math:IN} correspond to the empty graph for the given subject area are not listed), we arrive at \numprint{103229} 
graph comparisons 
using our 7 similarity measures, three of which are vertex (SV1-3) and four hypertext structure-oriented.\footnote{If all INs were non-empty, this number would equal $\#_{\text{similarities}} \cdot \#_{\text{topics}} \cdot \#_{\text{comparisons/language}} = 7 \cdot 25 \cdot ({35 \choose 2} + 35) = \numprint{110250}$.}
Since we randomize each IN 100 times to perform the same procedure for each random setting, the final number of graph comparisons performed equals \numprint{10322900}. 

We start our experiment with three subject areas, whose analyses span the similarity spectrum observed by us: 
\textit{chemical element} (OECD class \textit{Chemical sciences}), 
\textit{disease} (class \textit{Health sciences}) and language (\textit{Languages and Literature}).
Let us first consider the set-based similarity measure GES.
In the first column of Row 4 of Table \ref{tab:Longtable} we see the corresponding heatmap of the 35 languages' INs from Table \ref{tab:languages}:
the greener, the more similar the INs of the respective language pair for this subject area.
Apparently, we find that \textit{chemical element} is treated very uniformly across languages in terms of hypertext structure. 
This subject area is the maximum of what we observed regarding this uniformity:
it best approximates the ideal under the hypothesis that different language Wikipedias report uniformly on the same topic -- note that the rows and columns of the heatmaps are ordered according to the orders $|V_i^x|$ of the INs $I_i^x$.
This is confirmed by the curve displayed below the heatmap: 
it shows the similarity values listed in the heatmap as a function of $\min(|V_i^x|, |V_j^x|)$ (minimum of the orders of the input graphs $I_i^x, I_j^x$). 
In this way we see the influence of graph order: 
similarity values of comparisons with smaller graphs move to the left, those where both graphs are larger move to the right.
In fact, we see for this subject area that most comparisons concern (equally) large graphs achieving high GES-values.
This is contrasted by the last column of Row 4 which depicts the heatmap of subject area \textit{language}:
now, small graphs dominate the distribution with high similarity values, while pairs of significantly larger INs tend to have much lower values.
This example demonstrates a specialization of a few language editions on a broad representation of this subject area, while the majority of editions tend to underrepresent it.
All in all, we arrive at a zigzag curve of similarity values, which does not indicate a clear trend.
In the mid of the range of these two examples in Row 4, we find the subject area \textit{disease}:
the heatmap now suggests that the language pairs are rather dissimilar from the perspective of this subject area.
This is particularly evident from the similarity curve, which shows an almost constant trend of small values with a small increase towards larger graphs.
Let us now look at the same examples from the perspective of DeltaCon (Row 5); 
the situation remains essentially the same: 
\textit{chemical element} induces a more homogeneous mass of simultaneously larger and similar graphs, \textit{language} exhibits again a zigzag pattern, and \textit{disease} reveals slightest similarities regardless of the INs' orders. 
Obviously, the latter subject area seems to contradict the ideal of homogeneous, equally informative Wikipedias very strongly.
This is also confirmed by the randomizations according to Section \ref{sec:Assessing Observed Similarities}, which are presented as boxplots embedded in the similarity curves:
blue is the boxplot of the similarity values shown in the curve itself and red is the boxplot of the similarity values of the randomizations.
Interestingly, in the case of \textit{language} randomized similarities are even higher than their empirical  counterparts.
This can be explained by the topology of ER graphs, which tend to have short diameters making shorter paths more probable.
Anyhow, the situation is almost the same as in the case of Row 4: 
observed similarities seem to deviate only slightly from their randomized counterparts;
similarities between these INs resemble those of corresponding random graphs.
Now we look at CGS (Row 6), which evaluates graph similarities more \enquote{pessimistically} than DeltaCon.
This is confirmed by the heatmaps: 
in the case of \textit{chemical element}, the high similarity values change in favour of a \enquote{chessboard} view, while the binary regime disappears in the case of \textit{language}, whose similarity progression now resembles that for \textit{disease}.
In the latter two cases, low similarities dominate, with the tendency of higher similarities only for pairs of larger graphs: 
based on CGS, pairs of small graphs are highly dissimilar (damping effect).
Interestingly, the graph similarities as a function of the minimum of the orders of the input graphs no longer predominantly show a zigzag pattern as in the case of GES and DeltaCon: 
the impression of gradual transitions now prevails (2nd line of Row 6). 

The latter assessments are confirmed by the set of 25 subject areas.
Table \ref{fig:fuzzyGraphs} shows the DeltaCon-based boxplots displayed as a curve:
medians are represented by straight lines while the value ranges between the 25th and 75th percentile are colored accordingly;
the blue curve represents observed similarities, the red one their random counterparts (\enquote{lower bound}) and the orange one the corresponding upper bounds (see Section \ref{sec:Assessing Observed Similarities}).
Table \ref{fig:fuzzyGraphs}, in which the subject areas are arranged according to decreasing median similarities of their INs, shows a clear trend:
in almost all cases, observed similarities are below both the similarities of randomized INs and the upper bounds.
That is, observed similarities are far away from ideally equally informative Wikipedias, which would cover the corresponding subject areas uniformly for all languages.
Even more: as far as DeltaCon considers transitive dependencies of nodes along the same paths, it turns out that the INs' random counterparts even tend to have a greater hypertext-structural similarity -- as explained above, this is partly due to their small diameters.
Obviously, randomness makes networks seem more similar than if one follows existing walks in the real networks simultaneously. 
This even applies to the maximum values measured (upper bound).
At the same time, we observe a broad spectrum of similarity values ranging from a minimum of about 20\% (\textit{disease}) to a maximum of about 75\% (\textit{chemical element}).
Therefore it matters very much in which subject area one reads Wikipedia -- our approach shows this in a fine-grained way for educationally relevant topics using a three-level topic model.
The picture becomes even clearer when we look at the distribution of GES-based values in Table \ref{fig:fuzzyGraphs}:
observed similarities are now hardly distinguished from their random counterparts and far away from their upper bounds.
A more chaotic picture, somewhat reminiscent of DeltaCon, results from the Jaccard-based variant EJS: 
observed similarities tend to be even smaller than their random counterparts, but not always.
In Table \ref{fig:fuzzyGraphs}, the least chaotic pattern is produced by CGS: 
we observe a monotonically decreasing function with small differences for minimum and maximum values, where the similarity values fall below those of the other measures.

From this lesson we learn that subject areas are rather unevenly distributed across Wikipedia's language editions, whether one measures their similarities using simpler (GES, EJS) or more complex measures (DeltaCon, CGS). 
But what picture do we get from looking at the content-related similarities of articles?
In Table \ref{tab:Longtable} we start with SV1 (Row 1, LDS). 
Again, the picture is tripartite:
\textit{chemical element} marks the upper limit of observed similarities (the larger the INs, the higher their similarity), \textit{language} the lower limit and \textit{disease} a middle case.
\textit{language} also shows that LDS-based similarities break down into two groups: for larger and for smaller graphs, while members of both groups show no or little similarities between each other.
In any event, observed similarities tend to exceed their random counterparts (embedded blue (observed) and red (randomized) boxplots).
This tripartition is basically confirmed by SV2 (thematic similarities measured by text2ddc) (Row 2 in Table \ref{tab:Longtable}).
But now \textit{disease} turns out as a subject area whose article graphs (INs) consist of thematically more homogeneous articles -- more or less irrespective of the number of articles considered.
In any event, thematic similarity is again concentrated more or less on large graphs in the case of \textit{chemical element}.
The third case (Row 3 in Table \ref{tab:Longtable}) concerns quantitative text structures.
Here too, the general picture is confirmed: concentration on high values for large graphs in the case of \textit{chemical element}, more evenly distributed, but high values in the case of \textit{disease} and a bipolar picture in the case of \textit{language}. 
As before, randomized counterparts are exceeded.

The rather exceptional case of SV2 (thematic similarities) is confirmed by the overall view for the 25 subject areas in Table \ref{fig:fuzzyGraphs}:
observed similarities along SV2 are below the values for LDS (SV1) and QTS (SV3), while the latter are more evenly distributed across the subject areas, possibly reflecting a law-like behavior as described by quantitative linguistics \citep{Koehler:Altmann:Piotrowski:2005}.
In any case, with few exceptions, the observed values are again within the interval spanned by their randomized variants and upper bounds.
This points to similarity distributions far away from ideally homogeneously structured Wikipedia articles, which in addition would manifest almost the same topic distributions: 
actually, they do not.
Apparently hypertextual dissimilarity is parallel to textual dissimilarity:
\textit{what is rather dissimilar in terms of intertextual structure, tends to be dissimilar in terms of intratextual structure as well.}
However, we also observe the case of examples such as \textit{chemical element}, which has both high similarity values in terms of DeltaCon and SV2 (thematic similarity): 
this is, so to speak, the maximum of simultaneous inter- and intratextual similarity observed here.
In any event, our study combines intra- and intertextual measurements where the former are based on three views, regarding LDS, QTS and thematic text structures.
In this way, we obtain a more precise, broader view of article content than has been possible with the methods of related research.

\newpage

\begin{longtable}{|l|*{3}{m{0.29\textwidth}}|}
	\hline \centering
	& \centering chemical element & \centering disease & language \\
	\hline \centering 
	\rotatebox[origin=c]{90}{(1) SV1}&%
	\centering \includegraphics[width=.9\linewidth]{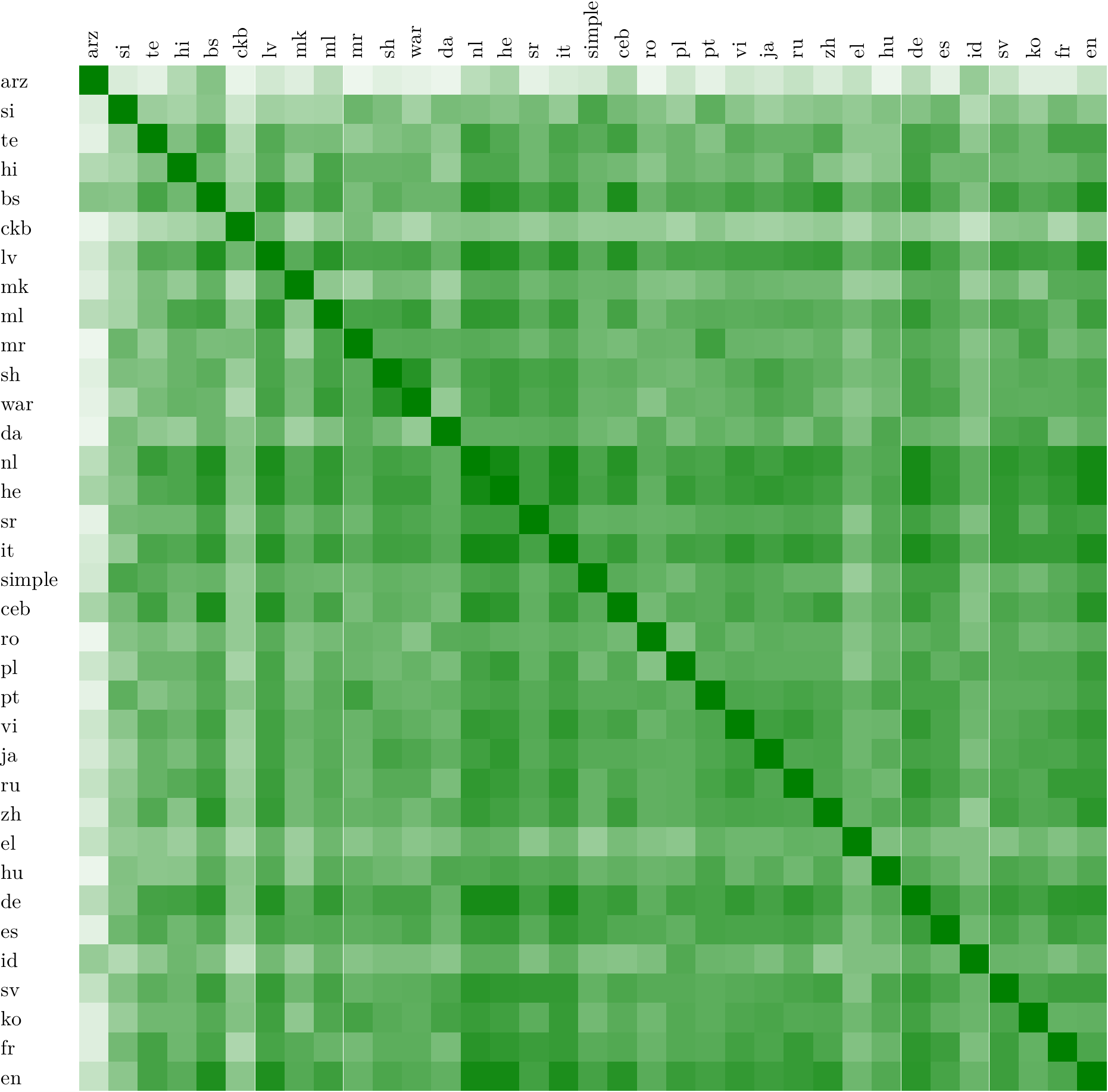}&%
	\centering \includegraphics[width=.9\linewidth]{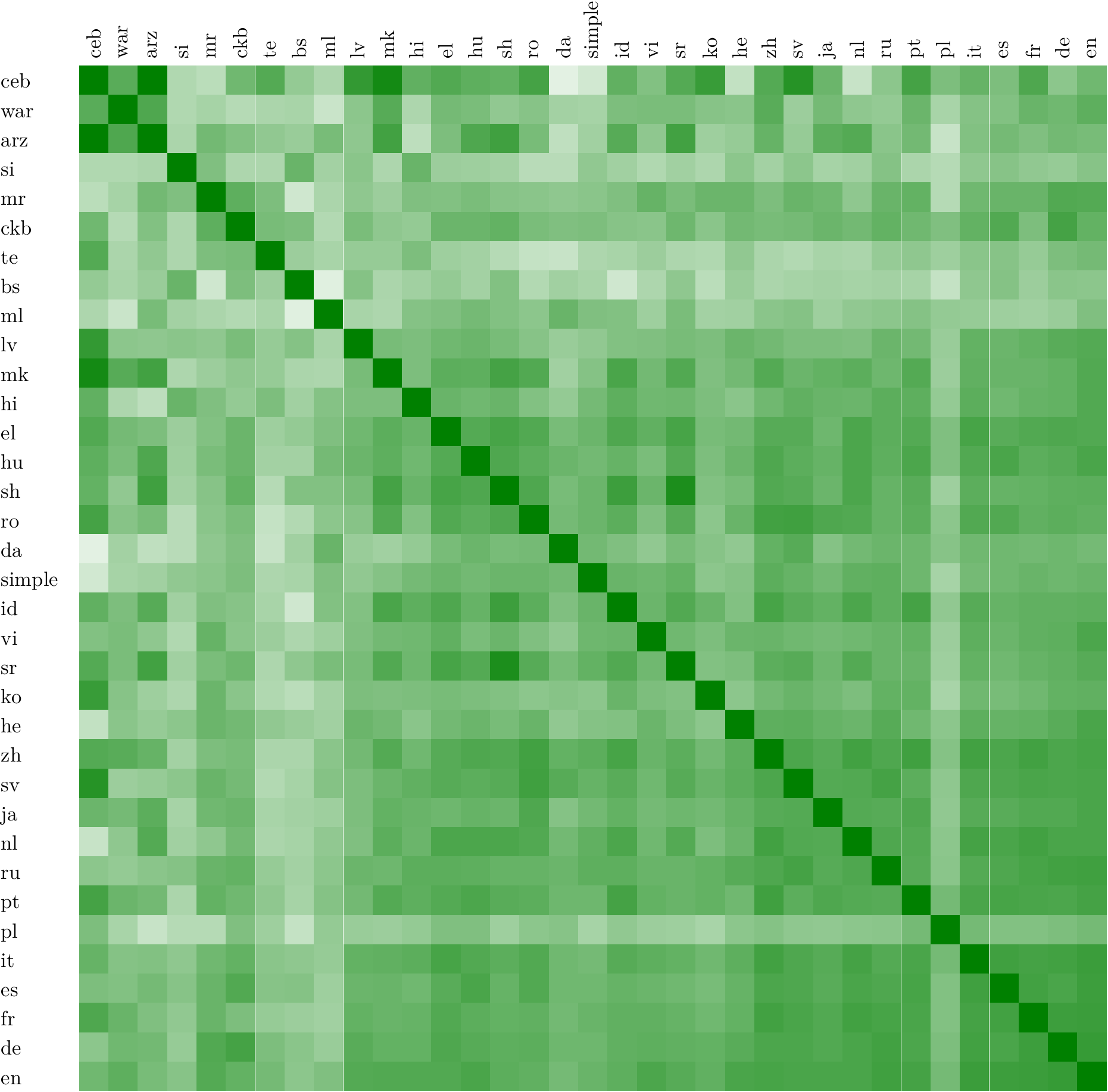}&%
	\includegraphics[width=.9\linewidth]{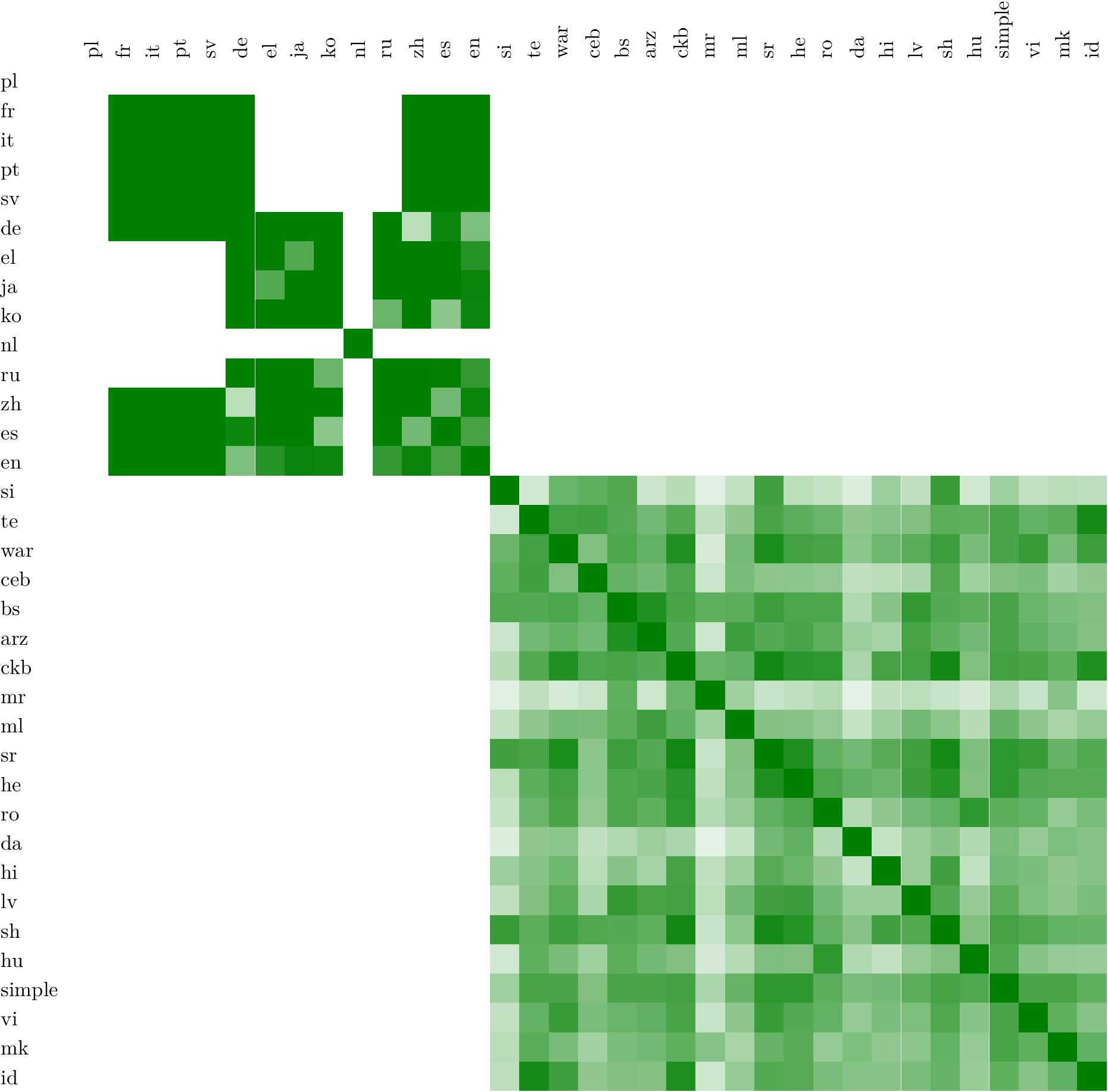}%
	\\%
	&%
	\centering \includegraphics[width=0.9\linewidth]{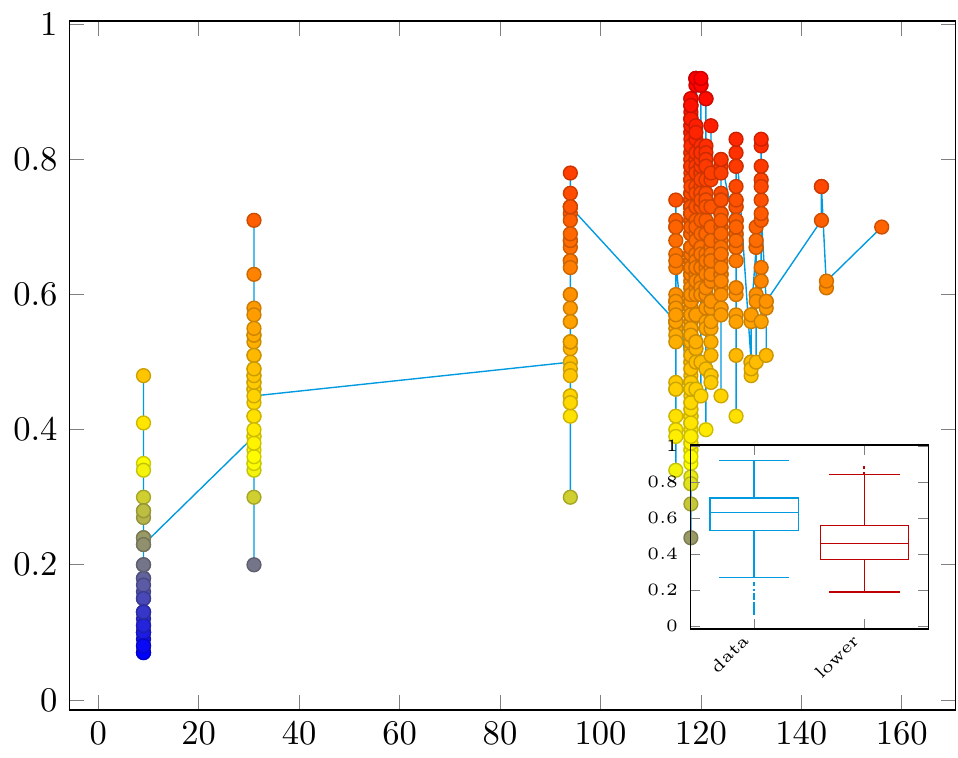}&
	\centering \includegraphics[width=.9\linewidth]{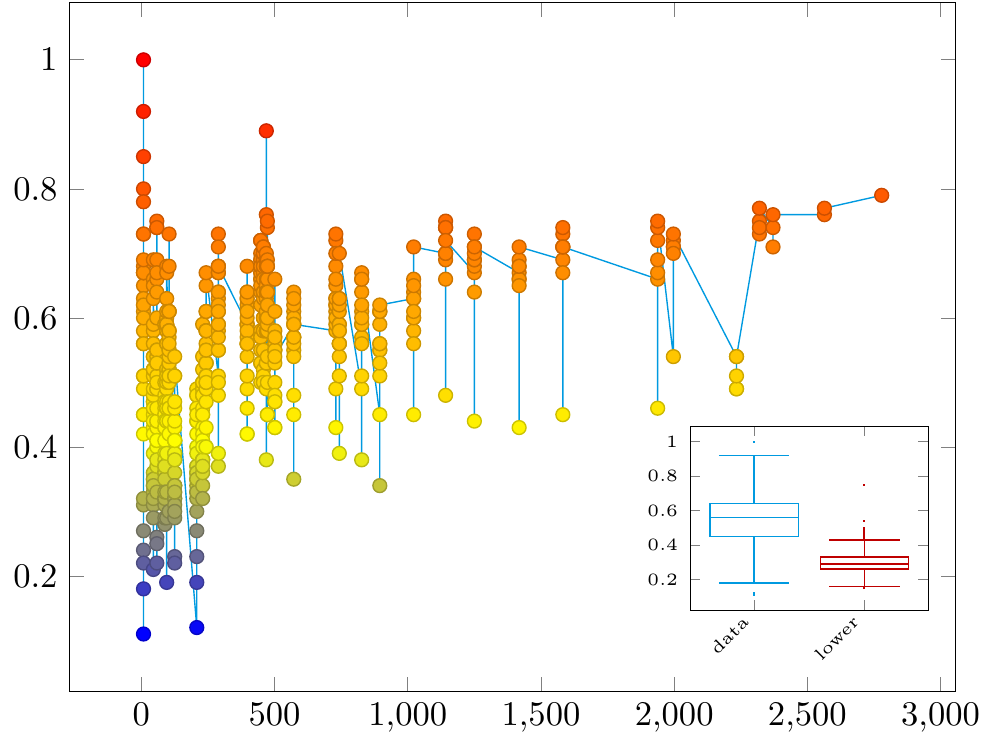} &
	\includegraphics[width=.9\linewidth]{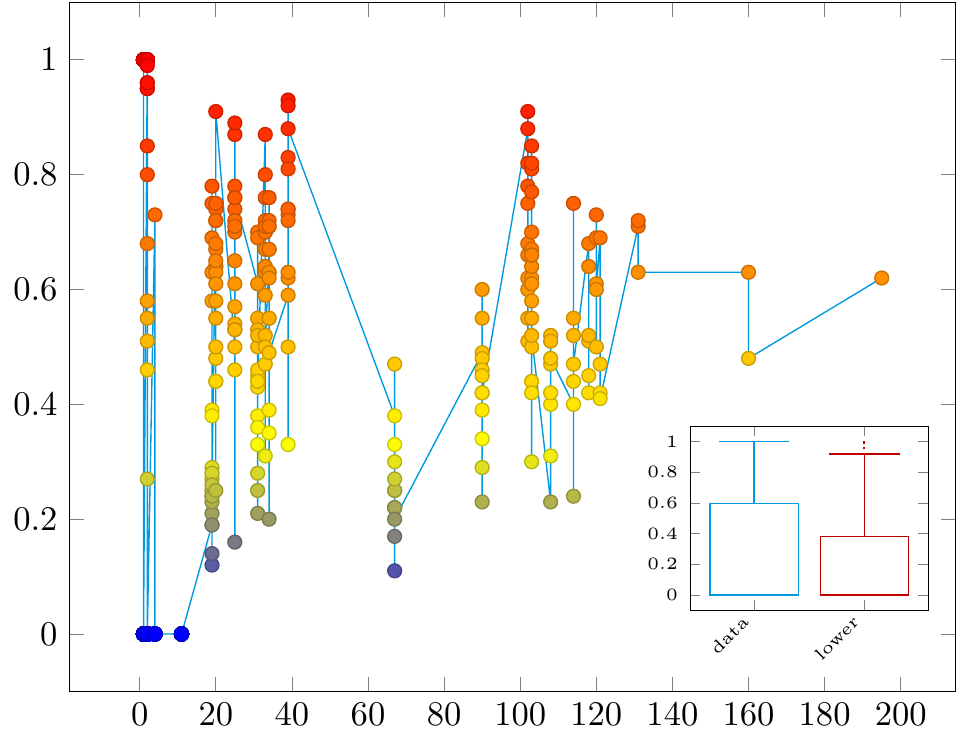} \\
	\hline \centering
	\rotatebox[origin=c]{90}{(2) SV2} &
	\centering \includegraphics[width=.9\linewidth]{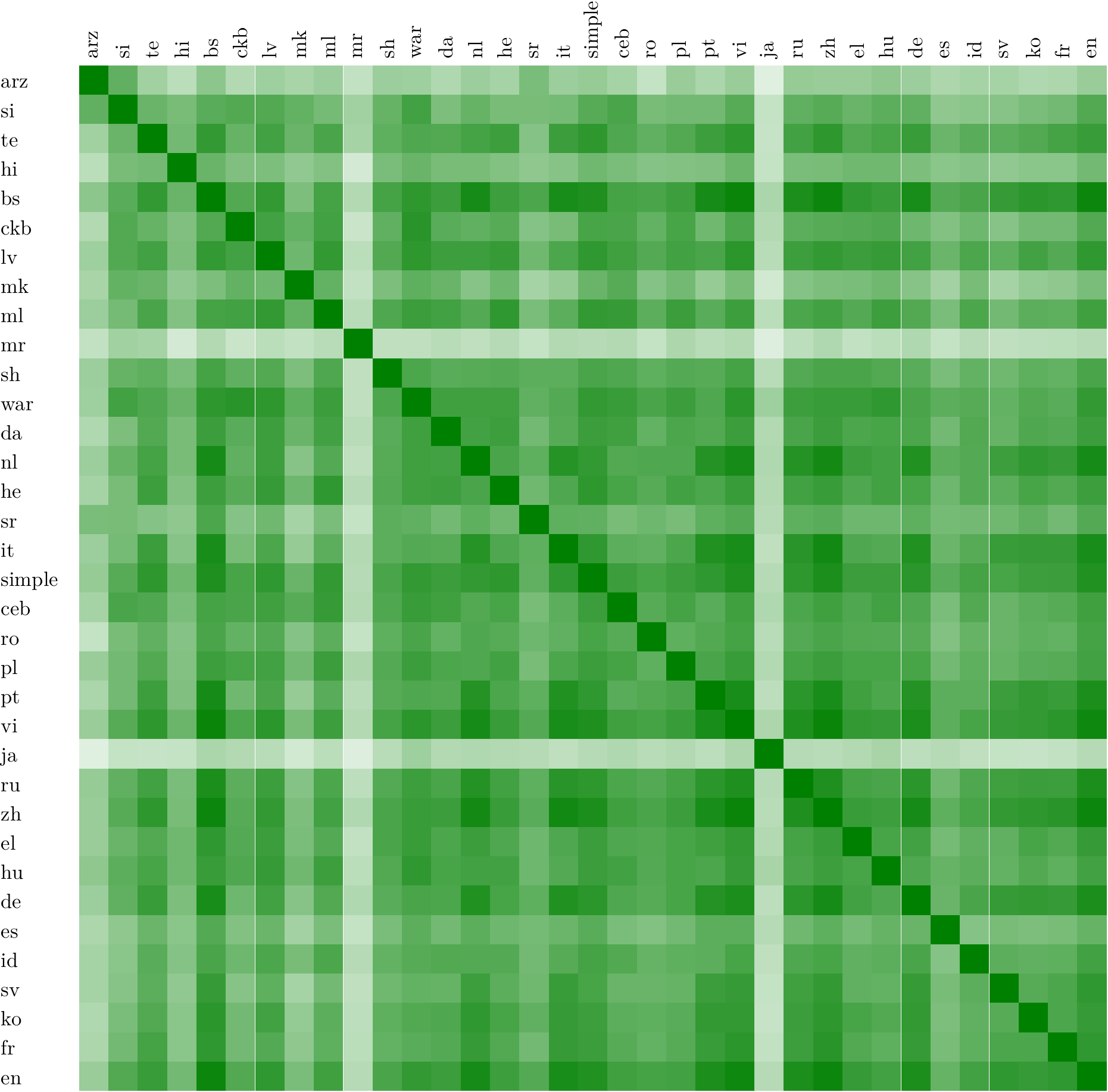}&
	\centering \includegraphics[width=.9\linewidth]{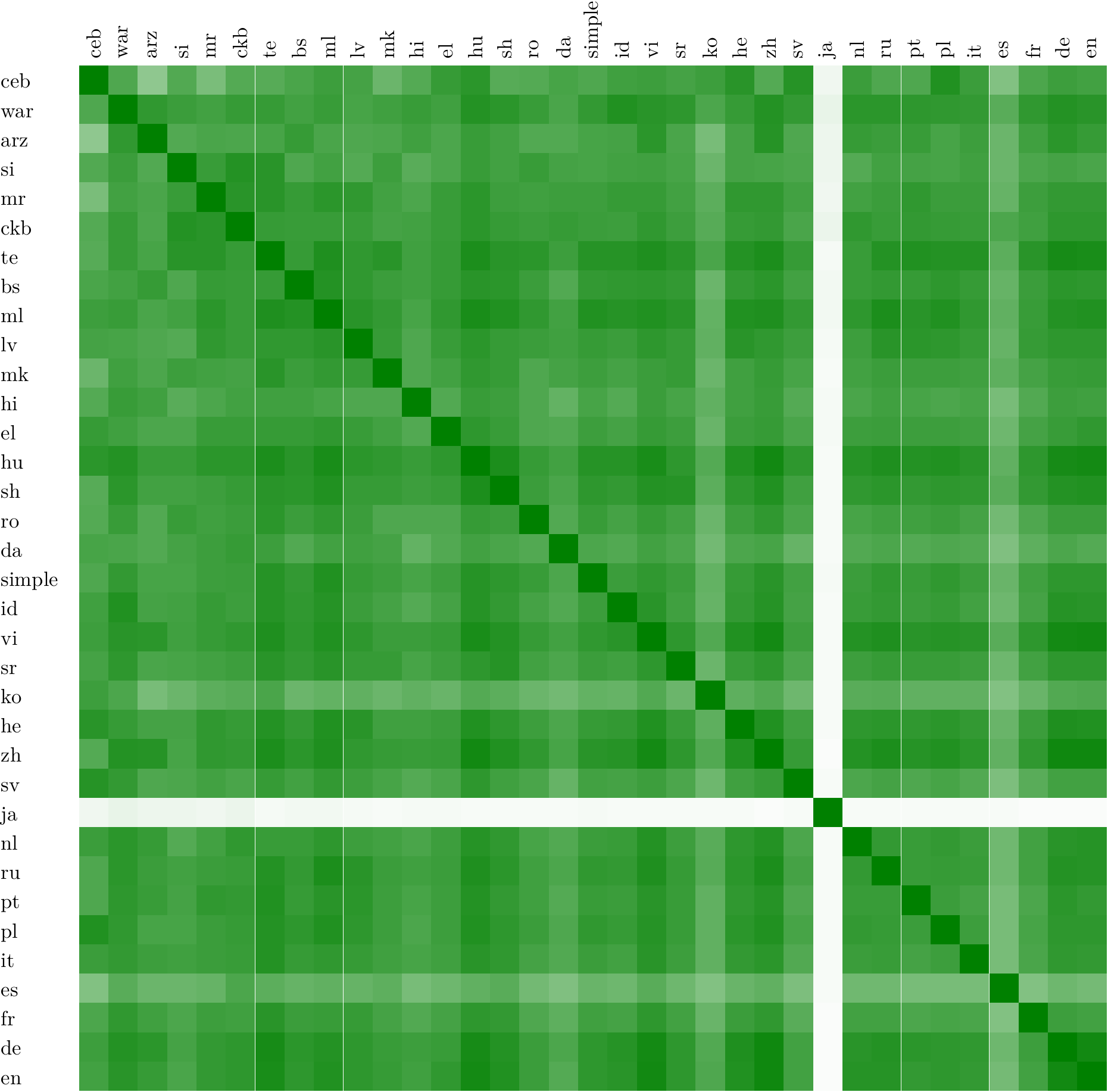} &
	\includegraphics[width=.9\linewidth]{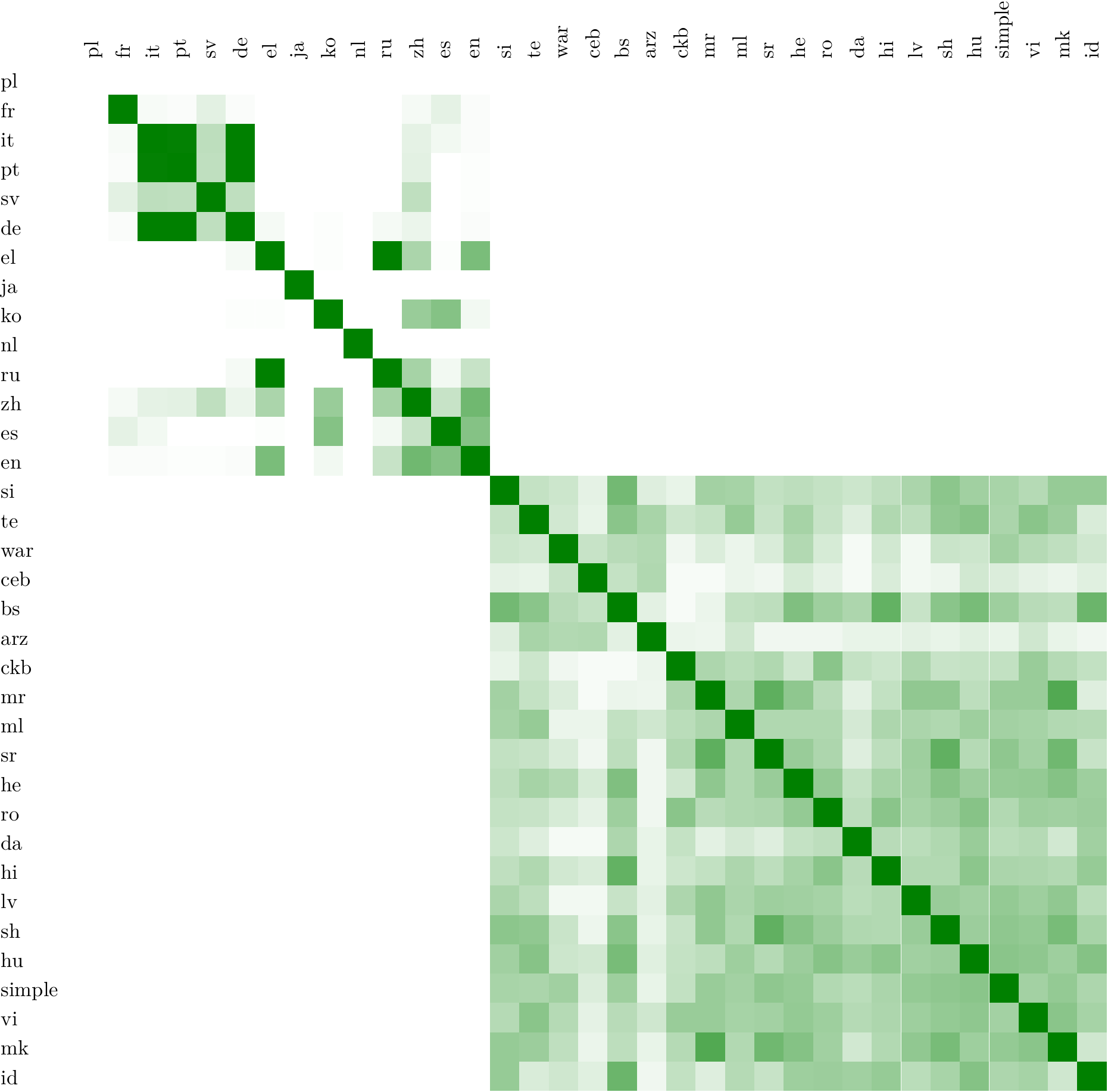} \\ 
	&
	\centering \includegraphics[width=0.9\linewidth]{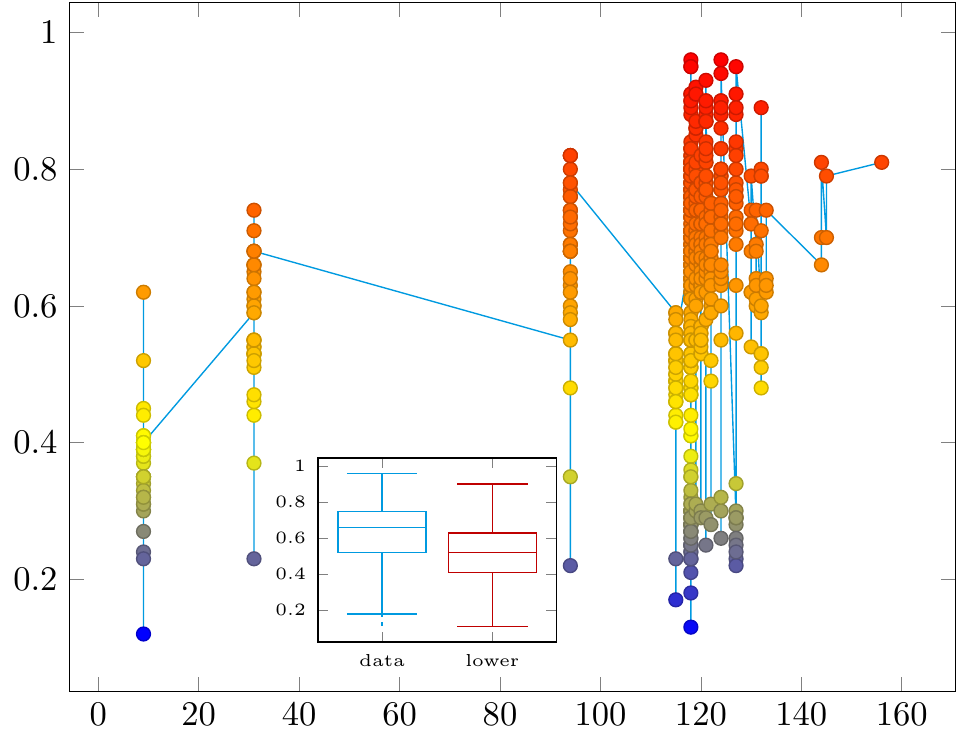}&
	\centering \includegraphics[width=0.9\linewidth]{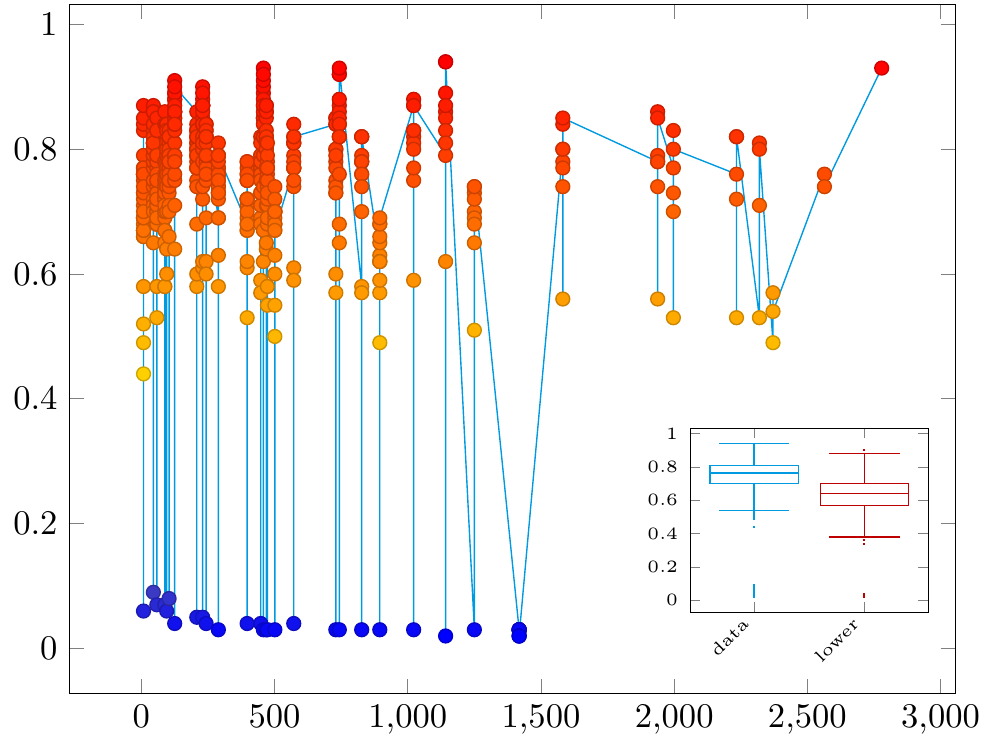} &
	\includegraphics[width=0.9\linewidth]{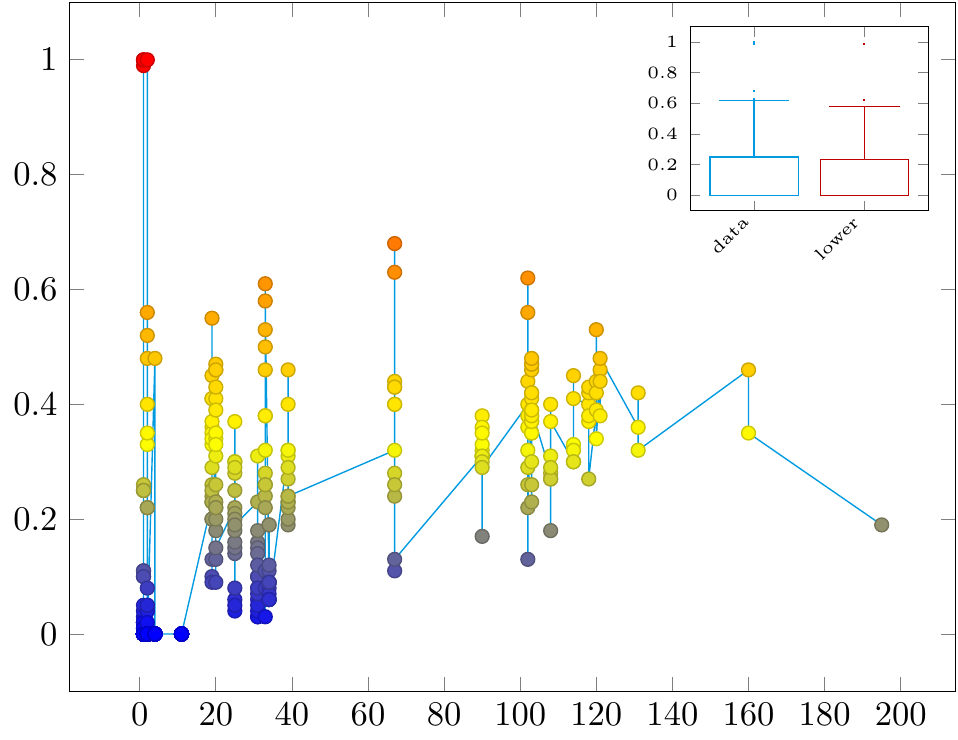} \\
	\hline \centering
	\rotatebox[origin=c]{90}{(3) SV3} &
	\centering \includegraphics[width=.9\linewidth]{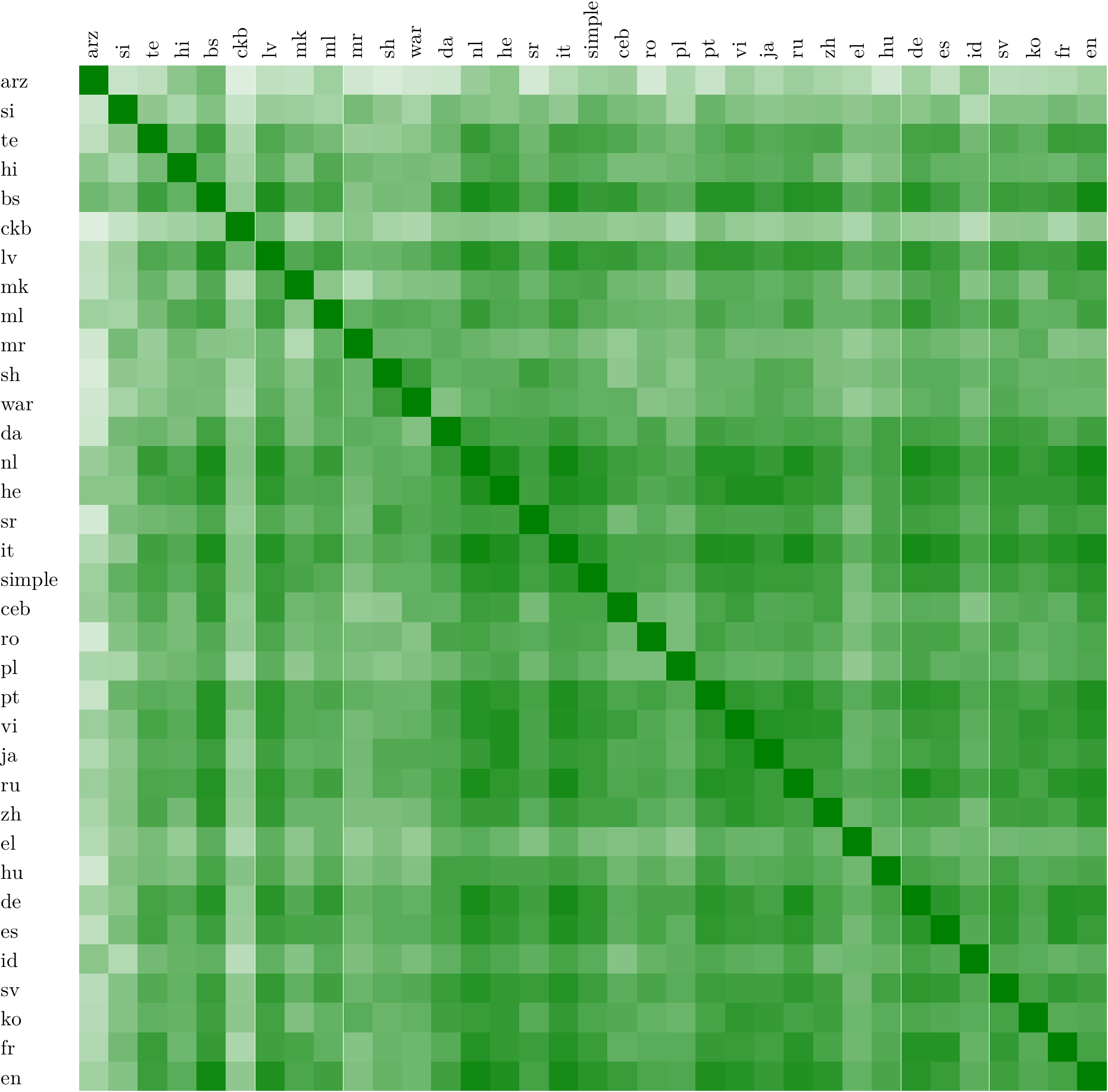}&
	\centering \includegraphics[width=.9\linewidth]{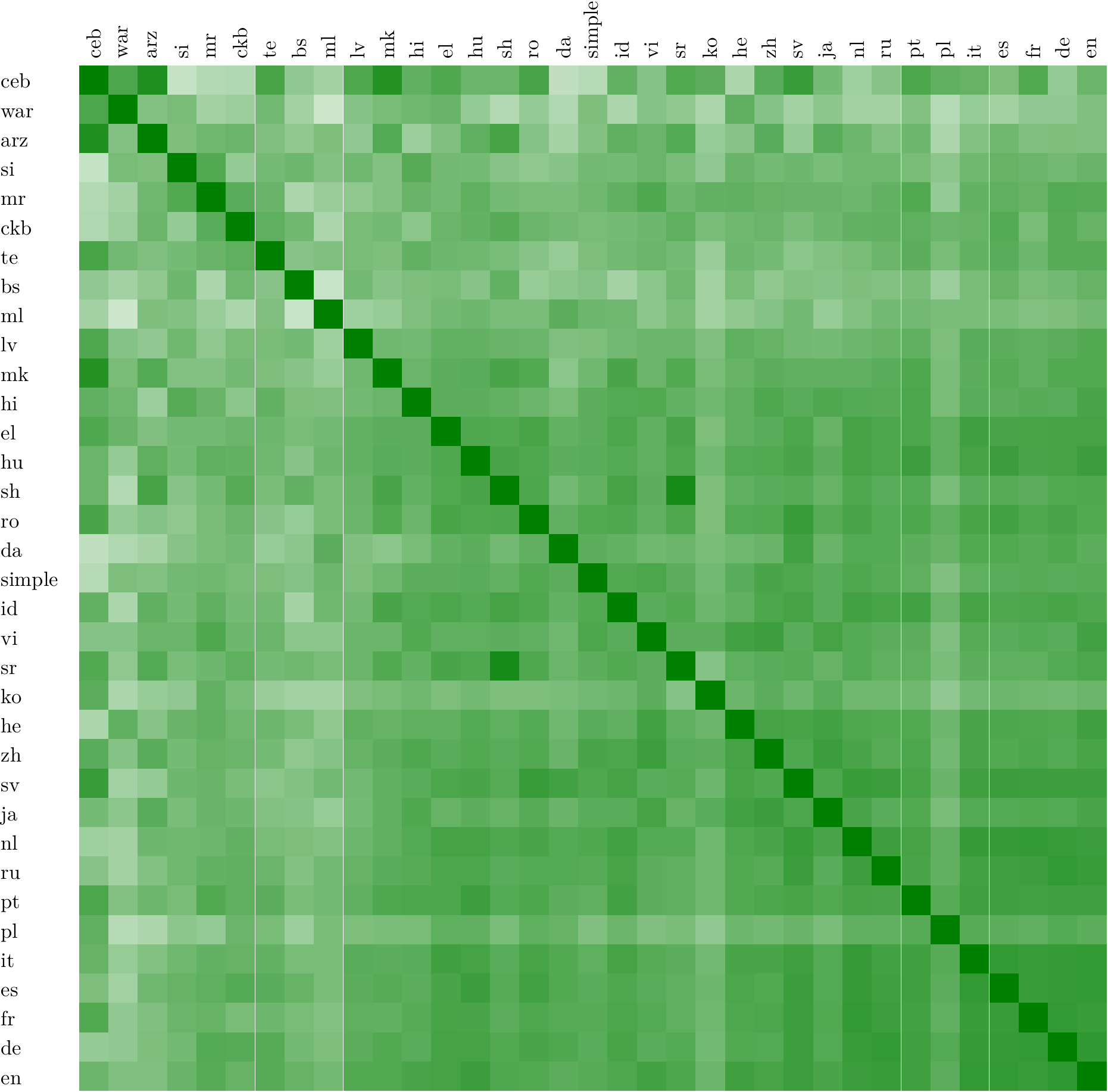} &
	\includegraphics[width=.9\linewidth]{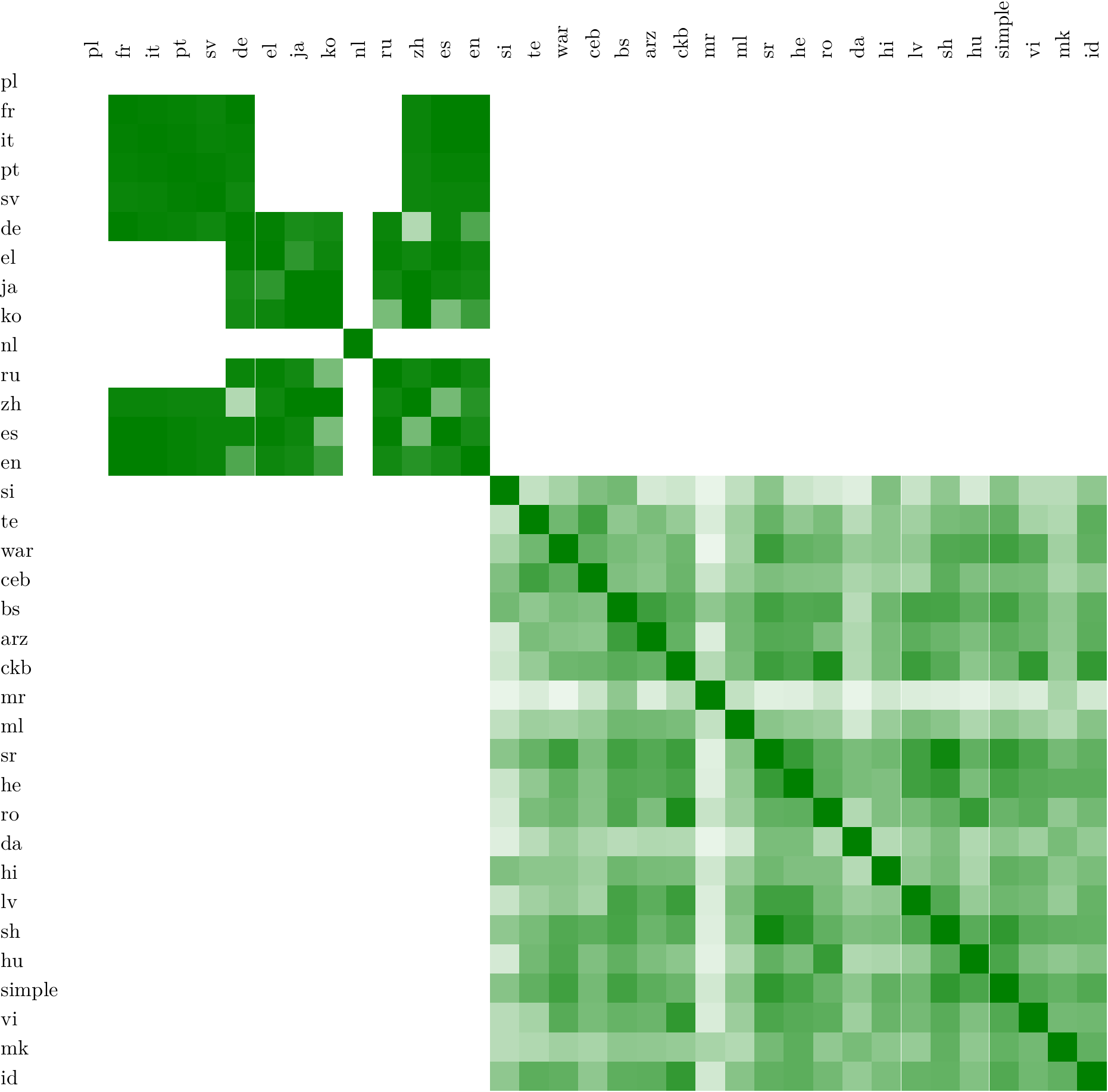} \\ 
	&
	\centering \includegraphics[width=0.9\linewidth]{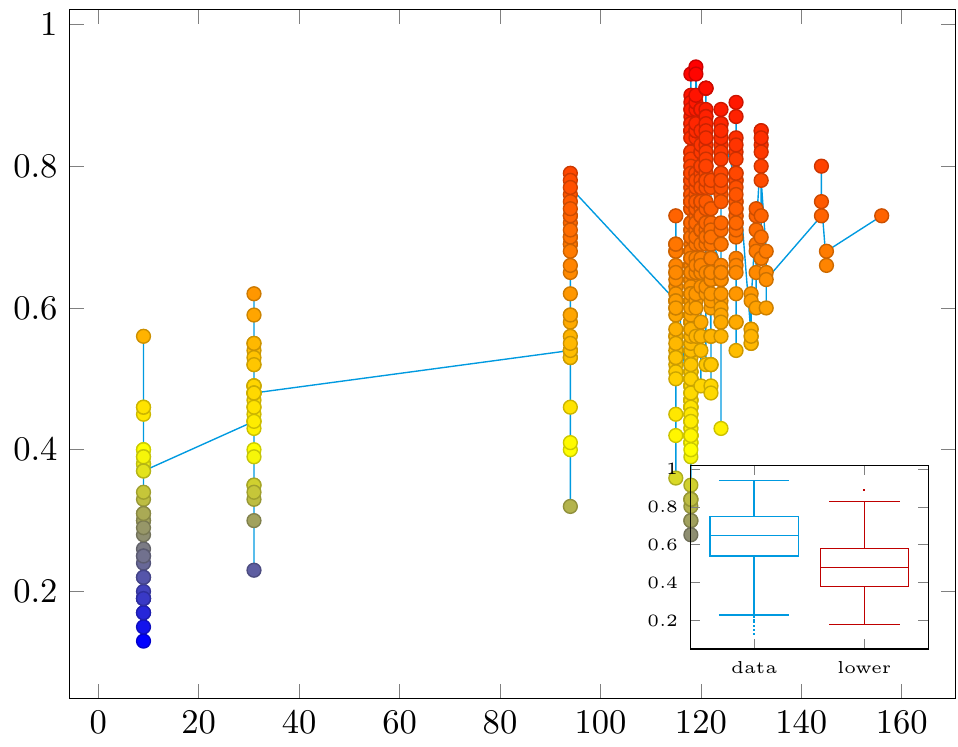} &
	\centering \includegraphics[width=0.9\linewidth]{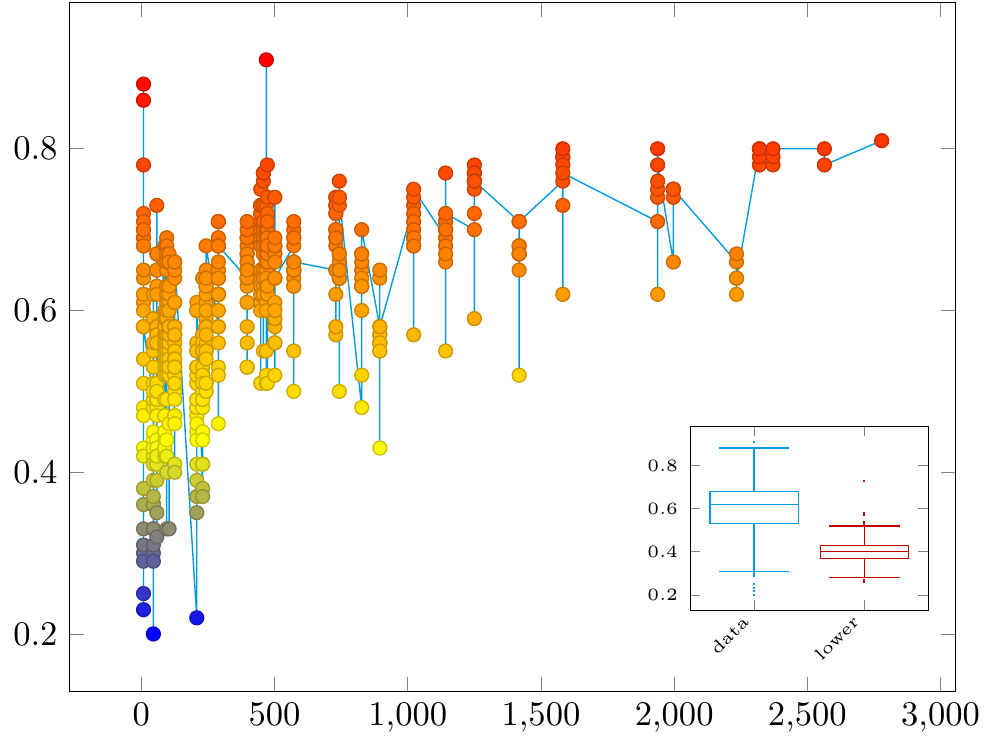} &
	\includegraphics[width=0.9\linewidth]{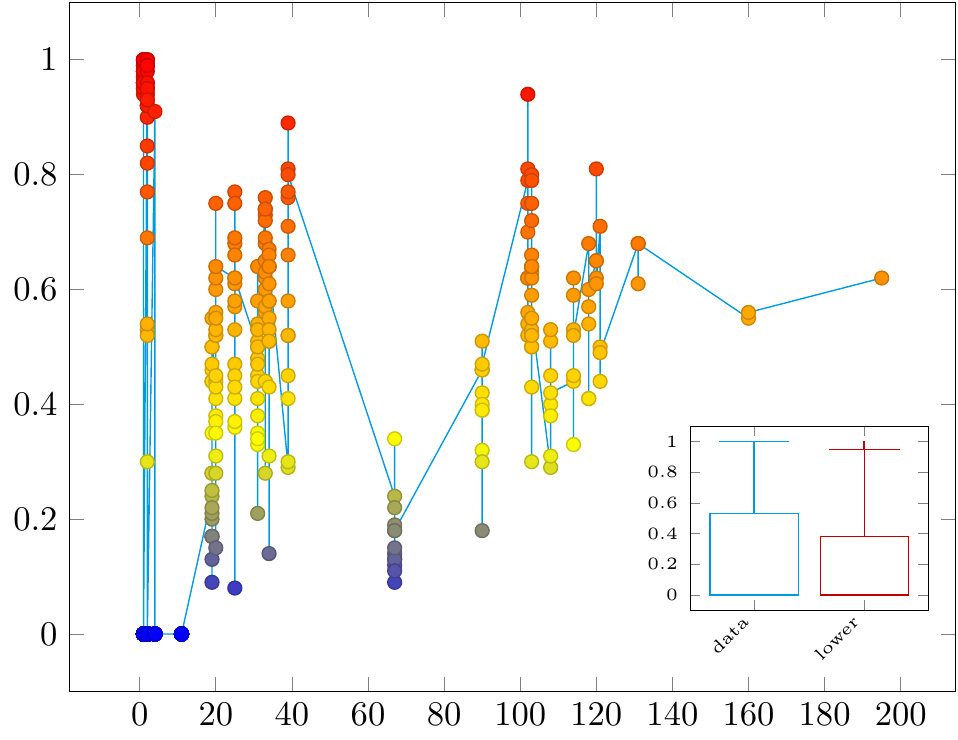}\\
	\hline \centering
	\rotatebox[origin=c]{90}{(4) GES} &
	\centering \includegraphics[width=.9\linewidth]{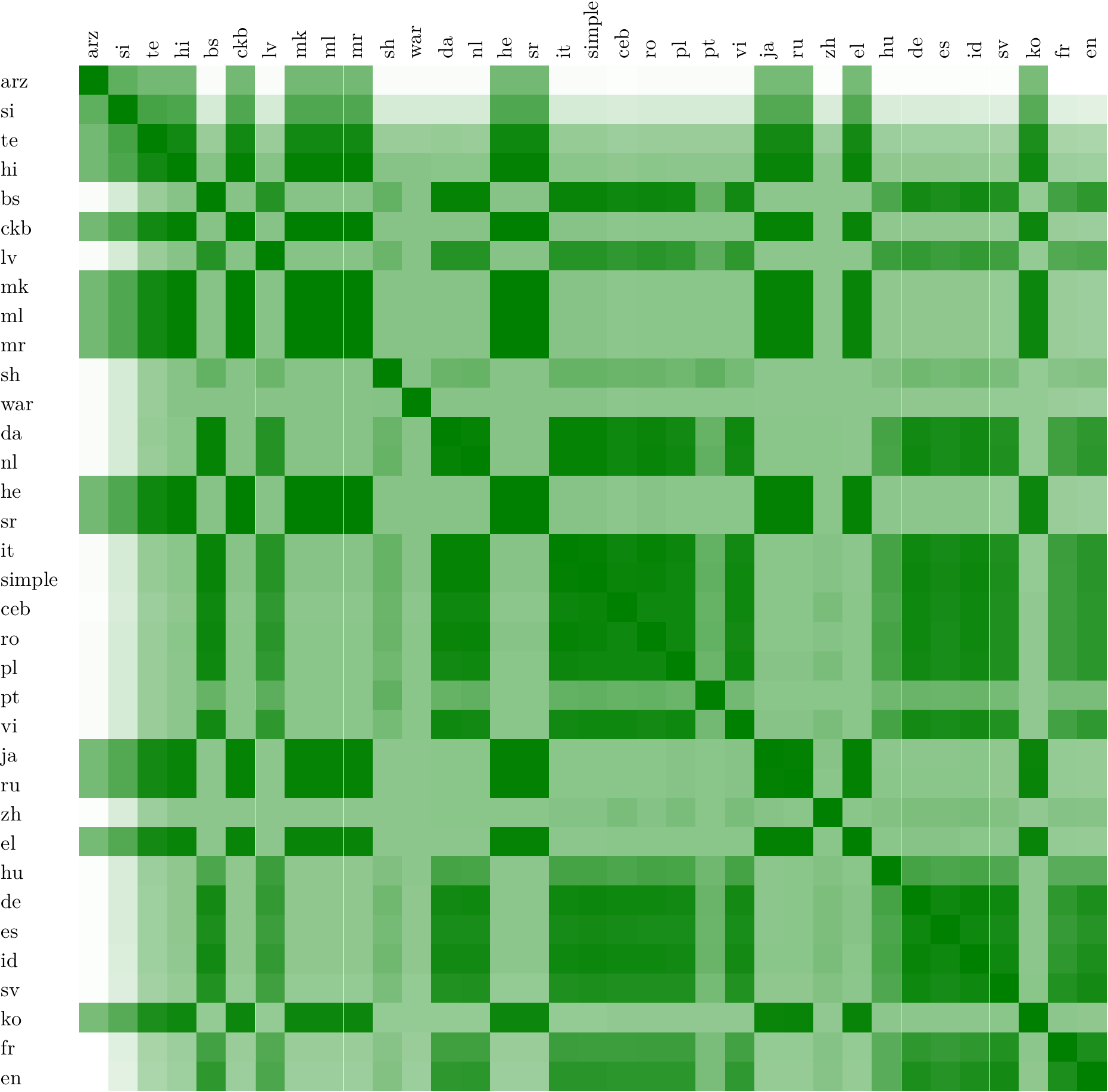}&
	\centering \includegraphics[width=.9\linewidth]{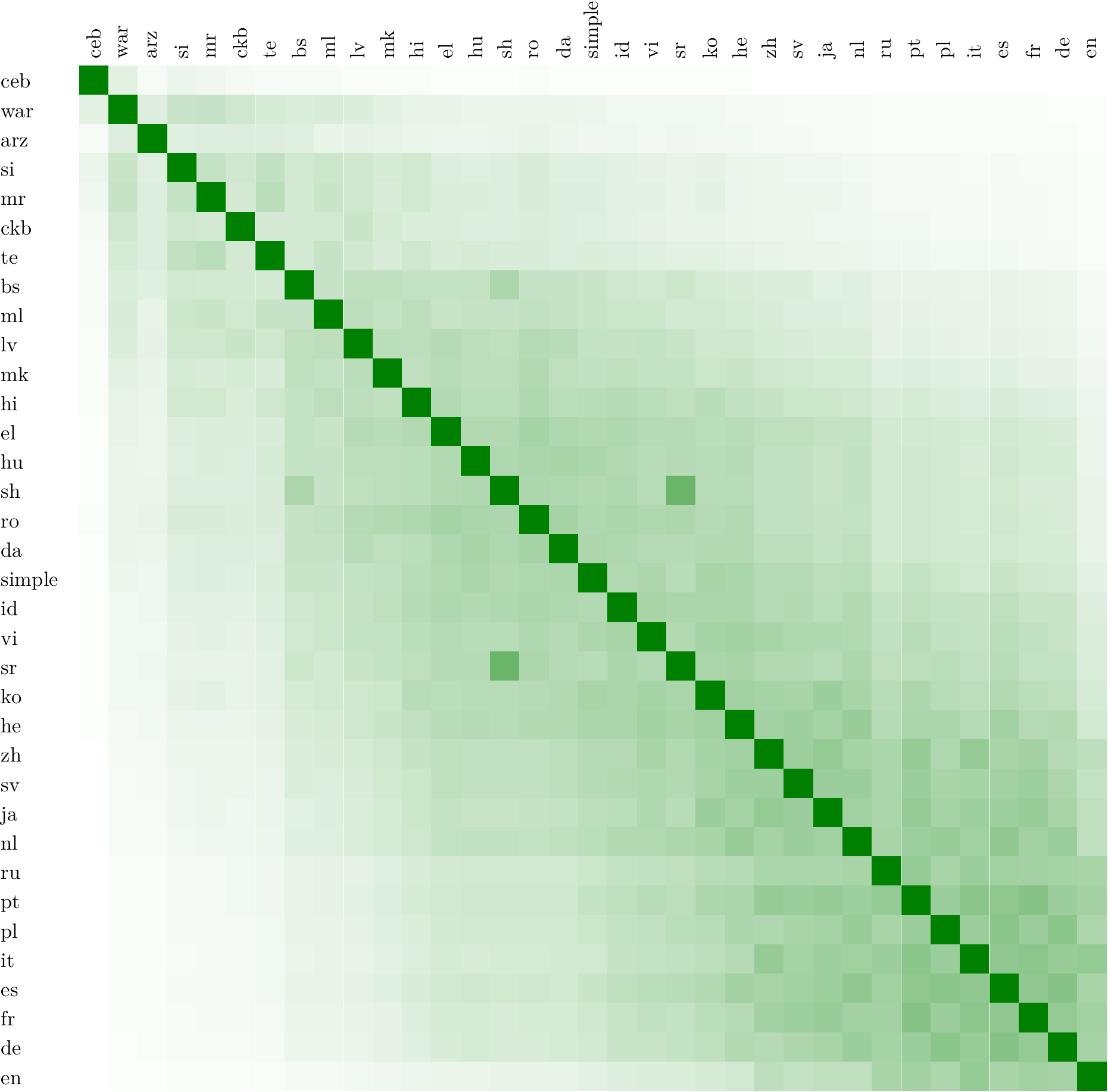} &
	\includegraphics[width=.9\linewidth]{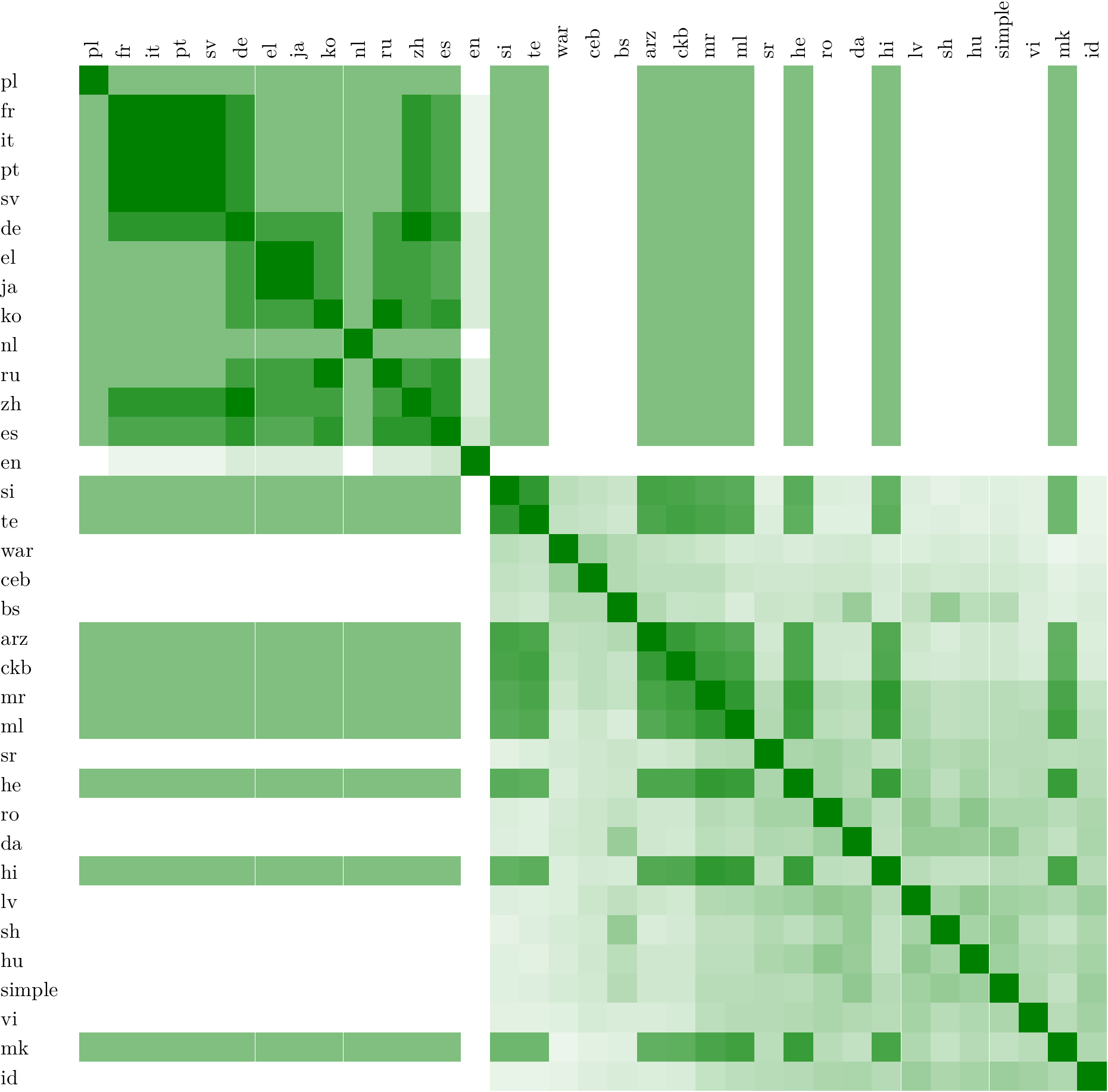}\\ 
	&
	\centering \includegraphics[width=0.9\linewidth]{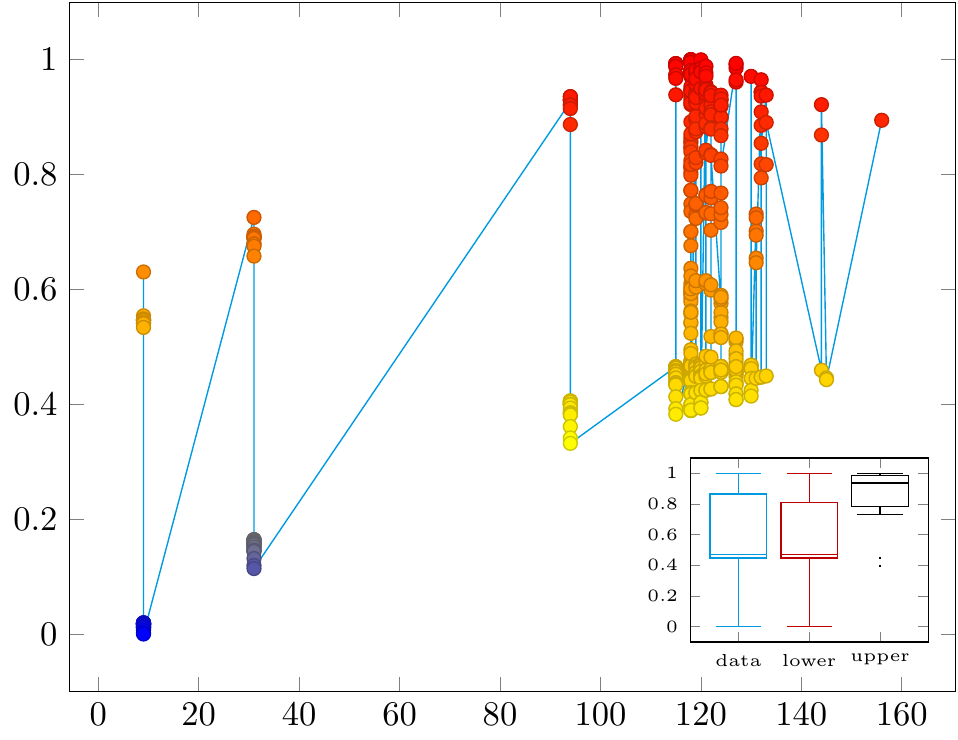}&
	\centering \includegraphics[width=0.9\linewidth]{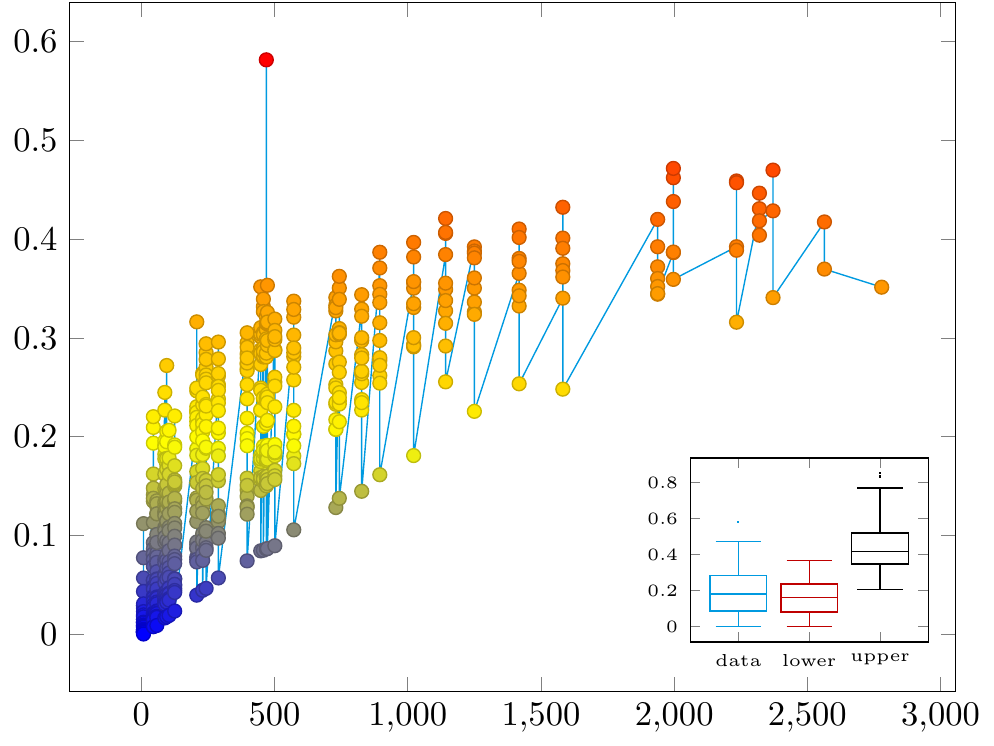} &
	\includegraphics[width=0.9\linewidth]{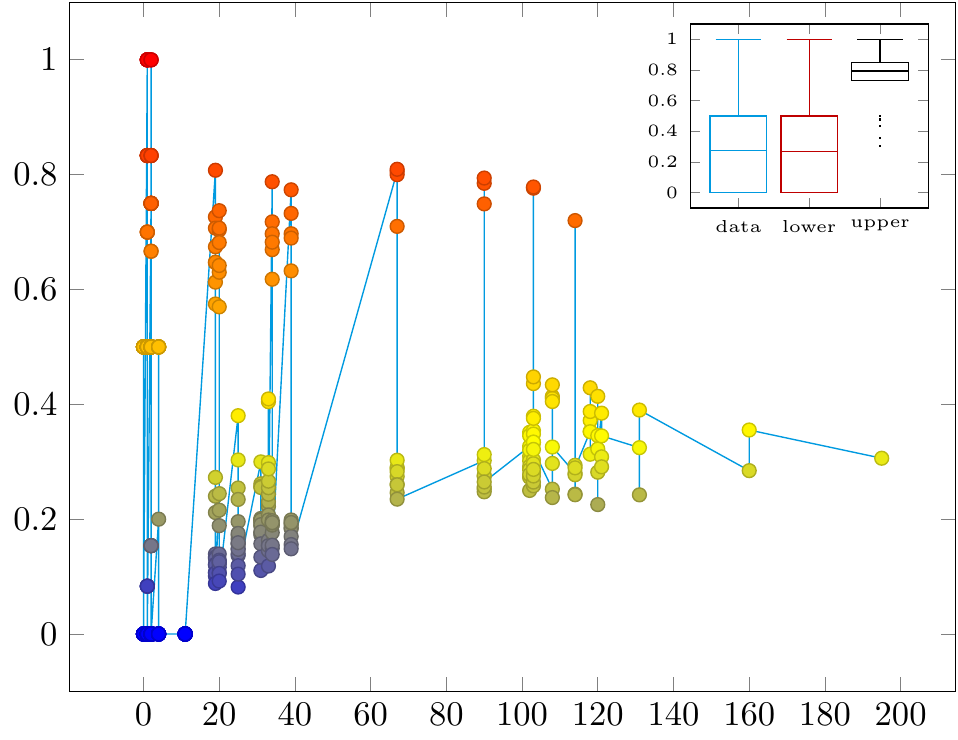}\\
	\hline \centering
	\rotatebox[origin=c]{90}{(5) DeltaCon} &
	\centering \includegraphics[width=.9\linewidth]{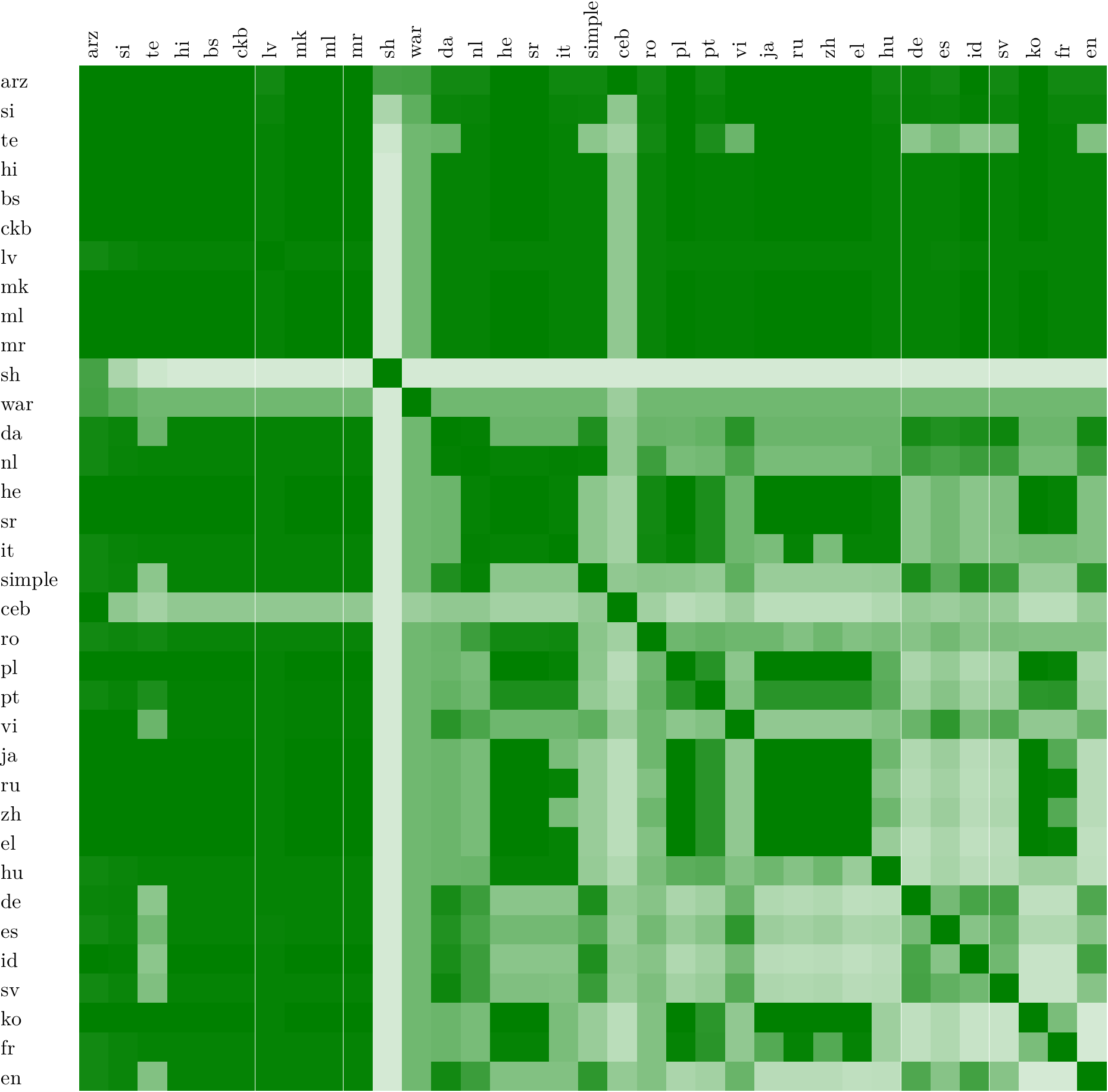}&
	\centering \includegraphics[width=.9\linewidth]{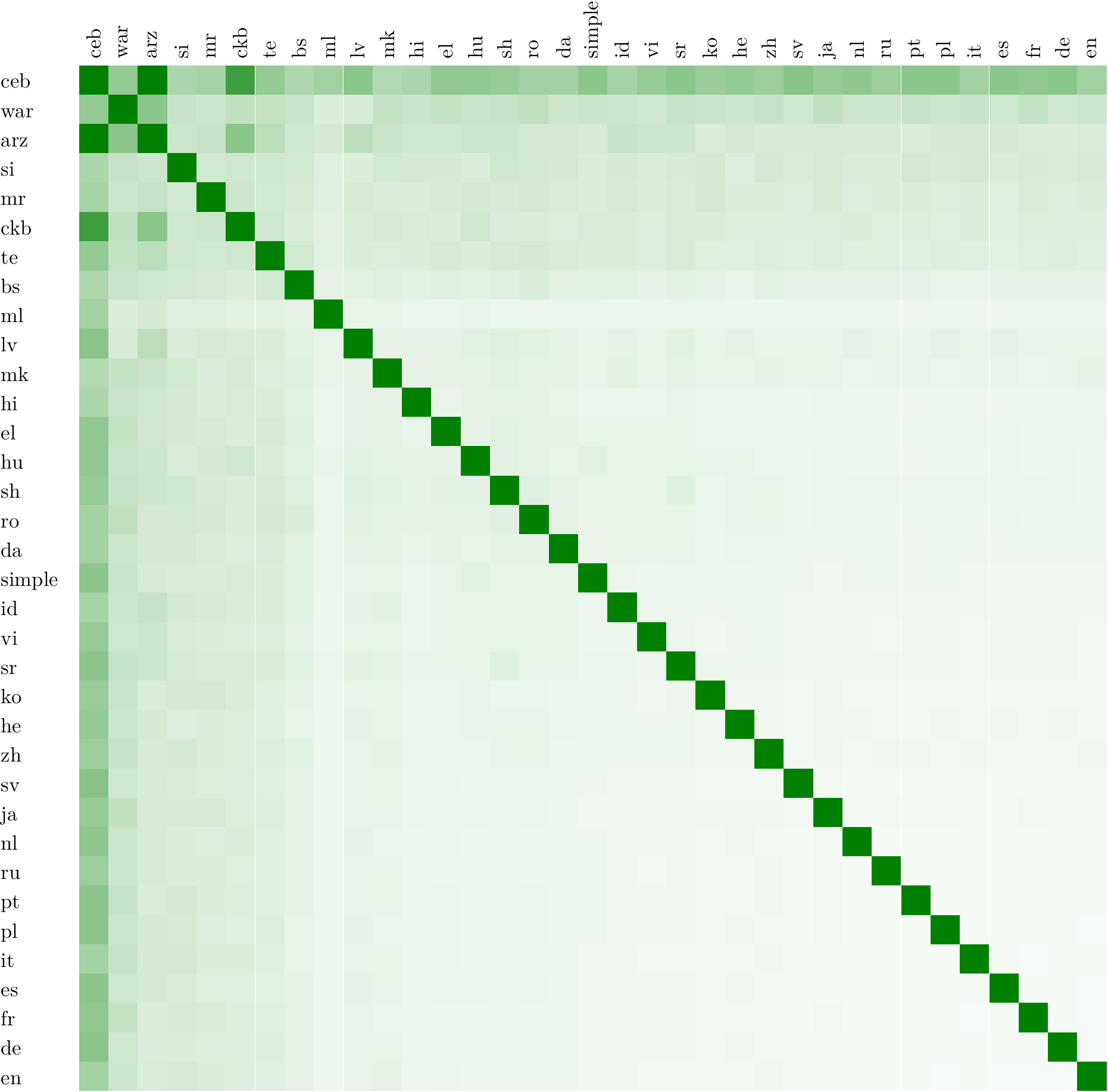} &
	\includegraphics[width=.9\linewidth]{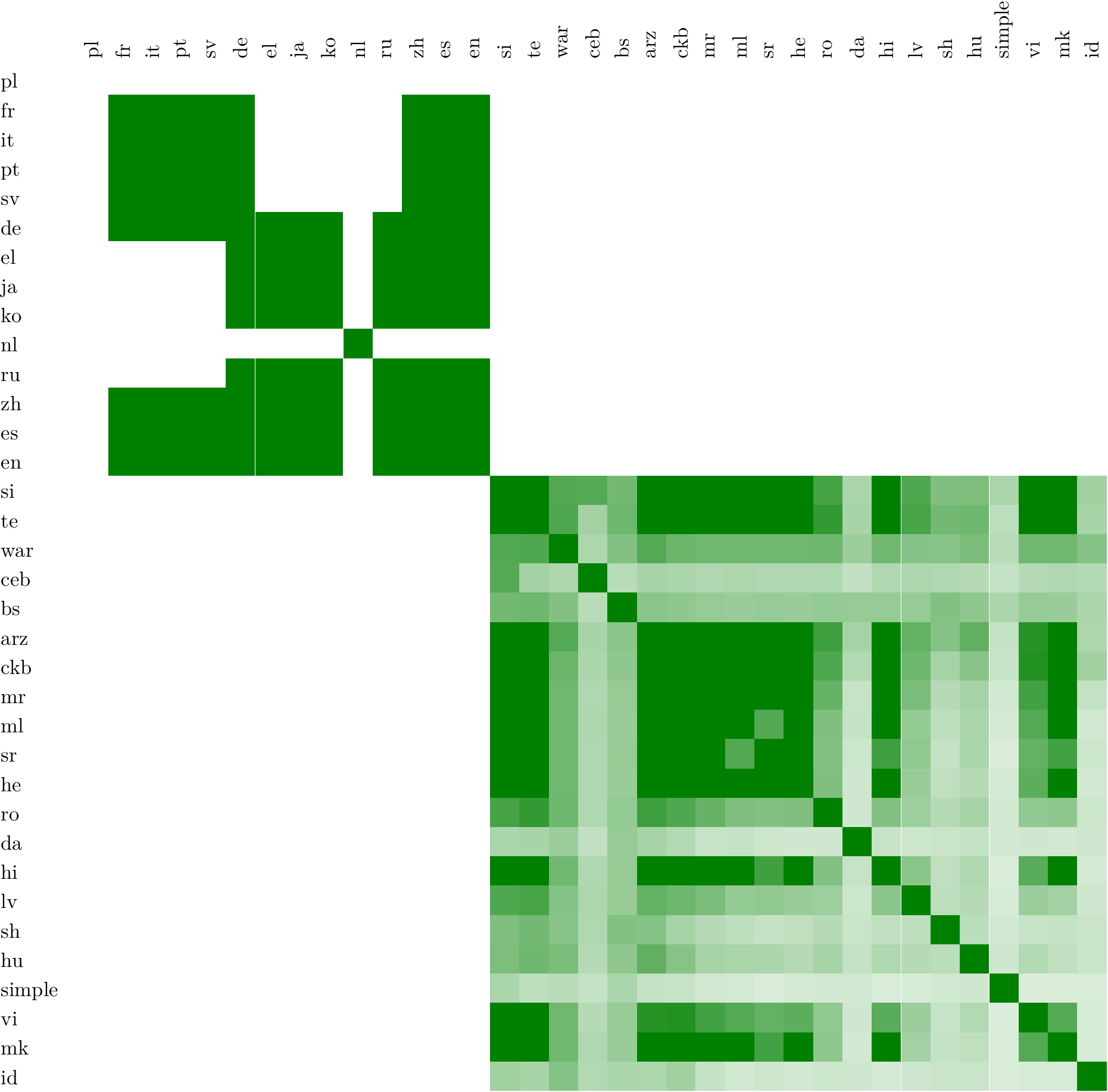}\\
	&
	\centering \includegraphics[width=0.9\linewidth]{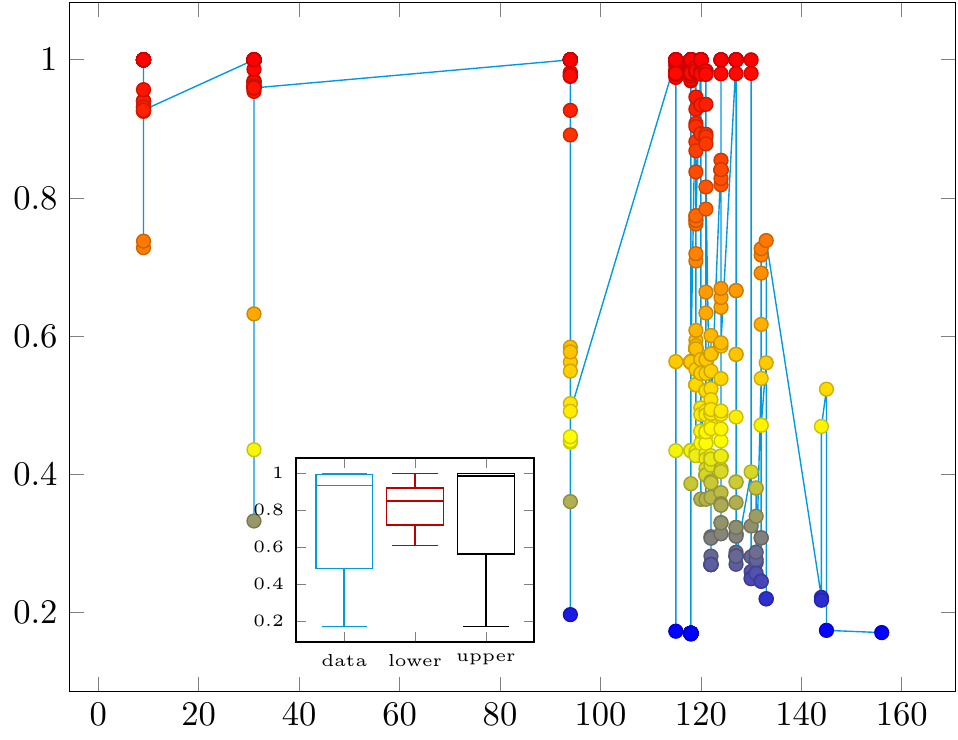}&
	\centering \includegraphics[width=0.9\linewidth]{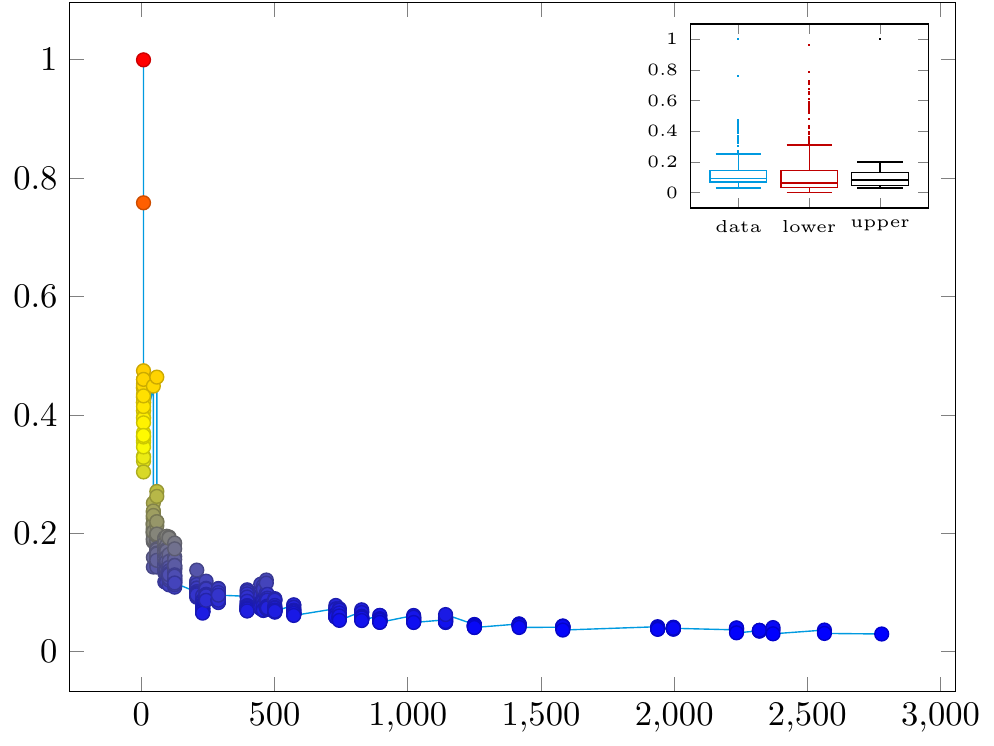} &
	\includegraphics[width=0.9\linewidth]{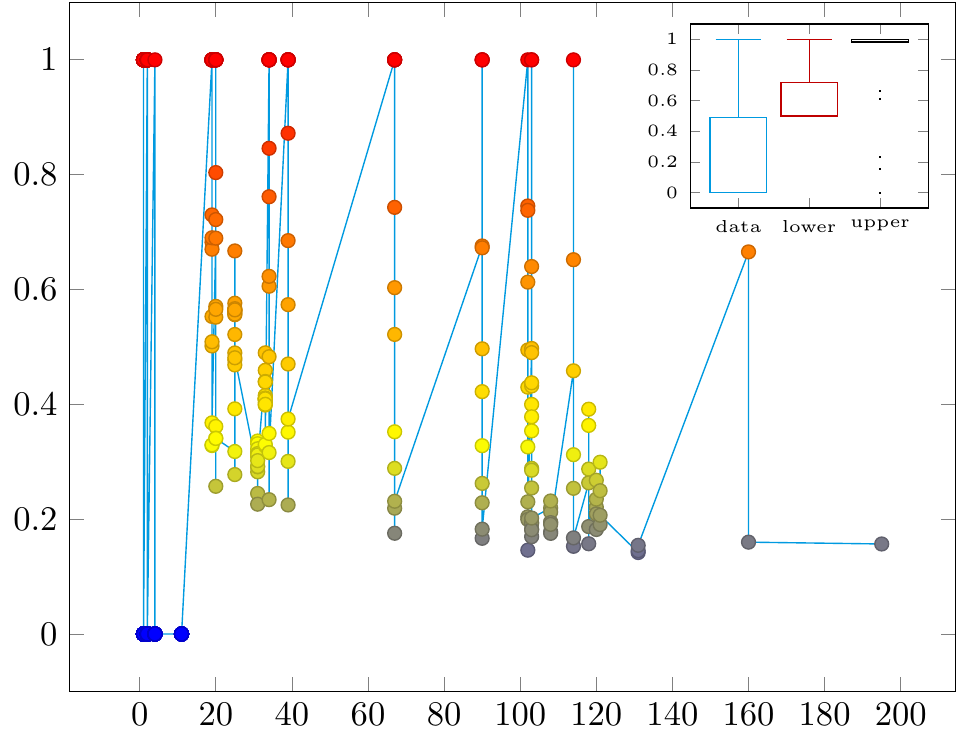}\\
	\hline 
	\rotatebox[origin=c]{90}{(6) CGS} &
	\centering \includegraphics[width=.9\linewidth]{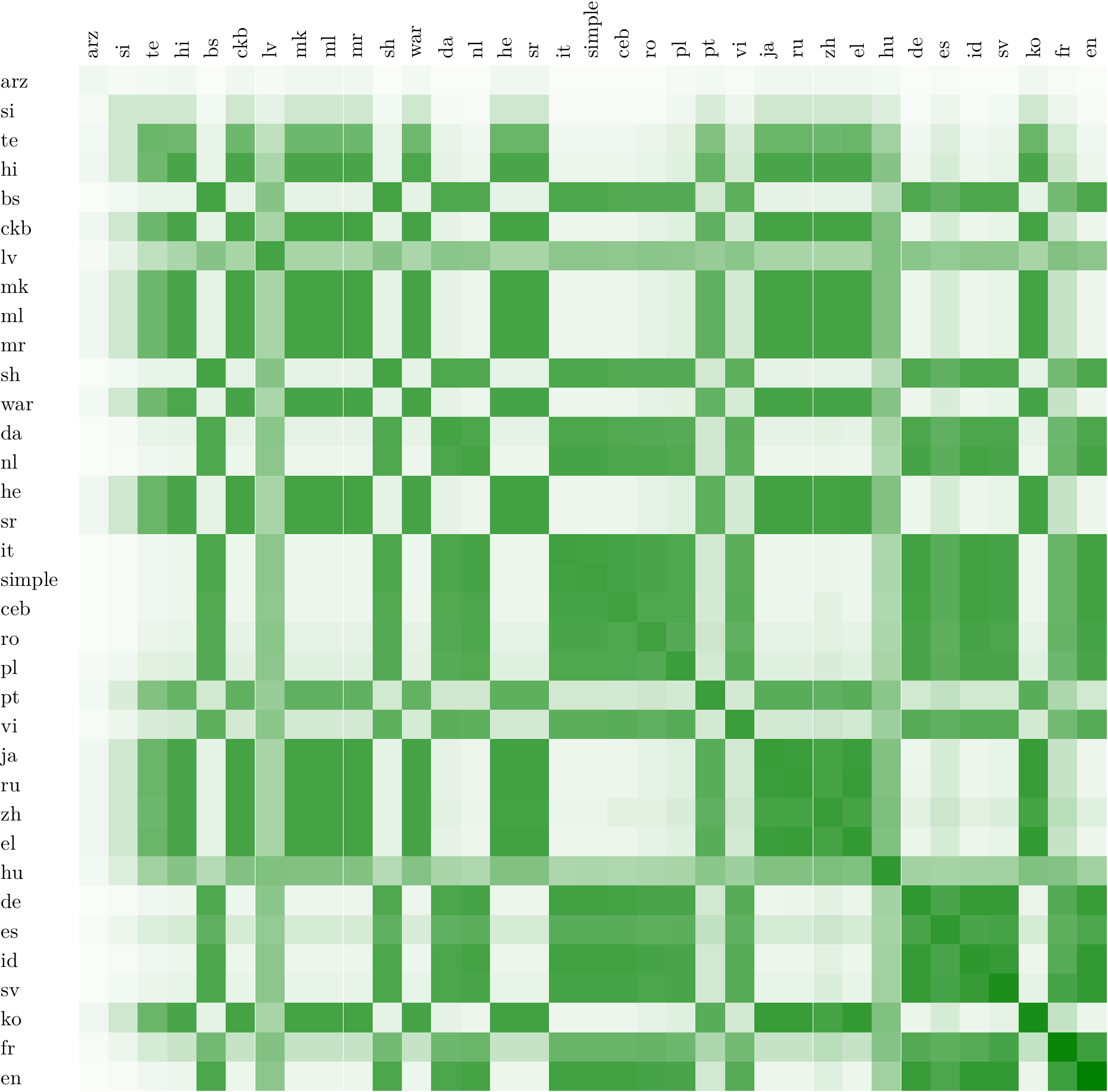}&
	\centering \includegraphics[width=.9\linewidth]{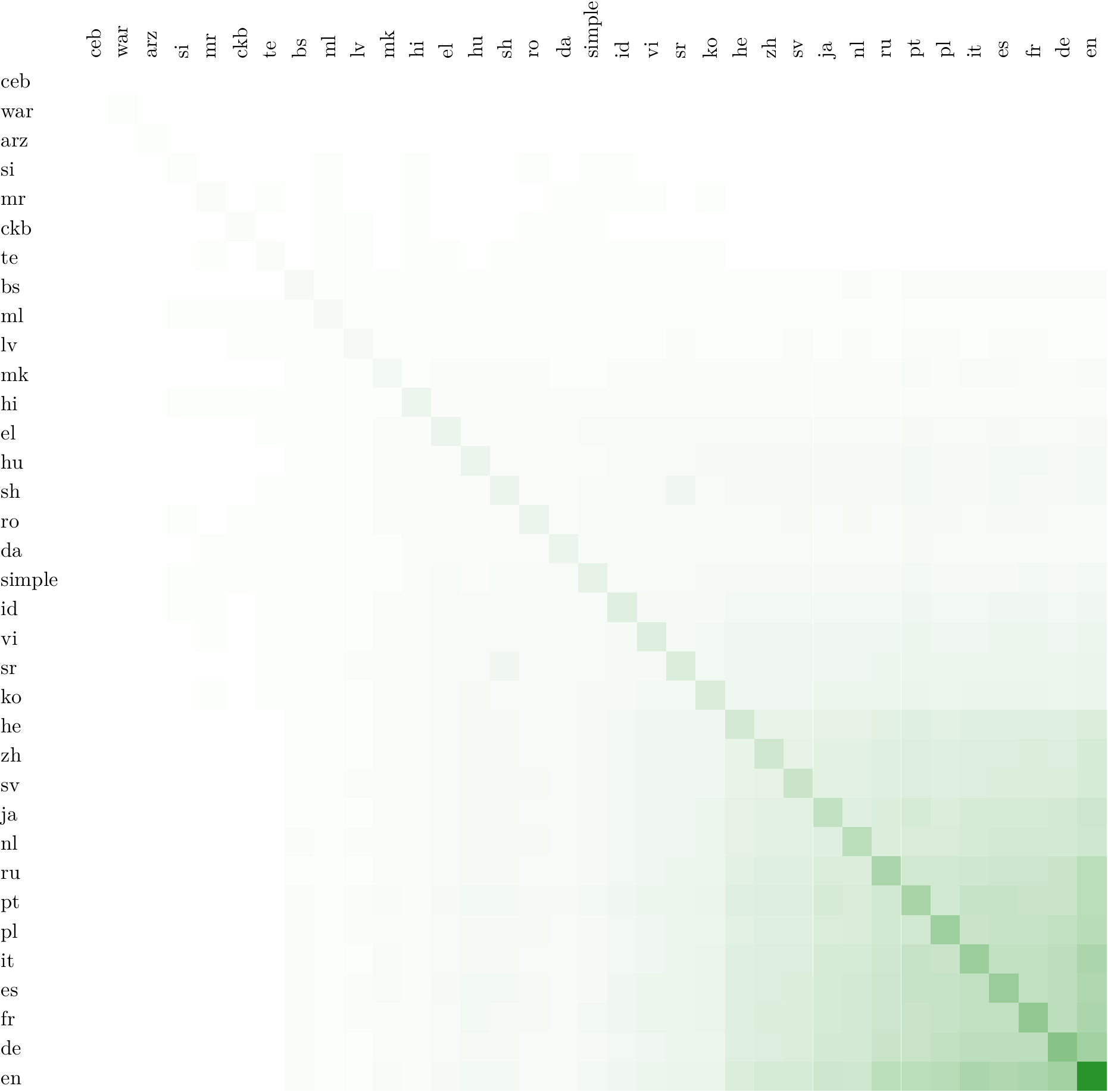} &
	\includegraphics[width=.9\linewidth]{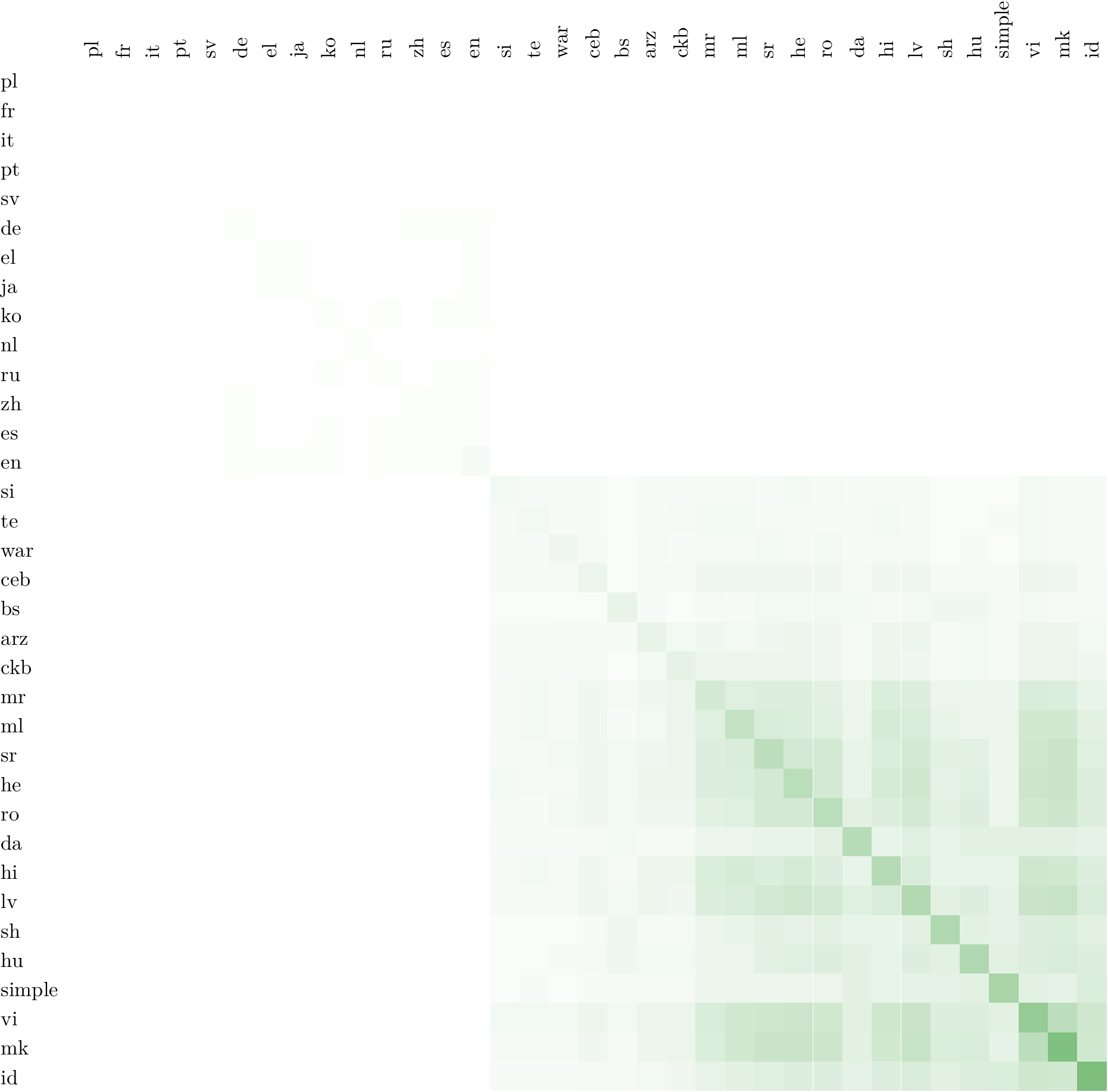}
	\\
	&
	\centering \includegraphics[width=0.9\linewidth]{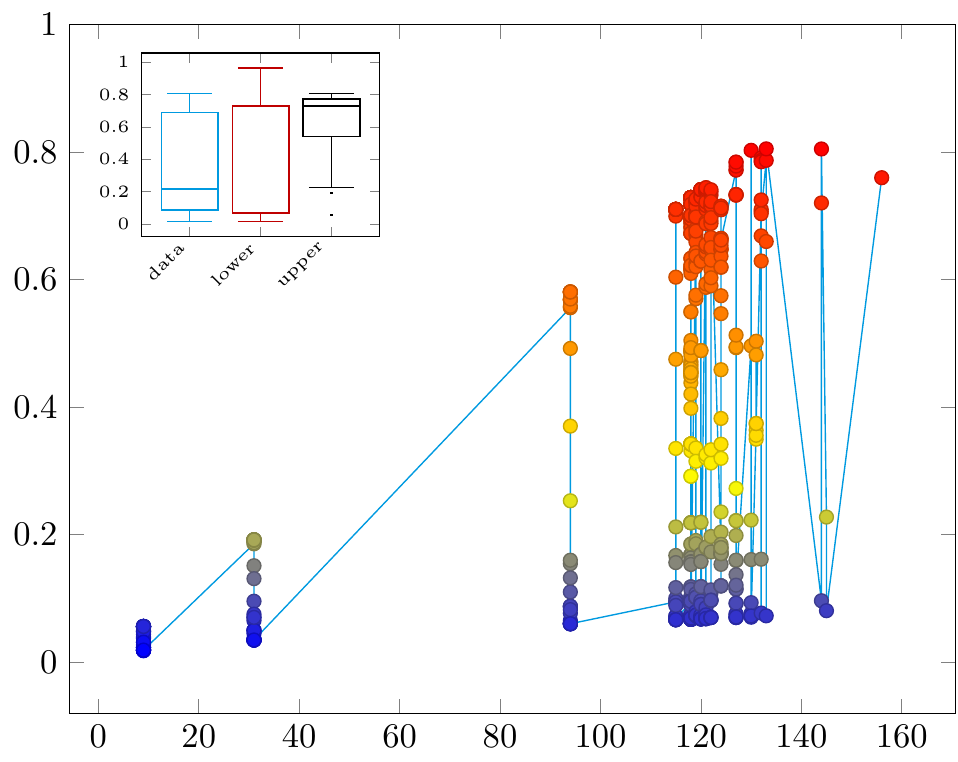}&
	\centering \includegraphics[width=0.9\linewidth]{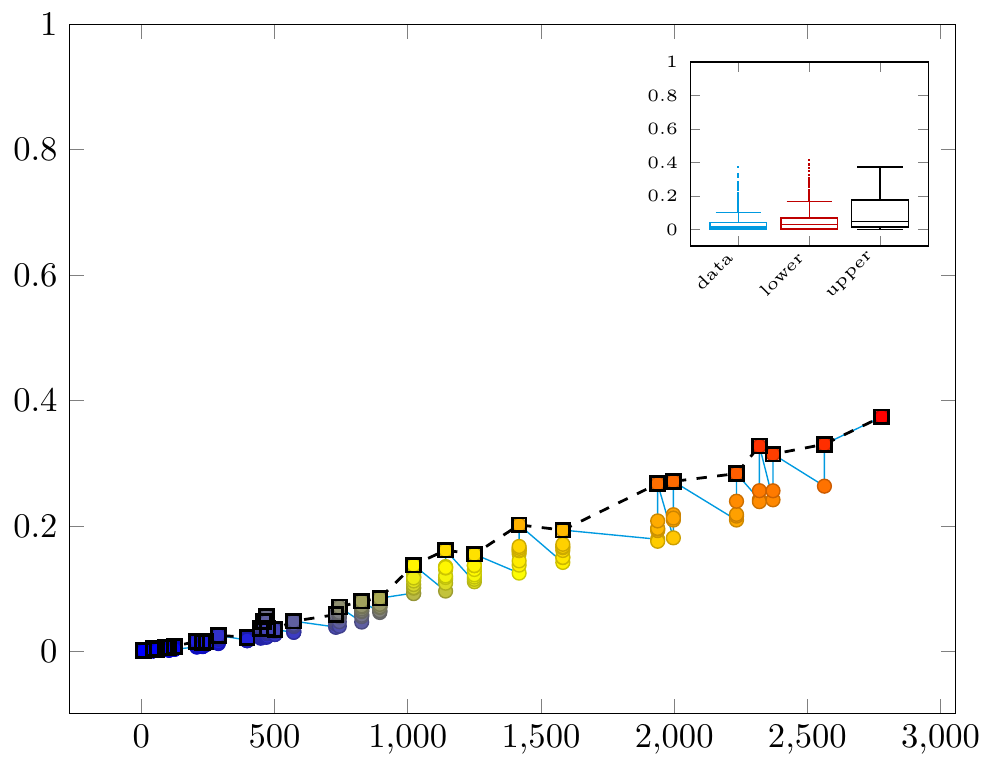}
	&
	\includegraphics[width=0.9\linewidth]{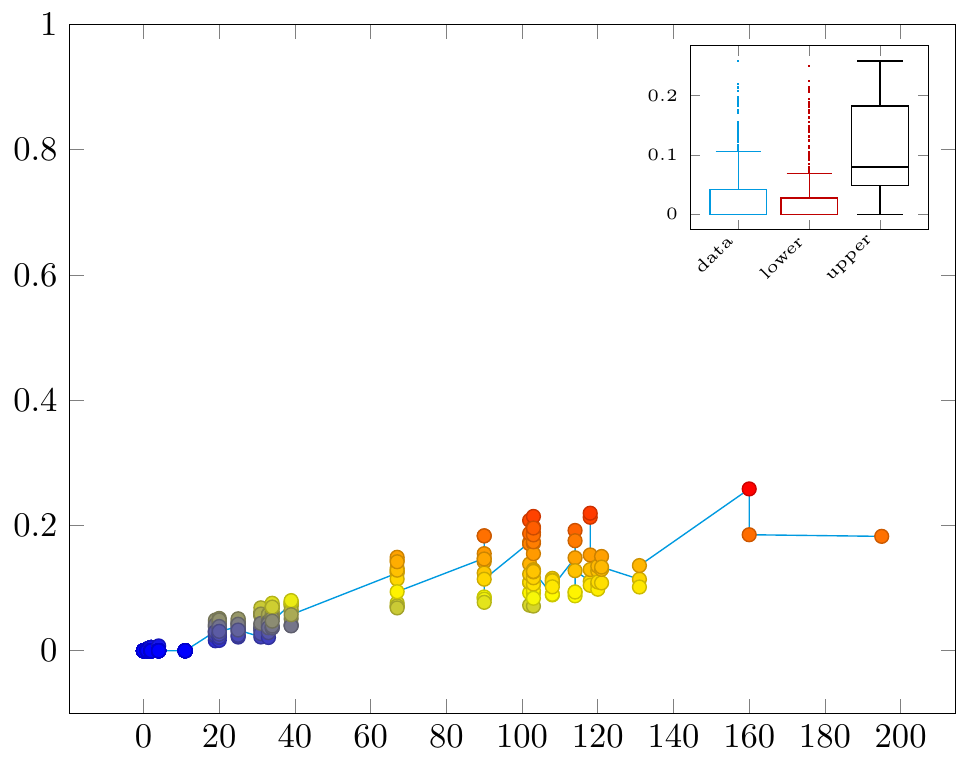}\\
	
	\hline
	\caption{Six measures (row) for computing the similarities of INs by example of three subject areas (column): rows and columns of the heatmaps correspond to languages (INs); curves below the heatmaps display similarity values as a function of the minimum order of the input graphs.}
	\label{tab:Longtable}
\end{longtable}

\begin{longtable}{*{3}{m{0.32\textwidth}}}
		\includegraphics[width=\linewidth]{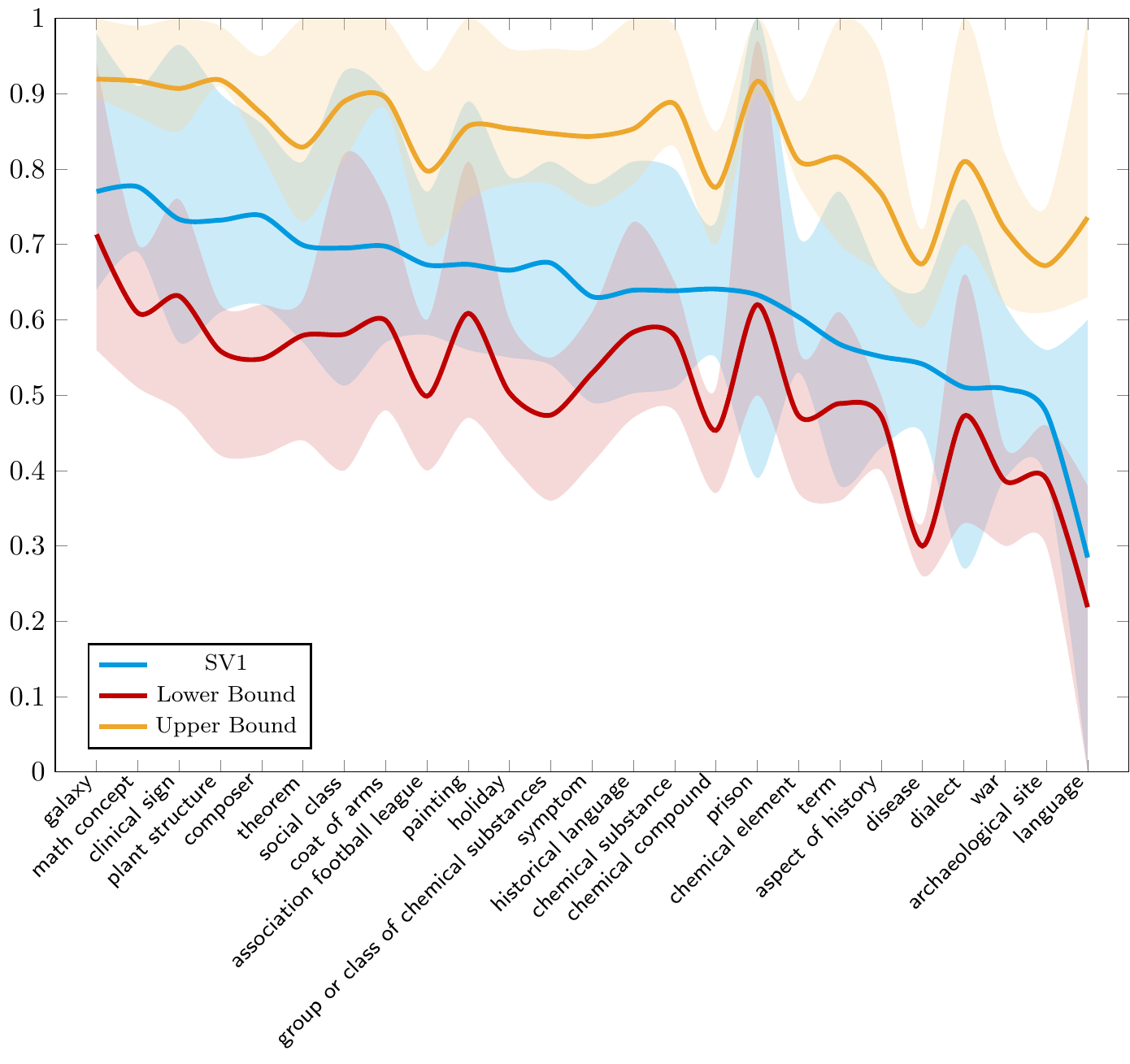} &
		\includegraphics[width=\linewidth]{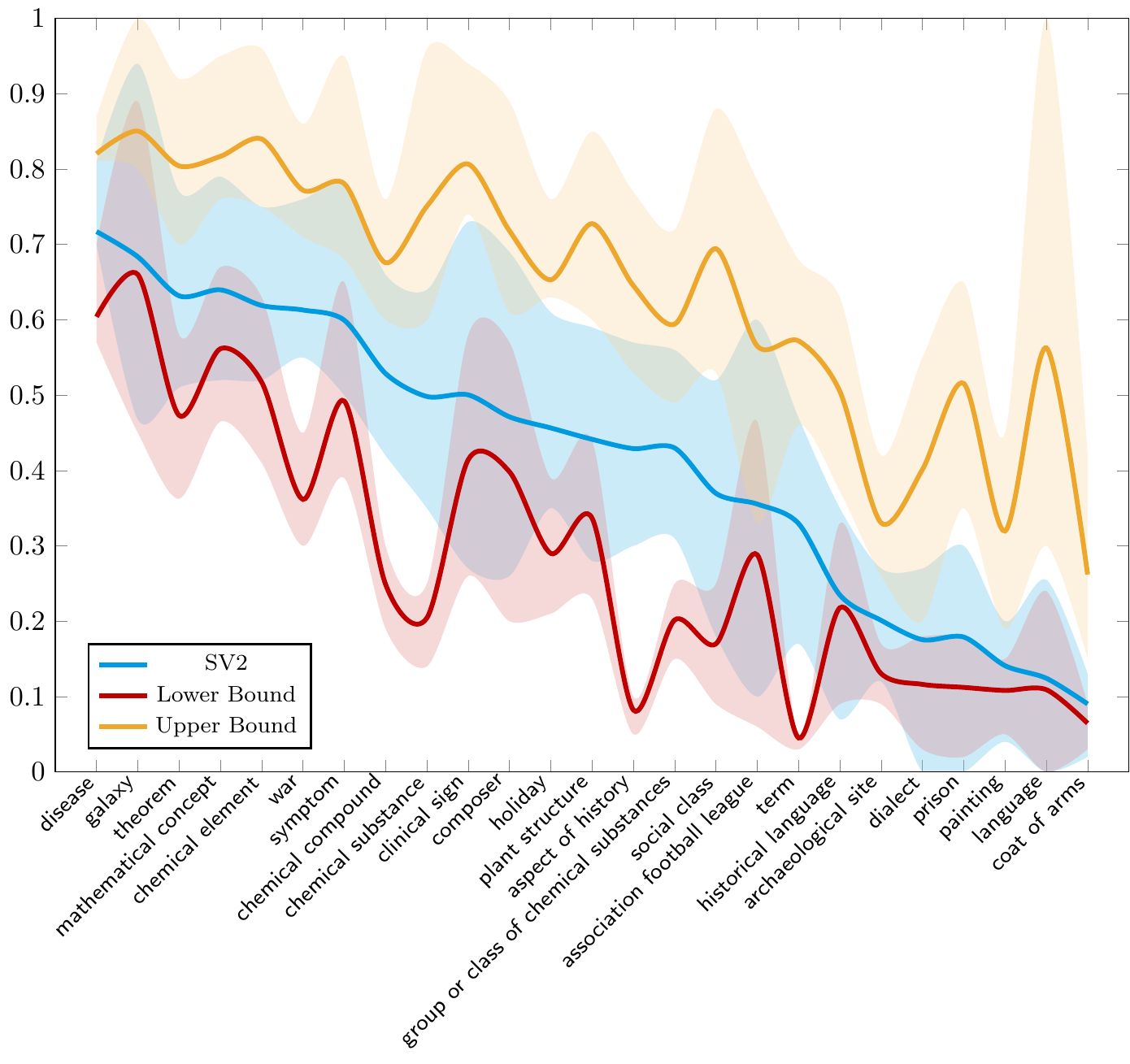} &
		\includegraphics[width=\linewidth]{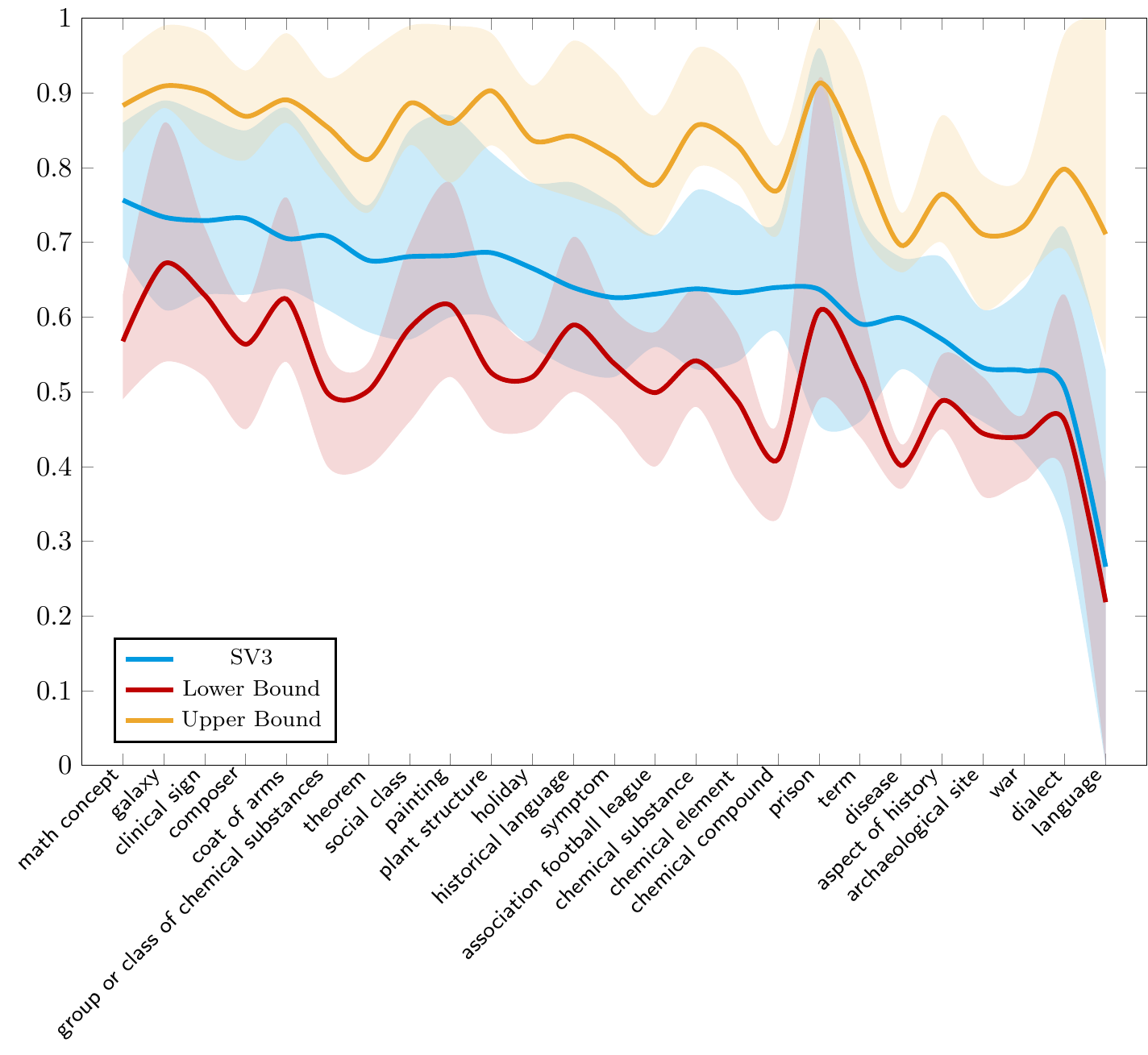} \\
		\includegraphics[width=\linewidth]{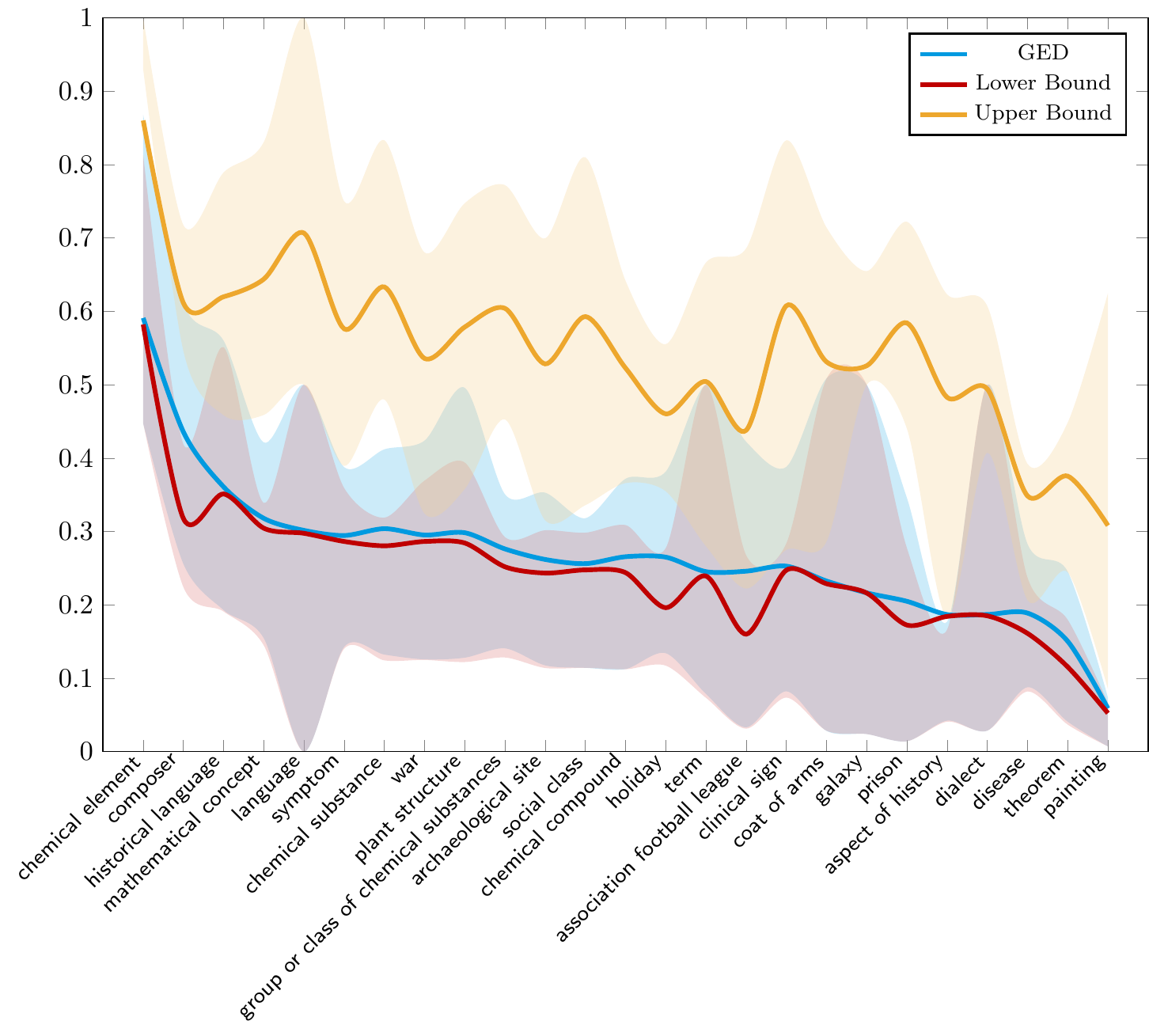} &
		\includegraphics[width=\linewidth]{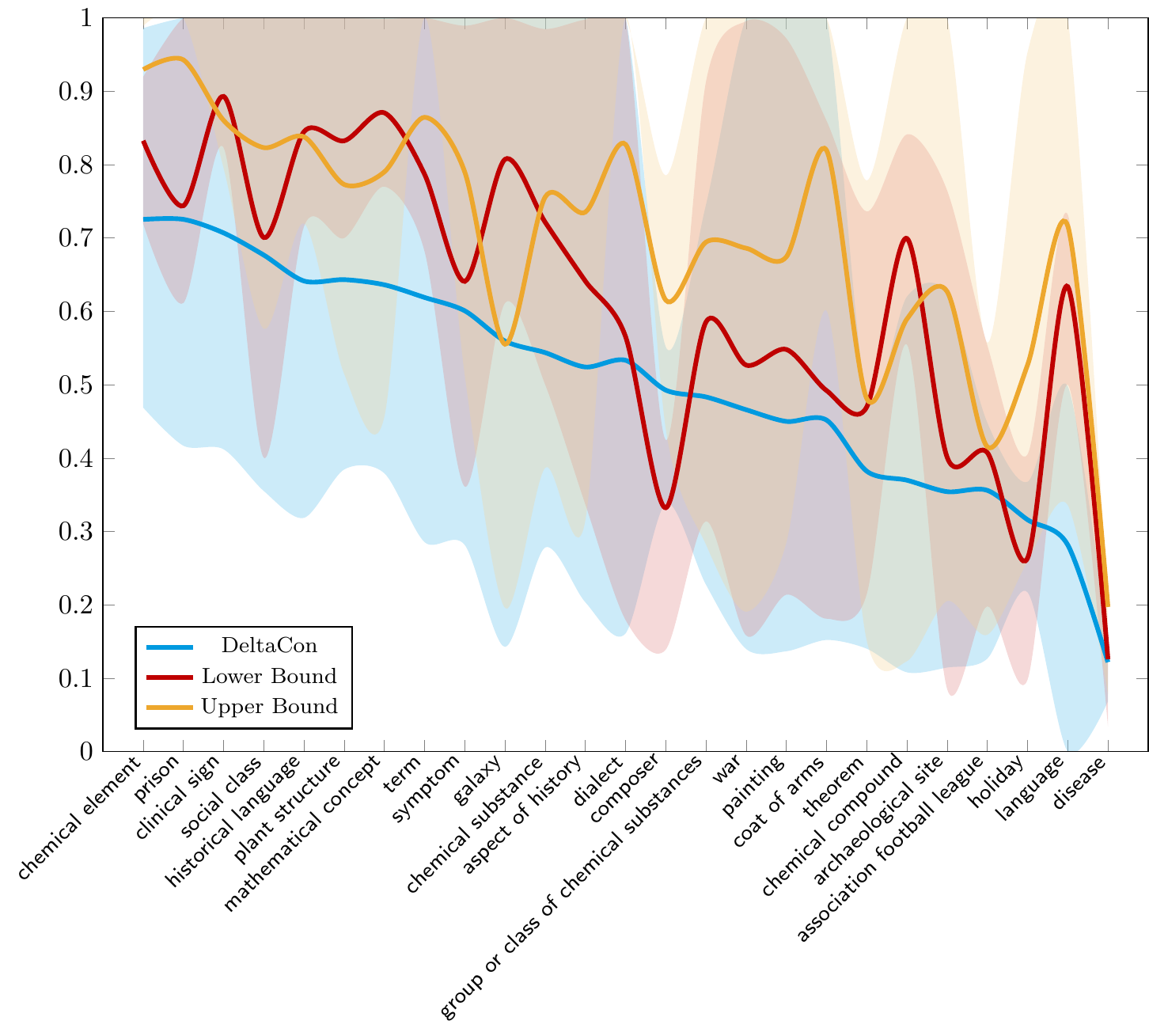} &
		\includegraphics[width=\linewidth]{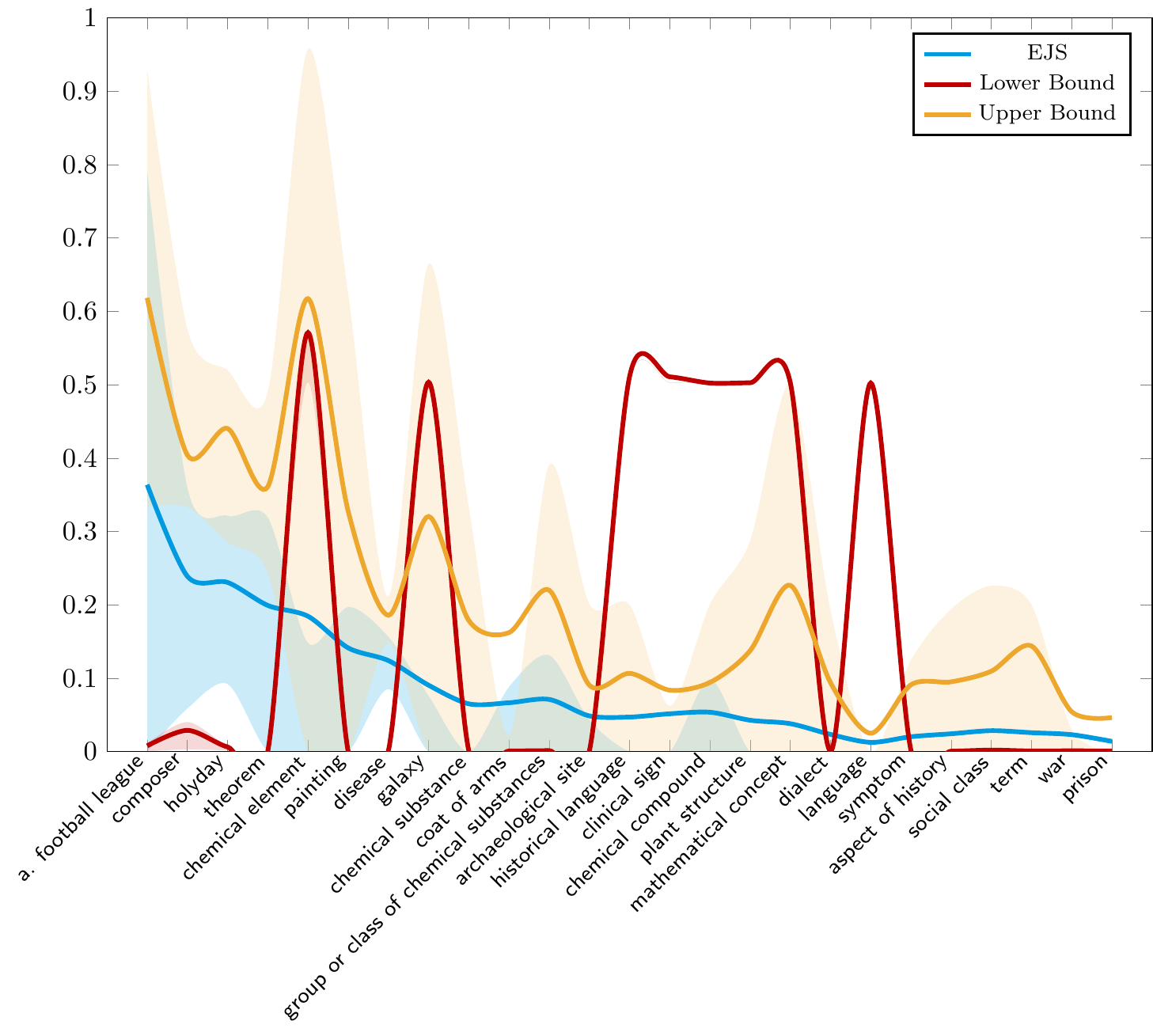} \\
		\includegraphics[width=\linewidth]{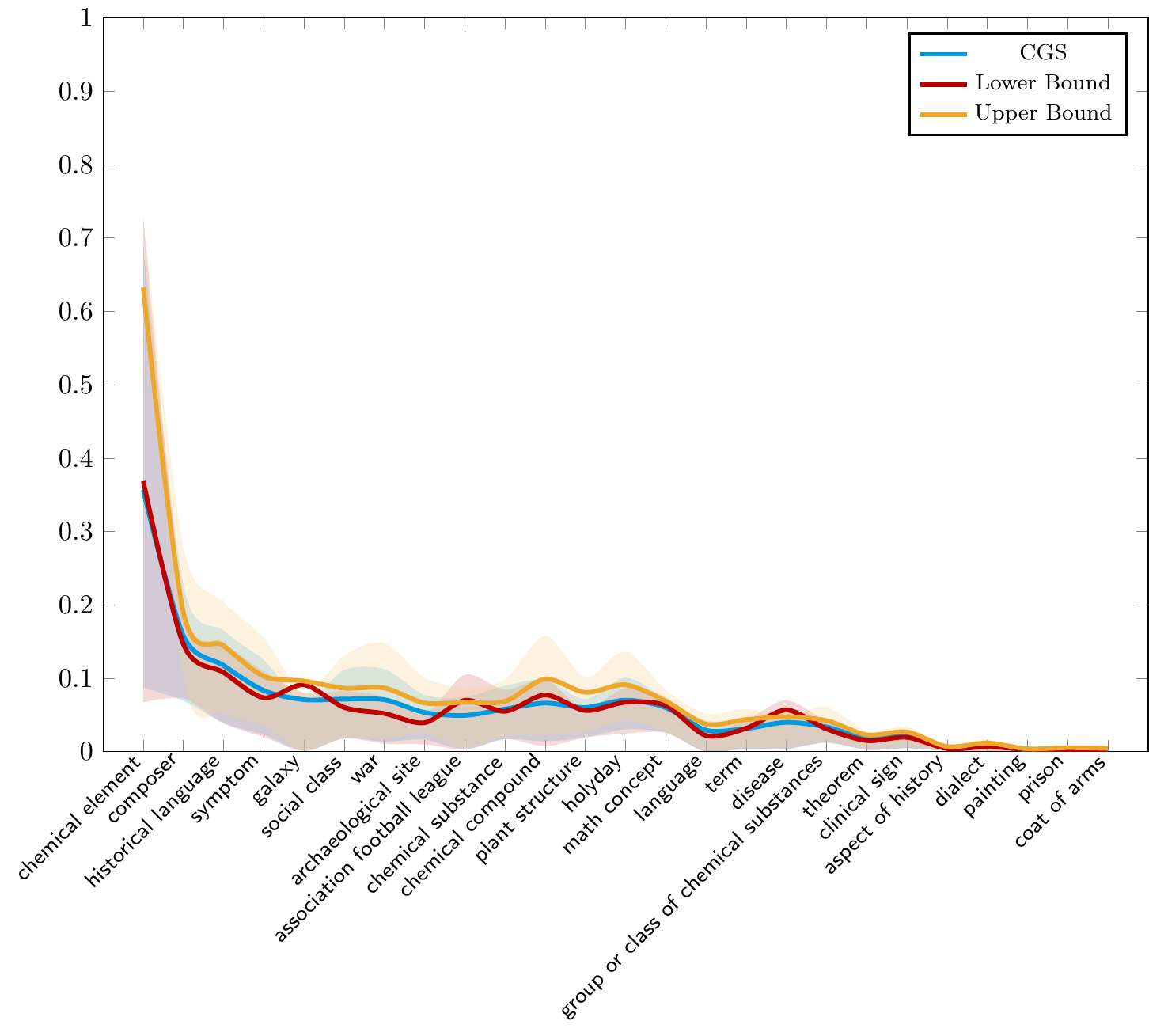} \\
		\caption{Distributions of similarity values of INs calculated by the similarity measures of Section \ref{sec:A Hybrid Approach to Measuring the Similarity of INs} and displayed per subject area. From left to right, top-down: SV1-3, GES, DeltaCon, EJS, CGS.}
		\label{fig:fuzzyGraphs}
\end{longtable}

We now turn to correlation analysis and ask about the dependence of our similarity analysis on the size of the graphs involved, based on the hypothesis that the size (and thus indirectly the degree of development or activity) of a subject area explains our results.
For this purpose, we calculate Spearman's rank correlation with respect to four data series, each of which is generated for two similarity views or 3 similarity measures (SV2, DeltaCon and GES).
\begin{figure}
    \centering
    \includegraphics[width=\linewidth]{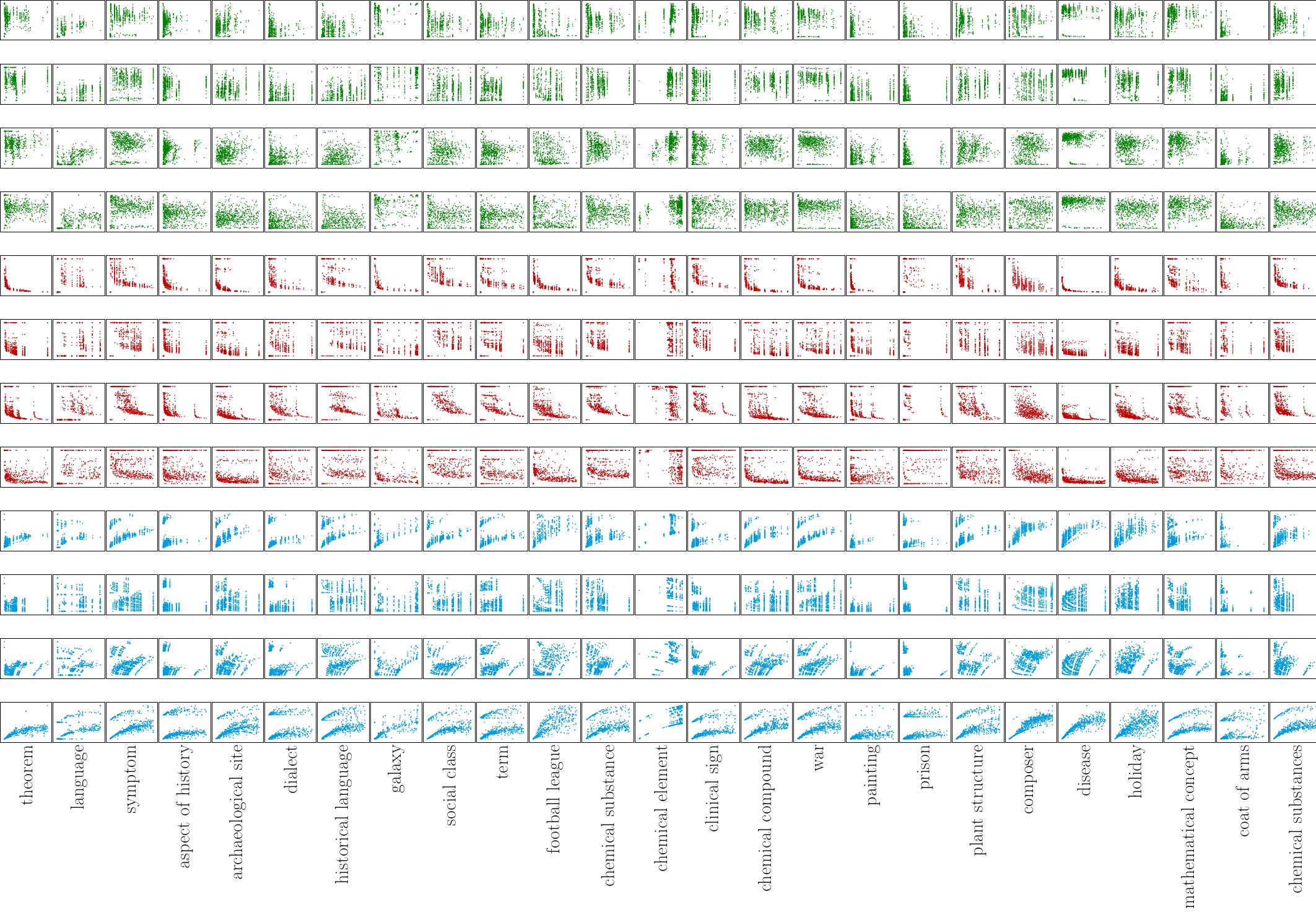}
    \caption{Similarities of languages as a function of the size (number of vertices) of the networks (INs) involved. Four alternatives of calculating the size $s(I_i^x, I_j^x)$ of pairs $(I_i^x, I_j^x)$ of network: 
    1st row: $s(I_i^x, I_j^x) = \min(|V_i^x|, |V_j^x|)$, 
    2nd row: $s(I_i^x, I_j^x) = \max(|V_i^x|, |V_j^x|)$, 
    3rd row: $s(I_i^x, I_j^x) = |V_i^x| + |V_j^x|$ and 
    4th row: $s(I_i^x, I_j^x) = \min(|V_i^x|, |V_j^x|) / \max(|V_i^x|, |V_j^x|)$.
    \textcolor{SeminarGruen}{Green:} similarity view SV2 (thematic intratextual similarity),
    \textcolor{SeminarRot}{red:} similarity view SV4-DeltaCon (intertextual similarity),
    \textcolor{SeminarBlau}{blue:} SV4-GES.
    }
    \label{fig:Correlation:1}
\end{figure}
The data series contrast measured similarities with the orders $|V_i^x|, |V_j^x|$ of the networks $I_i^x, I_j^x$ involved. 
\begin{figure}
    \includegraphics[width=0.49\linewidth]{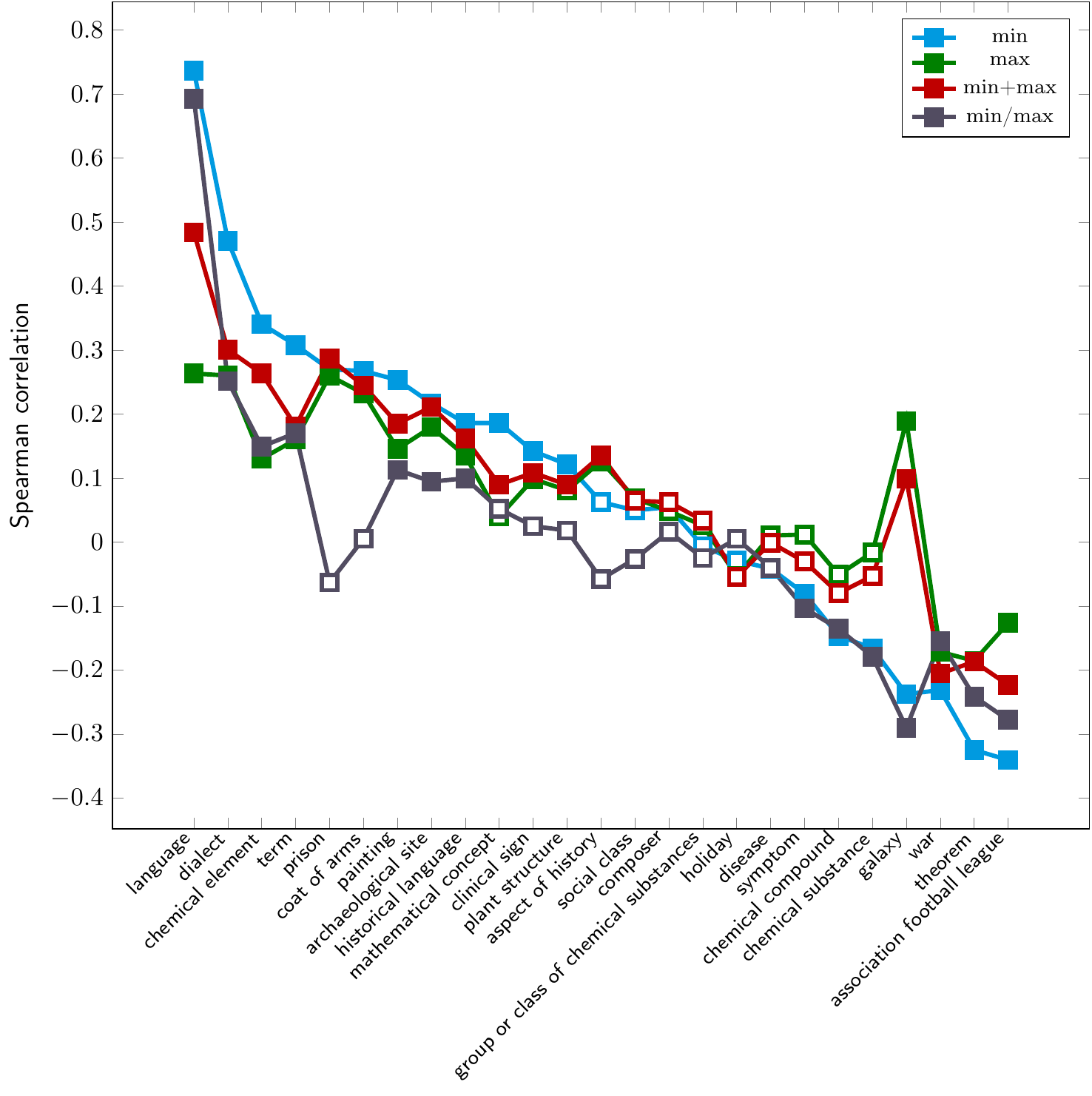}
    \includegraphics[width=0.49\linewidth]{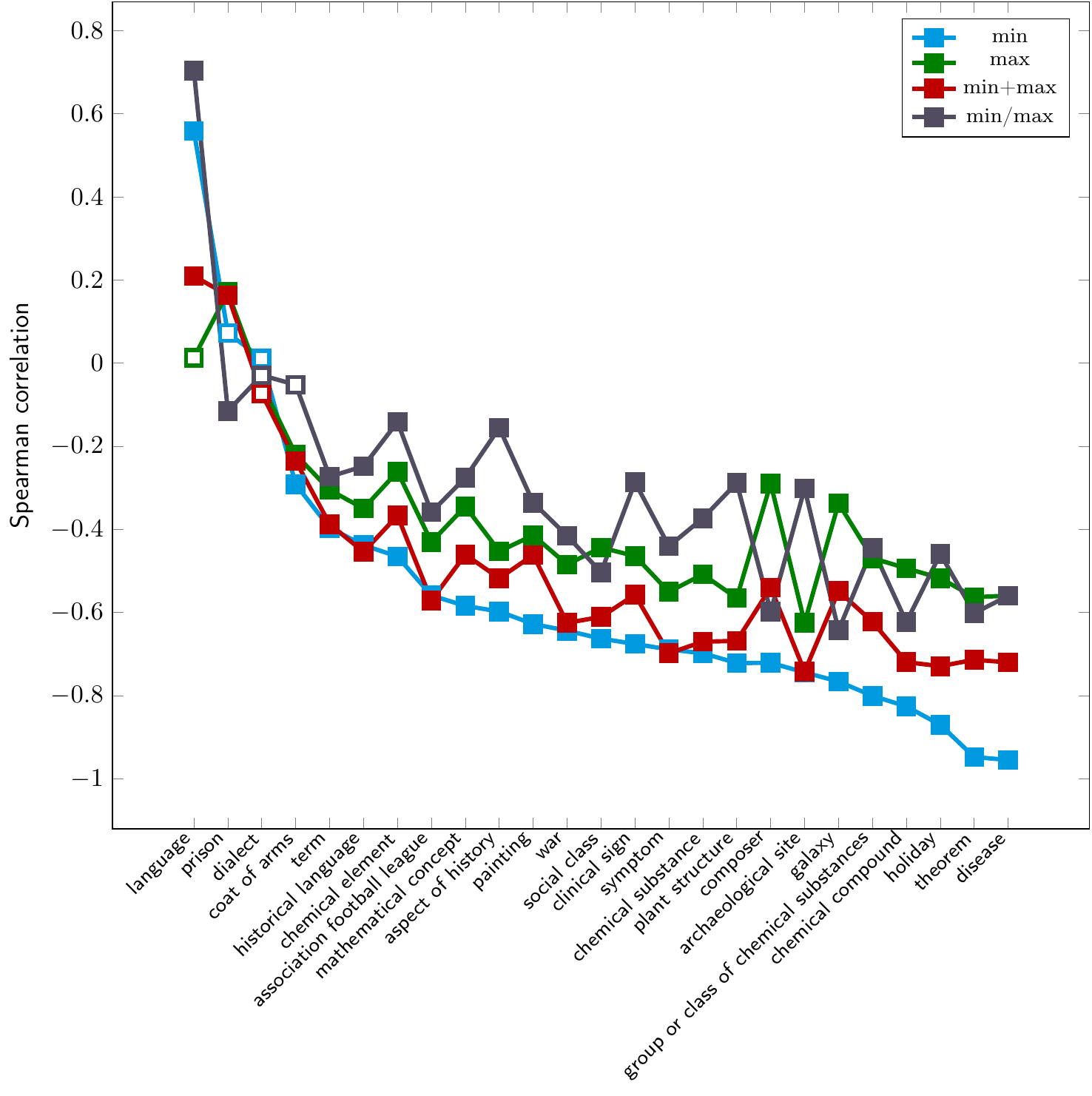}
    \includegraphics[width=0.49\linewidth]{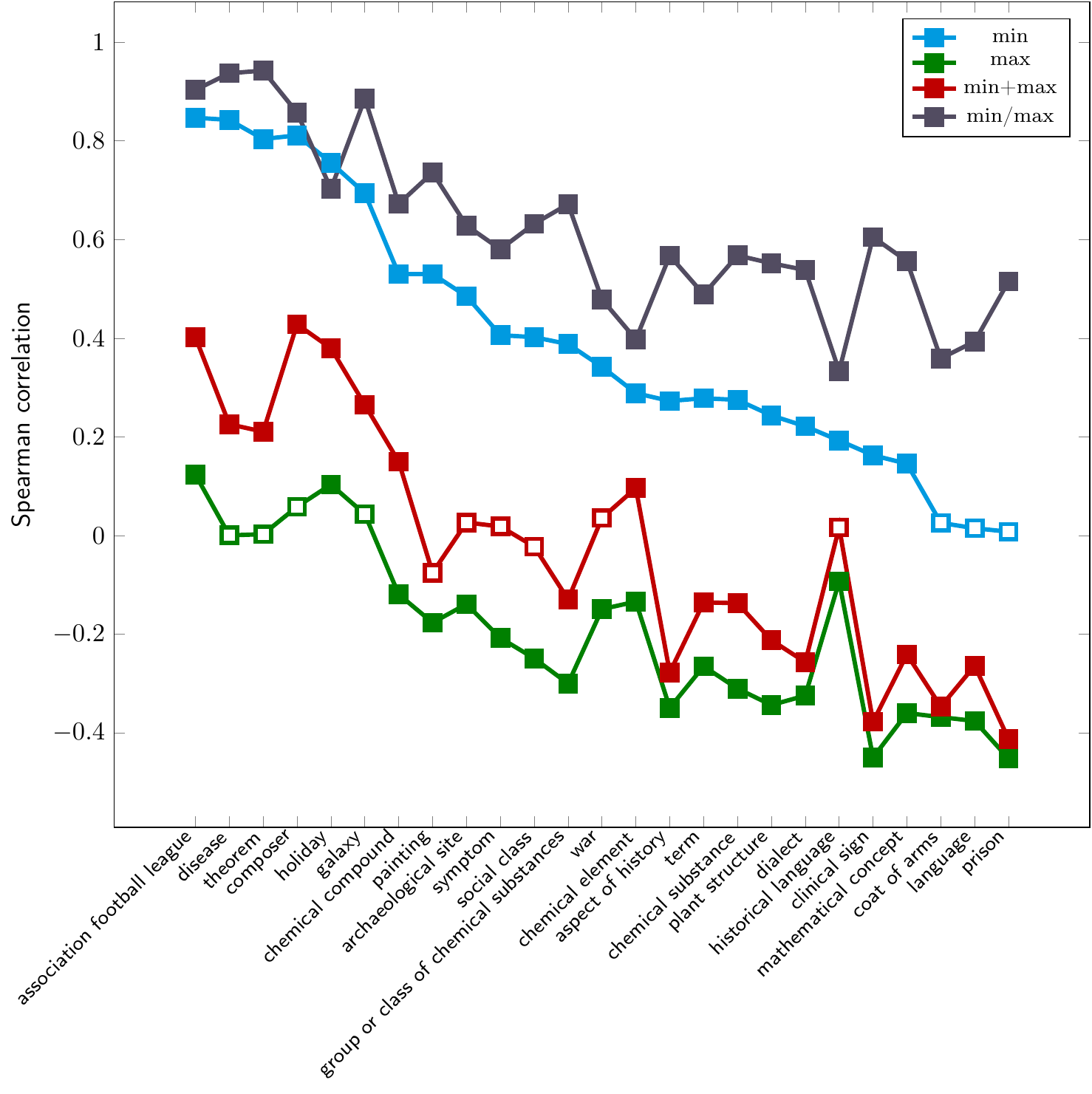}
    \includegraphics[width=0.49\linewidth]{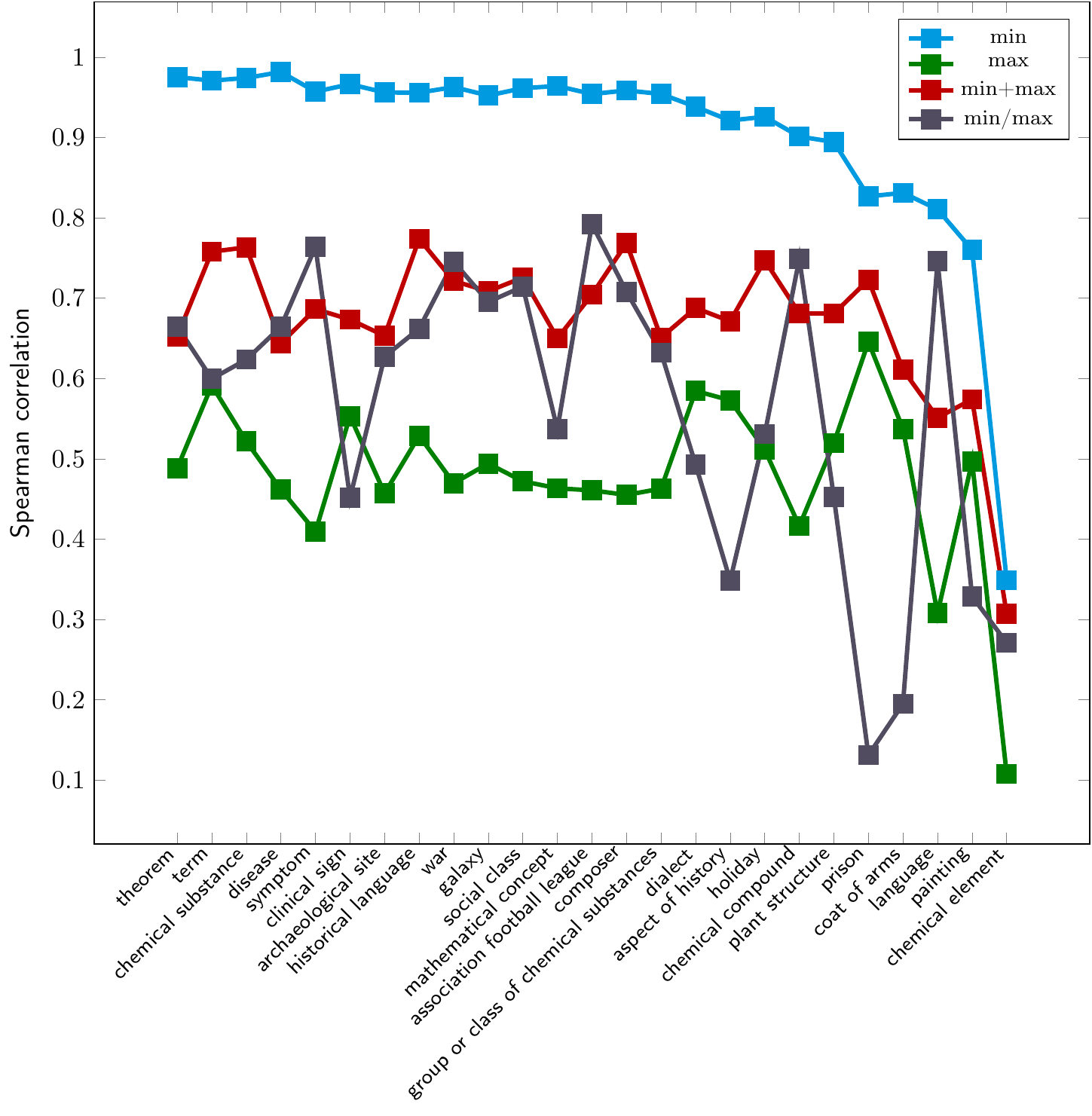}
    \caption{Rank correlation of the sizes (calculated as min (\textcolor{SeminarBlau}{blue}, used as a reference in descending order of correlation), max (\textcolor{SeminarGruen}{green}), sum (\textcolor{SeminarRot}{red}) and ratio (\textcolor{SeminarGrau}{gray})) of paired networks (INs) and their similarities computed per subject area using four measures: 
    upper left: SV2 (thematic intratextual similarity), 
    upper right: SV4-DeltaCon (intertextual similarity), 
    lower left: (SV4-GES),
    lower right: (SV4-CGS).
    Filled boxes indicate significant correlations ($p < 0.05$).
    }
    \label{fig:Correlation:2}
\end{figure}
That is, we ask whether the rank of a pair of networks (the higher their similarity $\sigma(I_i^x, I_j^x)$, the higher the rank) 
correlates with its rank according to size $s(I_i^x, I_j^x)$ (the larger the networks, the higher the rank), distinguishing four alternatives:
$s(I_i^x, I_j^x) = \min(|V_i^x|, |V_j^x|)$, $s(I_i^x, I_j^x) = \max(|V_i^x|, |V_j^x|)$, $s(I_i^x, I_j^x) = |V_i^x| + |V_j^x|$ and $s(I_i^x, I_j^x) = \min(|V_i^x|, |V_j^x|) / \max(|V_i^x|, |V_j^x|)$.
The distributions are shown in Figure \ref{fig:Correlation:1}, the corresponding correlations in Figure \ref{fig:Correlation:2} (which additionally plots the values for CGS).
For SV2, we observe a very strong effect regarding \textit{language} (\enquote{visually} confirmed by Table \ref{tab:Longtable})
(note that the curves are ordered according to variant $\min(|V_i^x|, |V_j^x|)$, sorted in descending order); in most other cases correlations are rather low (whether positive or negative).
From this picture we conclude that observed thematic similarities of articles in different languages on the same subject cannot be attributed to the sizes of the networks involved. 
Remarkably, for SV2 we observe that the data series mostly coincide with variant $\min(|V_i^x|, |V_j^x|)$ regarding the correlations' order.
This more or less also applies to DeltaCon in Figure \ref{fig:Correlation:2}.
However, we now mostly observe higher negative correlations.
Apparently, ranks in terms of structural similarity correlate negatively with size-related ranks. 
This means that if the INs are small, their similarities are likely to be large and vice versa.
In the case of GES (Figure \ref{fig:Correlation:2}) we find this assessment more or less reversed: 
the correlations (blue) are almost all significantly higher than 0.2; that is, the larger the networks involved, the higher the similarity.
From this perspective we can conclude that either both measures (GES and DeltaCon) contradict each other or (the more likely interpretation) they measure orthogonal aspects of graph similarity (the one set intersection-, the other walk-based).
An exception is again CGS, which shows stable, high rank correlations not only for small graphs, \textit{for which it computes very high values of dissimilarity}, but also depends more on size than SV2 and DeltaCon -- as motivated by the definition of CGS, size is a better predictor of it: 
the smaller the INs, the less their similarity and vice versa.
In any case, the picture we get from this analysis is ambiguous, so that we hesitate to conclude that hypertextual similarity is reliably correlated with the orders of the networks involved:
\textit{intratextually, the similarity of INs does not depend on size and intertextually it does not show a clear trend.}

\begin{table}
    \centering
    \includegraphics[width=0.45\linewidth]{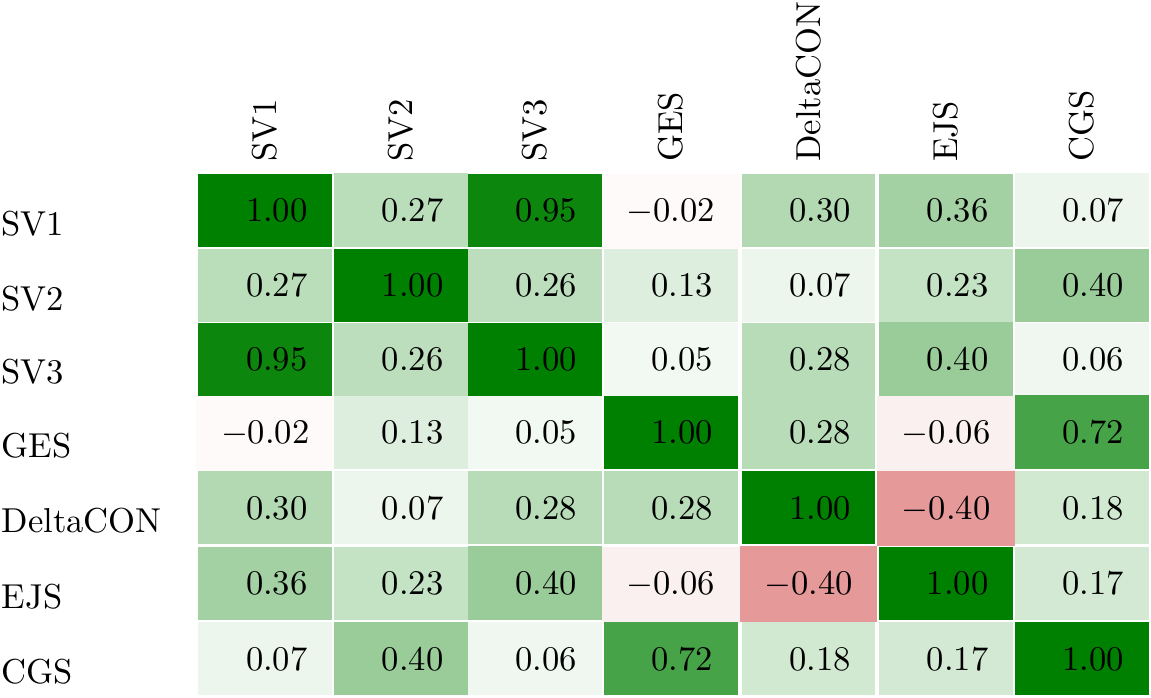}
    \hfill
    \includegraphics[width=0.45\linewidth]{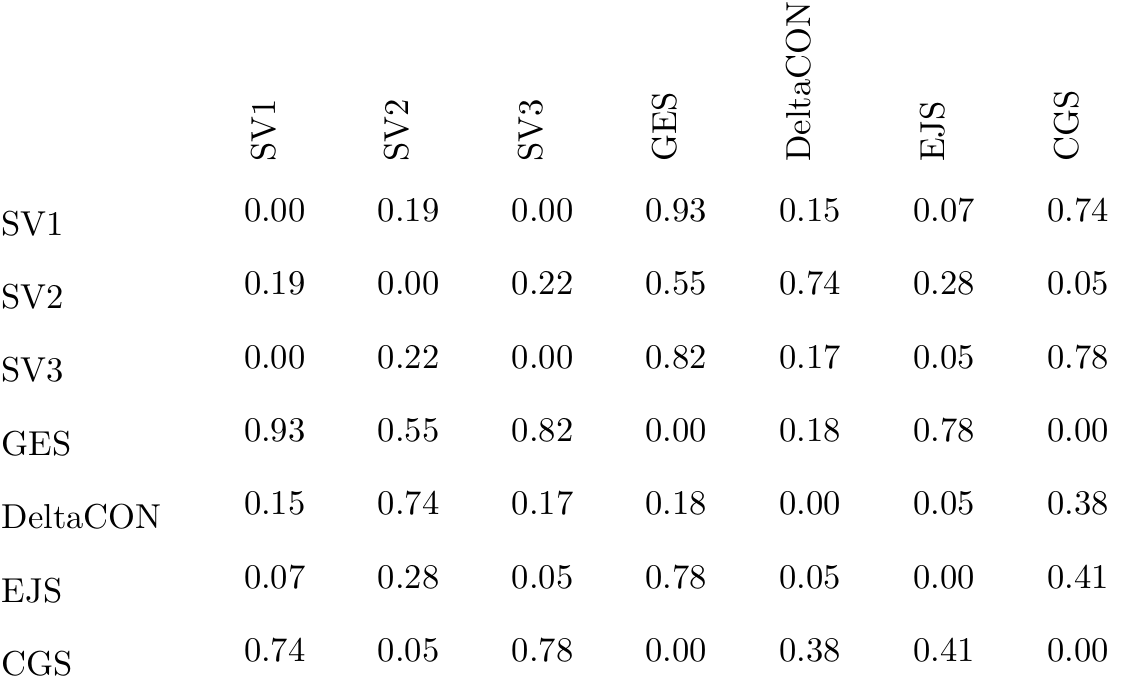}
    \caption{Rank correlations of the orderings of subject areas according to the average similarities obtained for the corresponding INs of language pairs where the similarities are computed by means of seven different similarity measures. 
    Left: Correlation, 
    right: p-Values.}
    \label{fig:Correlation:3}
\end{table}

Next, we ask about the status of subject areas as a function of their analyses along SV1-4.
Since each similarity measure induces a ranking of areas based on the average of the similarities observed for the language pairs (see Table \ref{fig:fuzzyGraphs}), we can ask whether the rankings induced by different measures correlate or not. 
Lower rank correlations would then indicate unsystematic similarity relations in the sense that intra- and intertextual similarities do not point in the same direction.
Lower rank correlations for either intra- or intertextual measures, in turn, would point to contradictory results.
All in all, such findings would indicate that the INs under consideration exhibit incoherent similarities -- contrary to the assumption of their uniform similarity along intra- \textit{and} intertextual dimensions.
This is essentially what we find in Table \ref{fig:Correlation:3}: 
rank correlations are mostly low and not significant.
One exception is the negative correlation of EJS and DeltaCon, which, apparently, measure measure different things.
Another exception are the few examples of high positive correlations such as those of GES and CGS.
This correlation analysis shows that, with few exceptions, the similarity-based ordering of subject areas along one similarity dimension (whether intra- or intertextual) does not allow us to infer their order along another dimension.

\begin{figure}[t]
\begin{tabular}{c|c|c}
	{\includegraphics[width=0.32\linewidth]{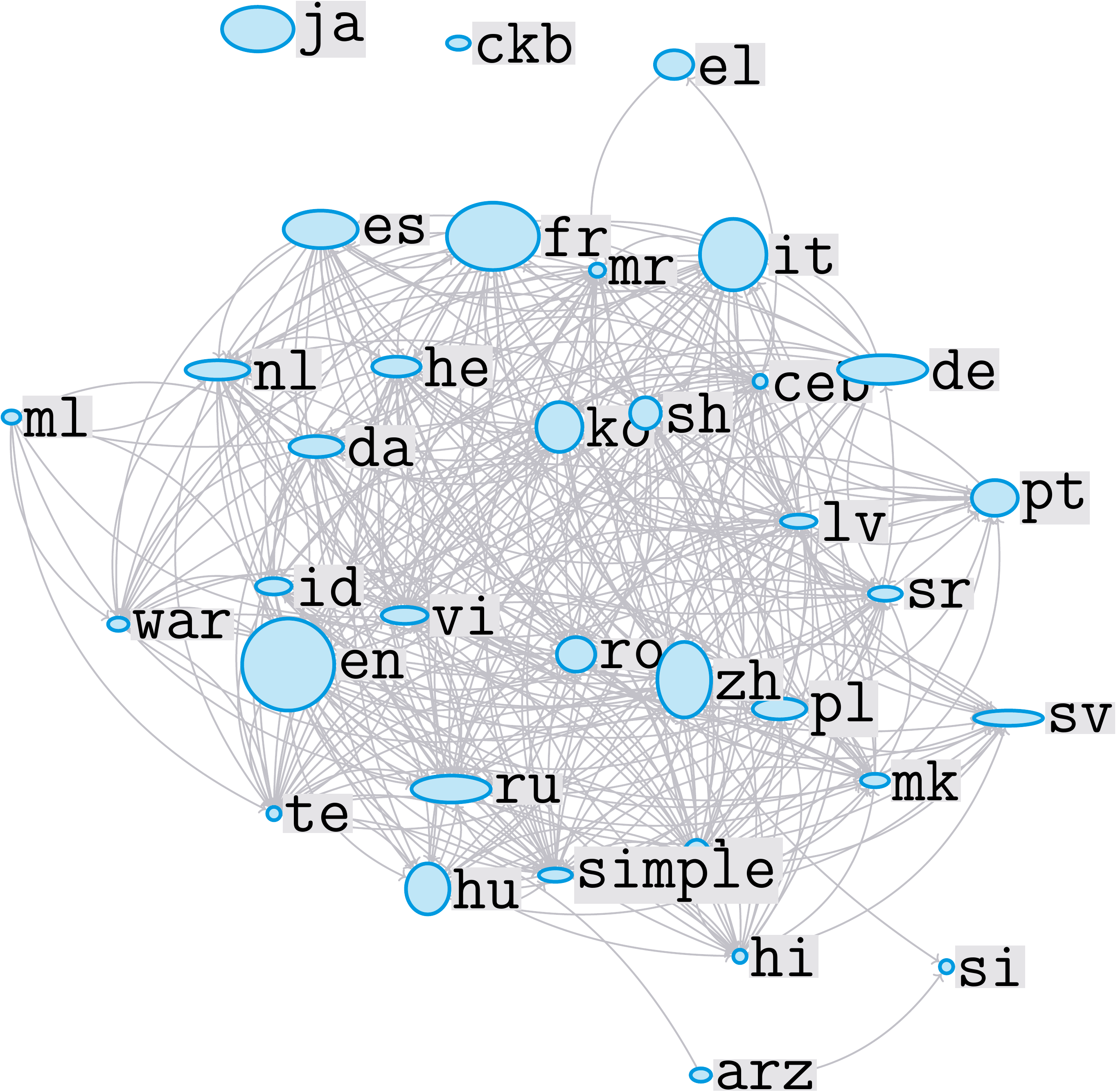}}
	&
	{\includegraphics[width=0.32\linewidth]{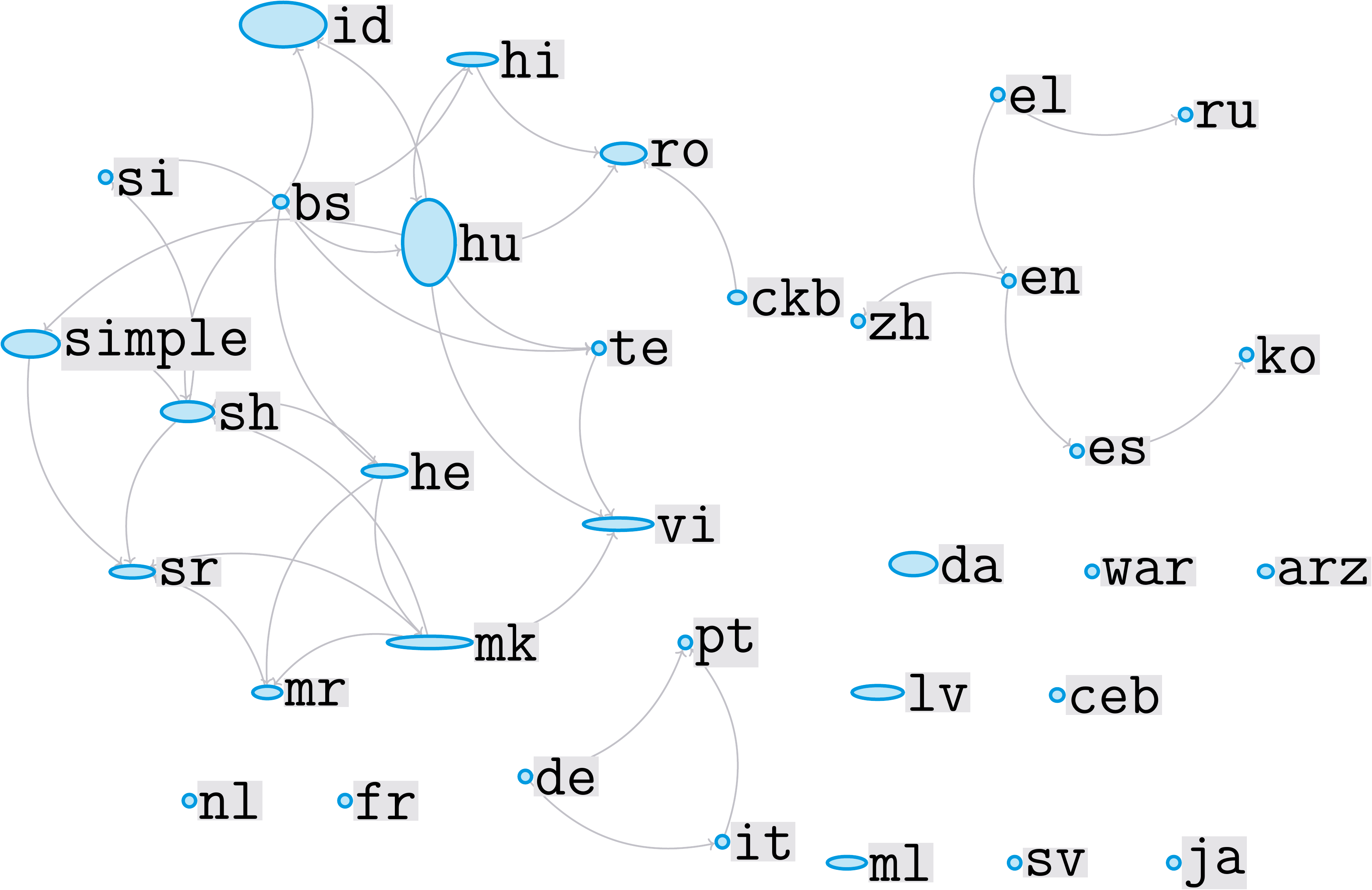}} 
	&
	{\includegraphics[width=0.32\linewidth]{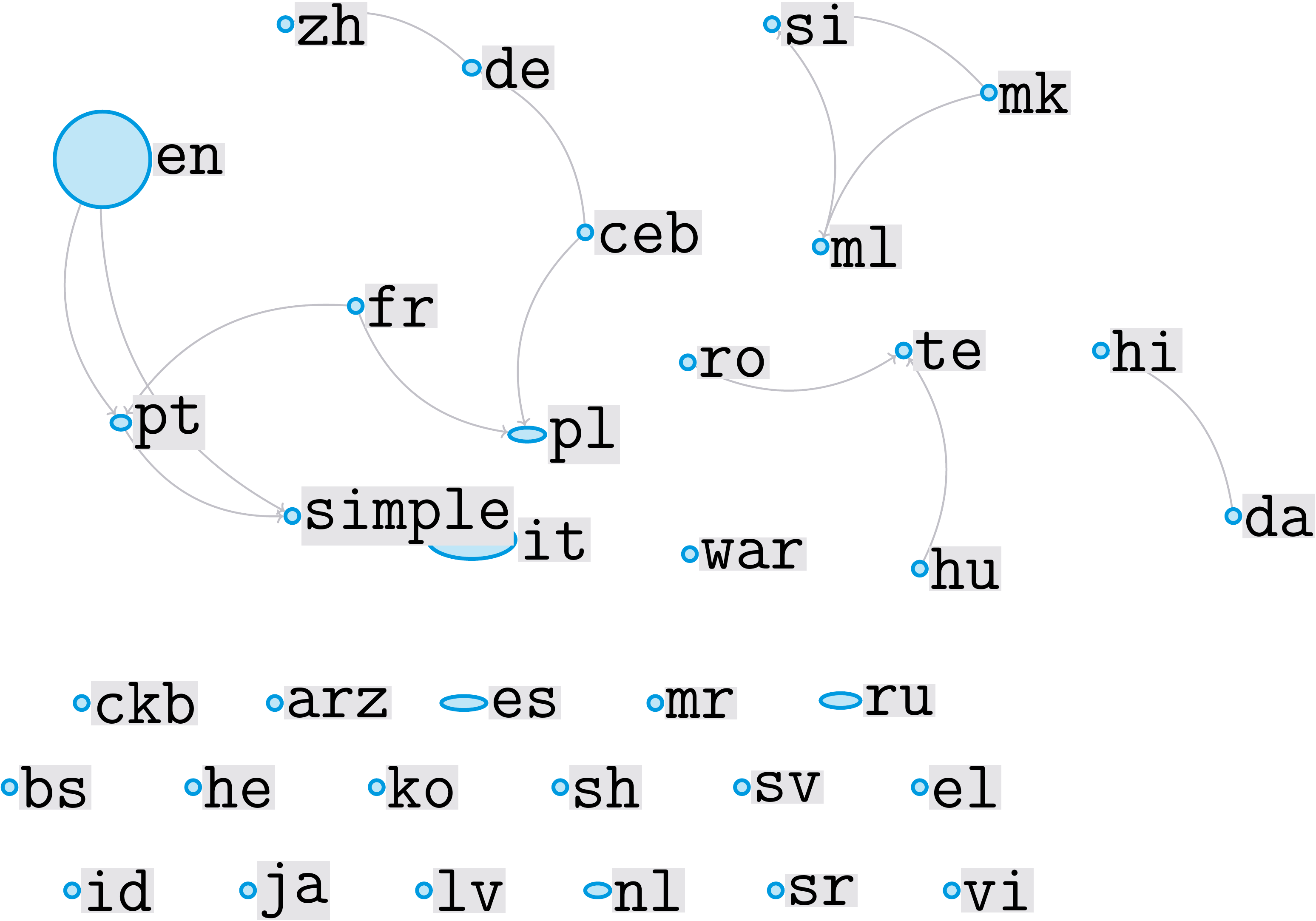}} \\ \hline
	{\includegraphics[width=0.32\linewidth]{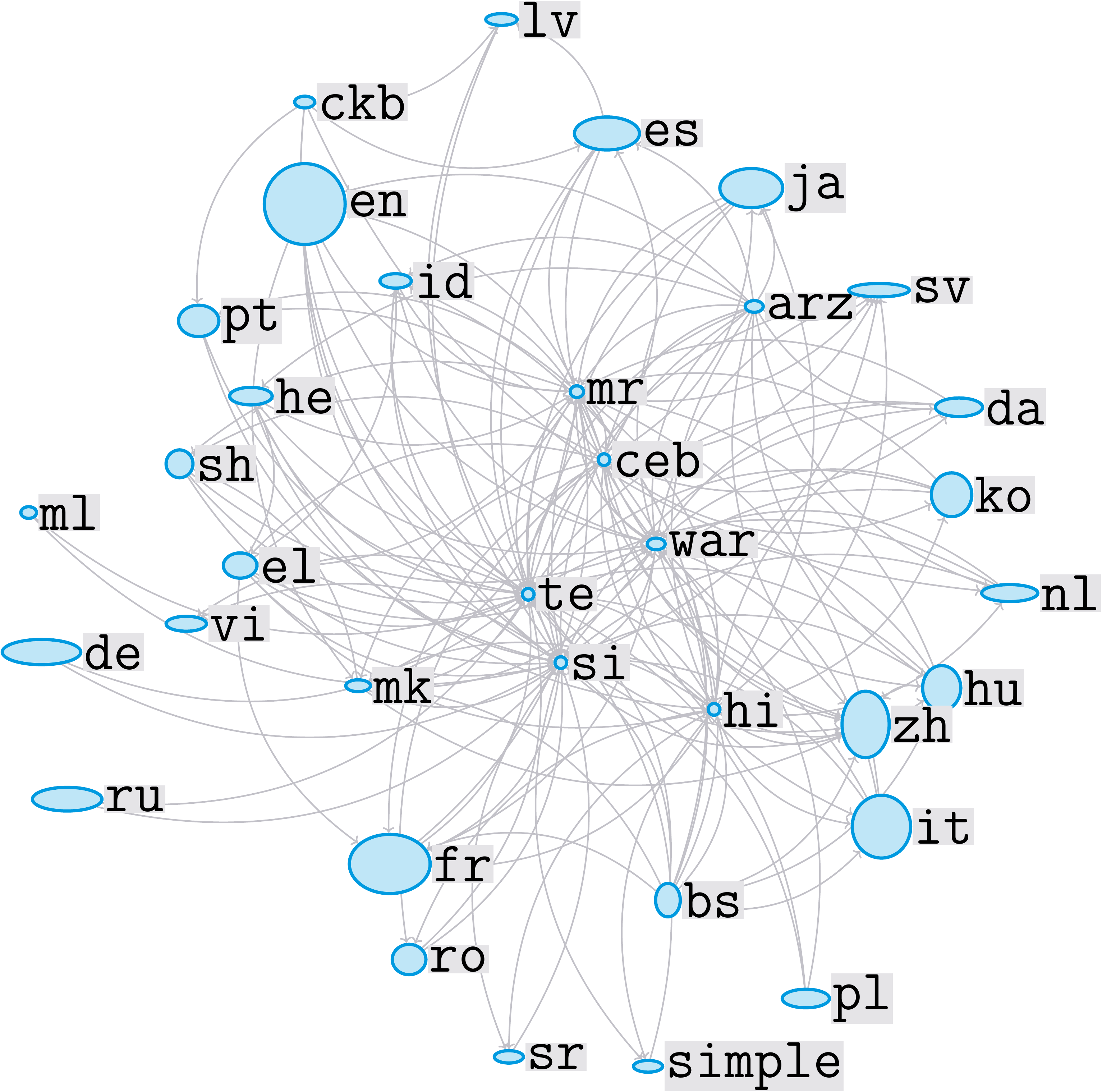}}  
	&
	{\includegraphics[width=0.32\linewidth]{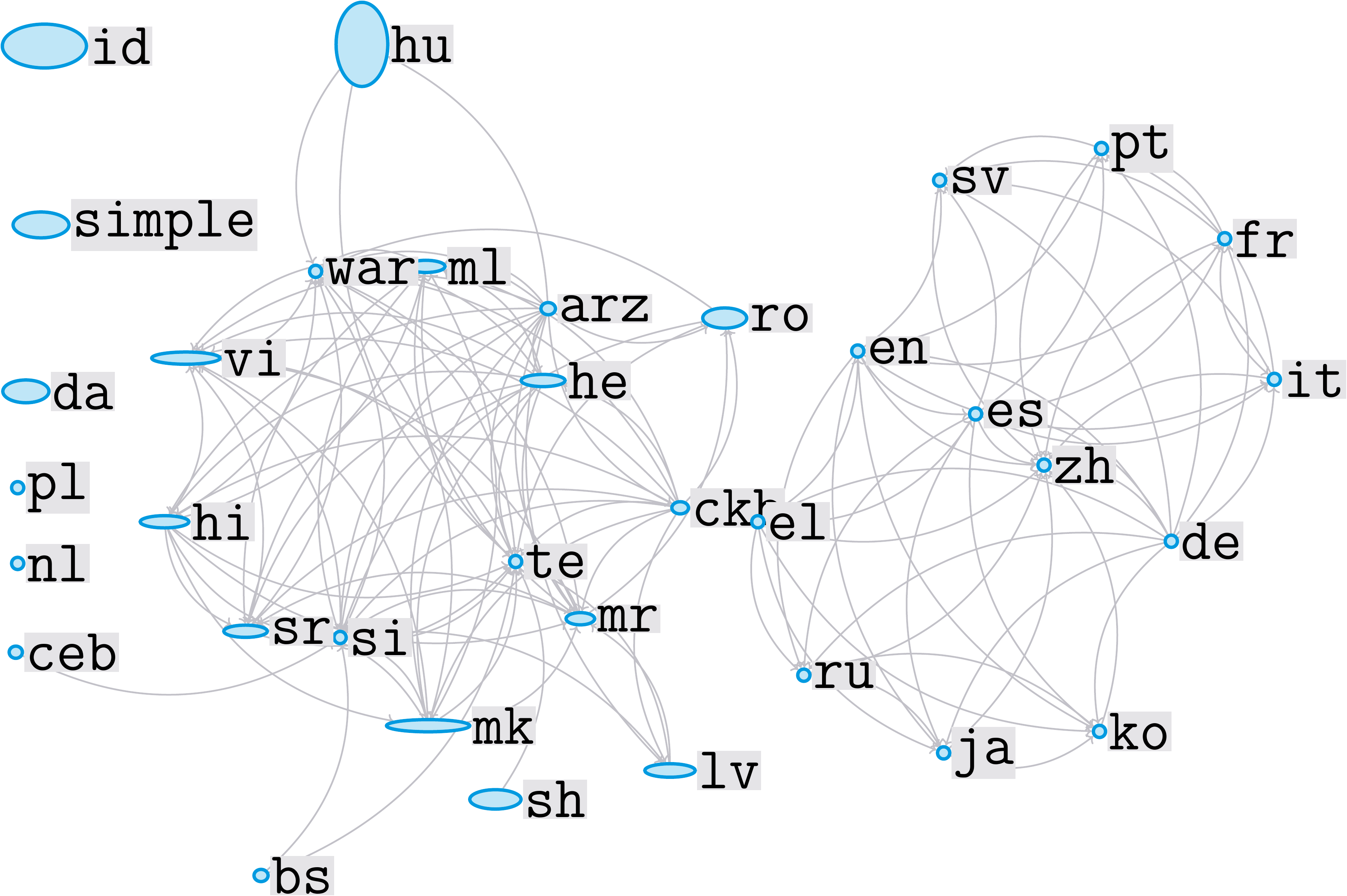}}
	&
	{\includegraphics[width=0.32\linewidth]{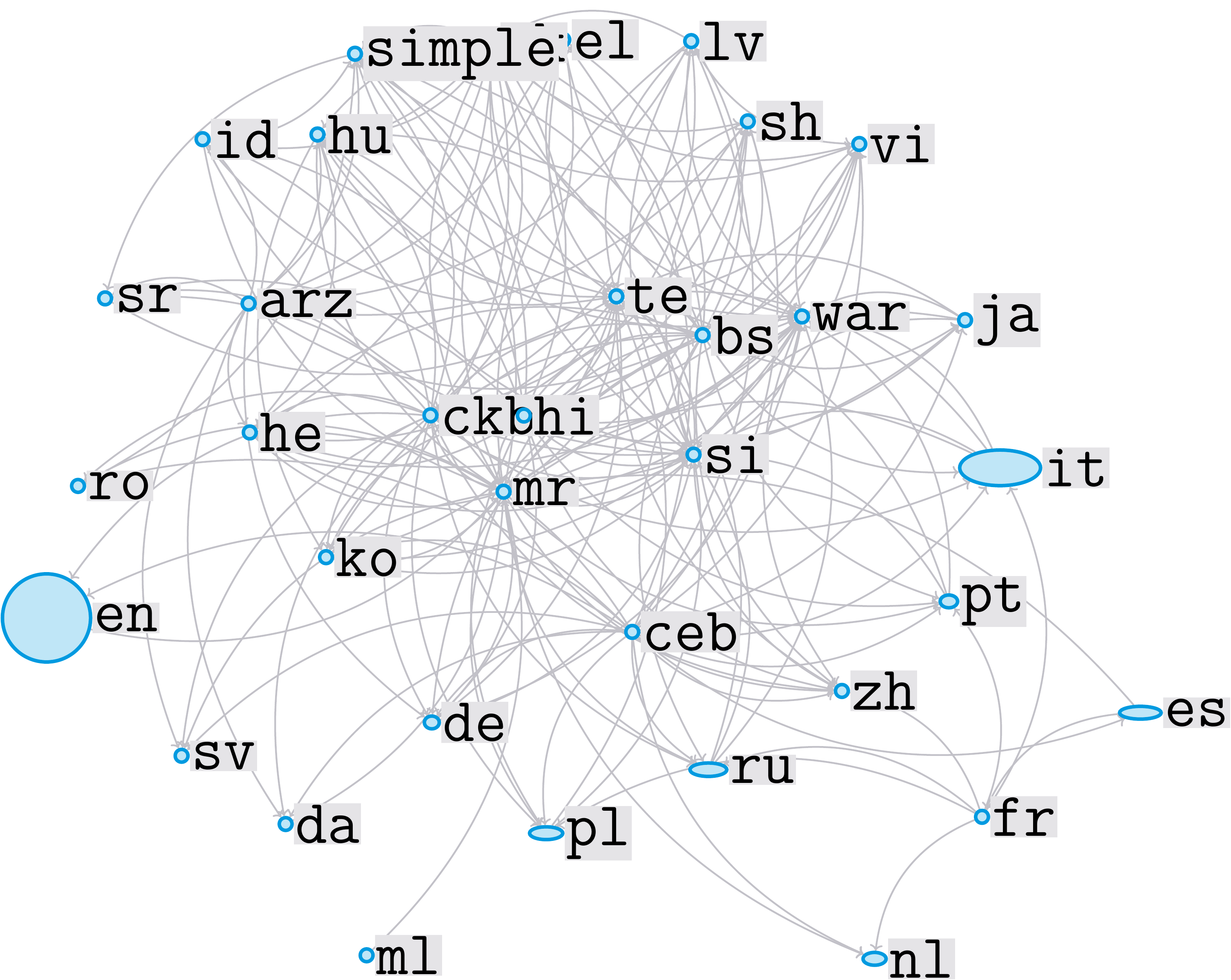}}
\end{tabular}
\caption{Language networks as a function of above average similarities computed by SV2 (1st row; thematic similarity) and SV4 (2nd row; DeltaCon). First column: subject area \textit{composer}; second column: \textit{language}; third column: \textit{painting}. The higher the number of vertices in the underlying IN, the \textit{wider} the resulting language vertex; the higher the IN's number of edges, the \textit{higher} the language vertex.}
\label{fig:Networks}
\end{figure}

Next, we consider language networks whose edges correspond to the similarity values of the underlying language pairs (related network analyses have been conducted by \citet{Miquel-Ribe:Laniado:2016,Samoilenko:et:al:2016}). 
We want to know in which subject areas which language clusters arise and whether different languages are center-forming in different subject areas (see question Q3).
%
%
Since the underlying similarity matrices (see Table \ref{tab:Longtable}) induce complete graphs, we filter out all edges whose similarity values are below the average similarity measured across all subject areas for the respective measure.
Figure \ref{fig:Networks} illustrates the result by the example of three subject areas (\textit{composer}, \textit{painting} and \textit{language}) and two similarity views (SV2 and SV4-DeltaCon).
The impression that we get already by this selection is rather confusing:
the graphs look very different, from the perspective of the similarity measure and from the perspective of the subject area.
A clear trend is not discernible (although English Wikipedia is usually prominently positioned).
From this brief network analysis, we conclude that the different languages may have different degrees of salience depending on the subject area, while there is no single language that dominates in all these cases. 

\begin{figure}[t]
\centering
\begin{tabular}{c|c|c}
	{\includegraphics[width=0.45\linewidth]{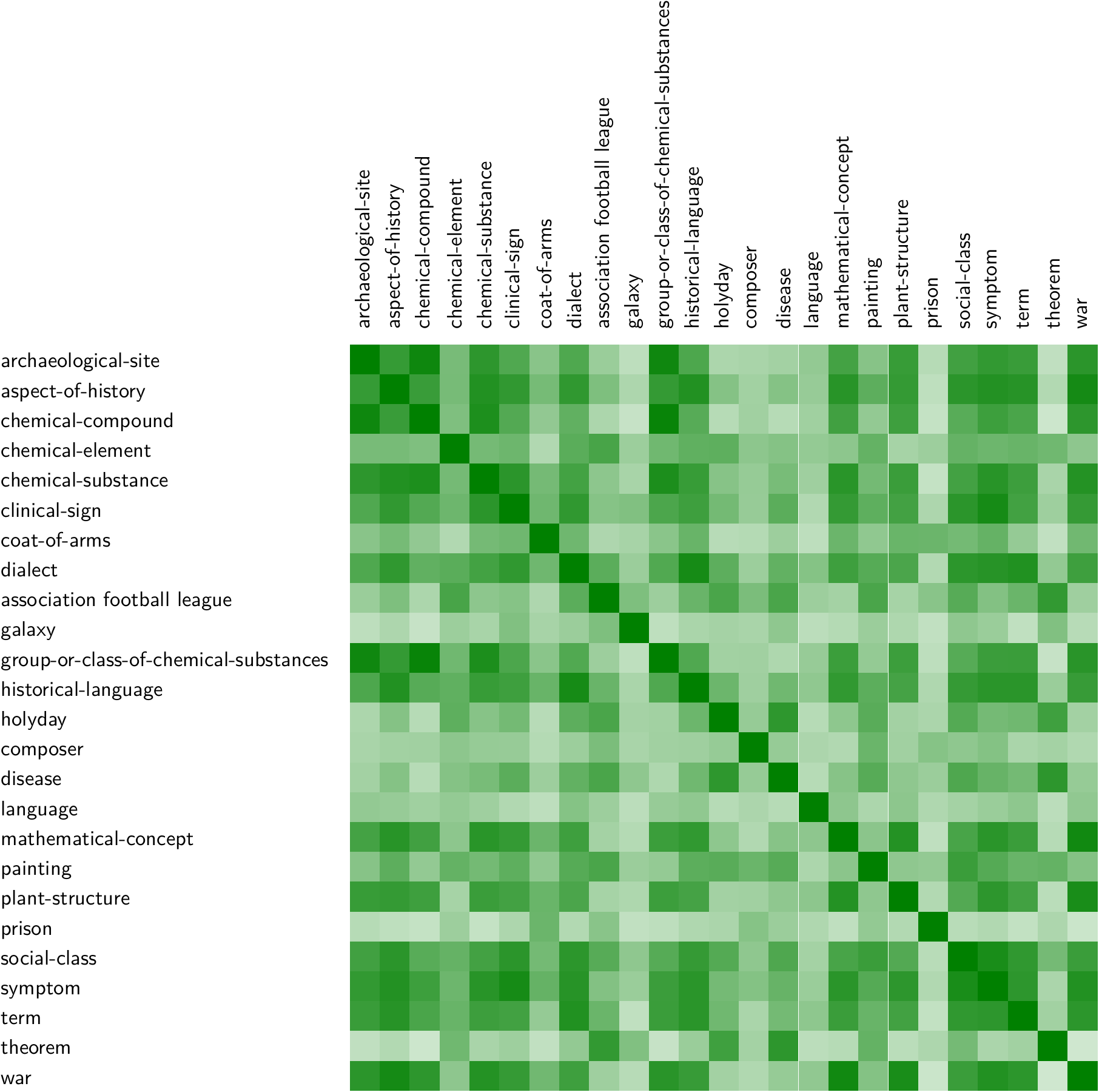}} &
	{\includegraphics[width=0.45\linewidth]{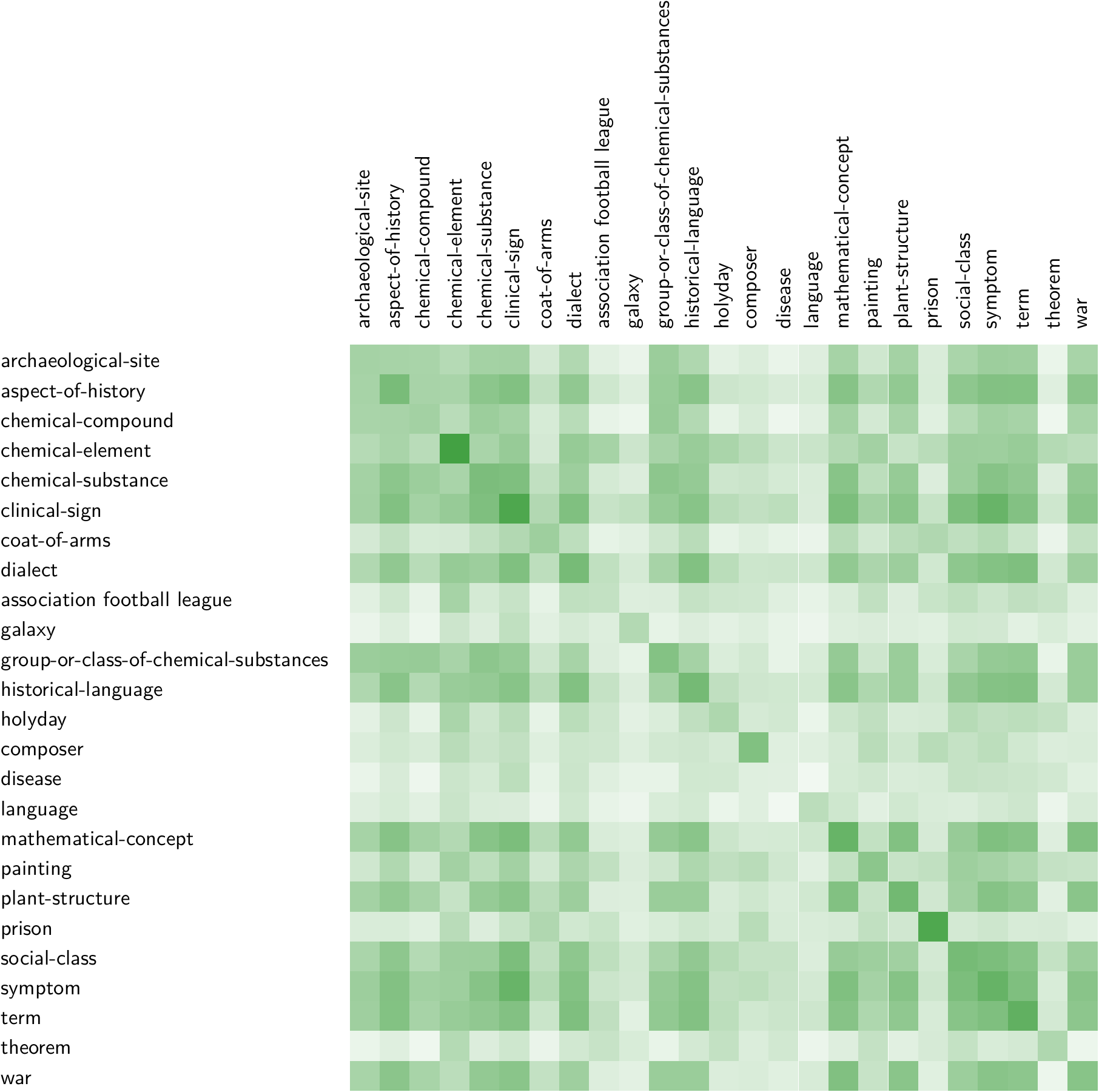}} \\[3pt] \hline
	{\includegraphics[width=0.45\linewidth]{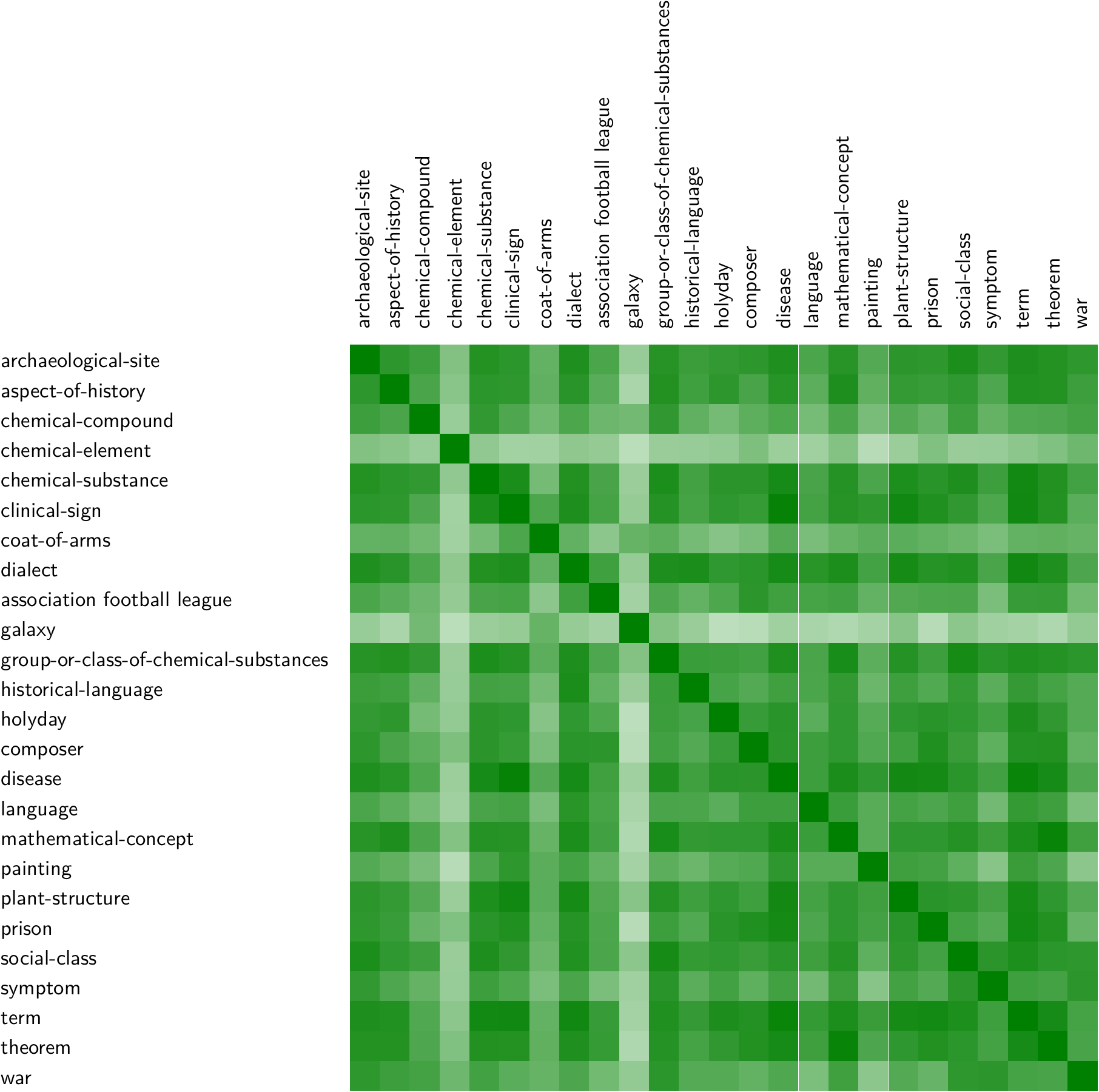}} &
	{\includegraphics[width=0.45\linewidth]{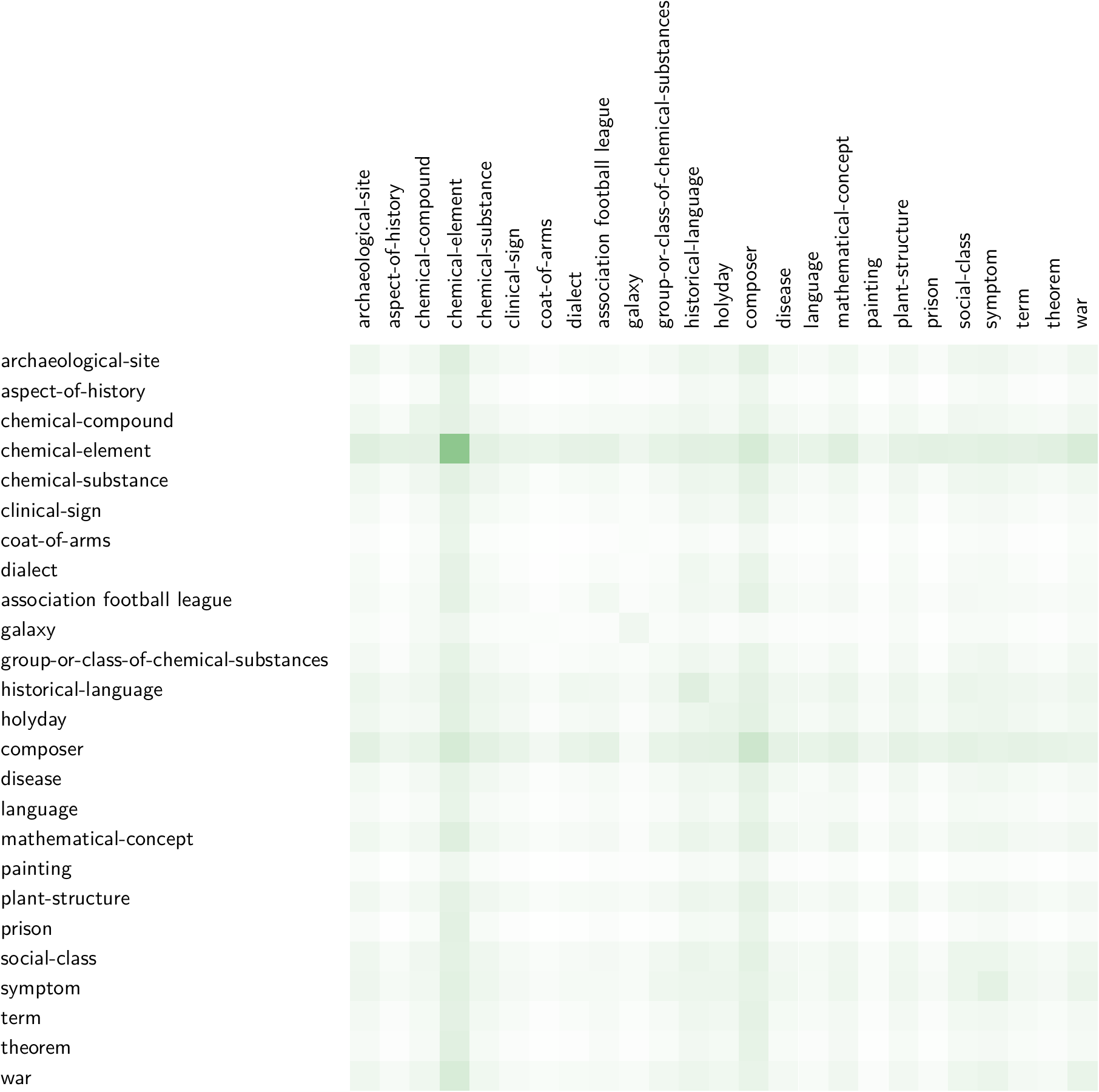}} 
\end{tabular}
\caption{From left to right: Unweighted and weighted distance correlations of the centralities of 35 languages in 25 language networks. First row: DeltaCon; second row: CGS.}
\label{fig:network_centrality_analysis}
\end{figure}

\begin{figure}[t]
\centering
\begin{tabular}{c|c|c}
	{\includegraphics[width=0.45\linewidth]{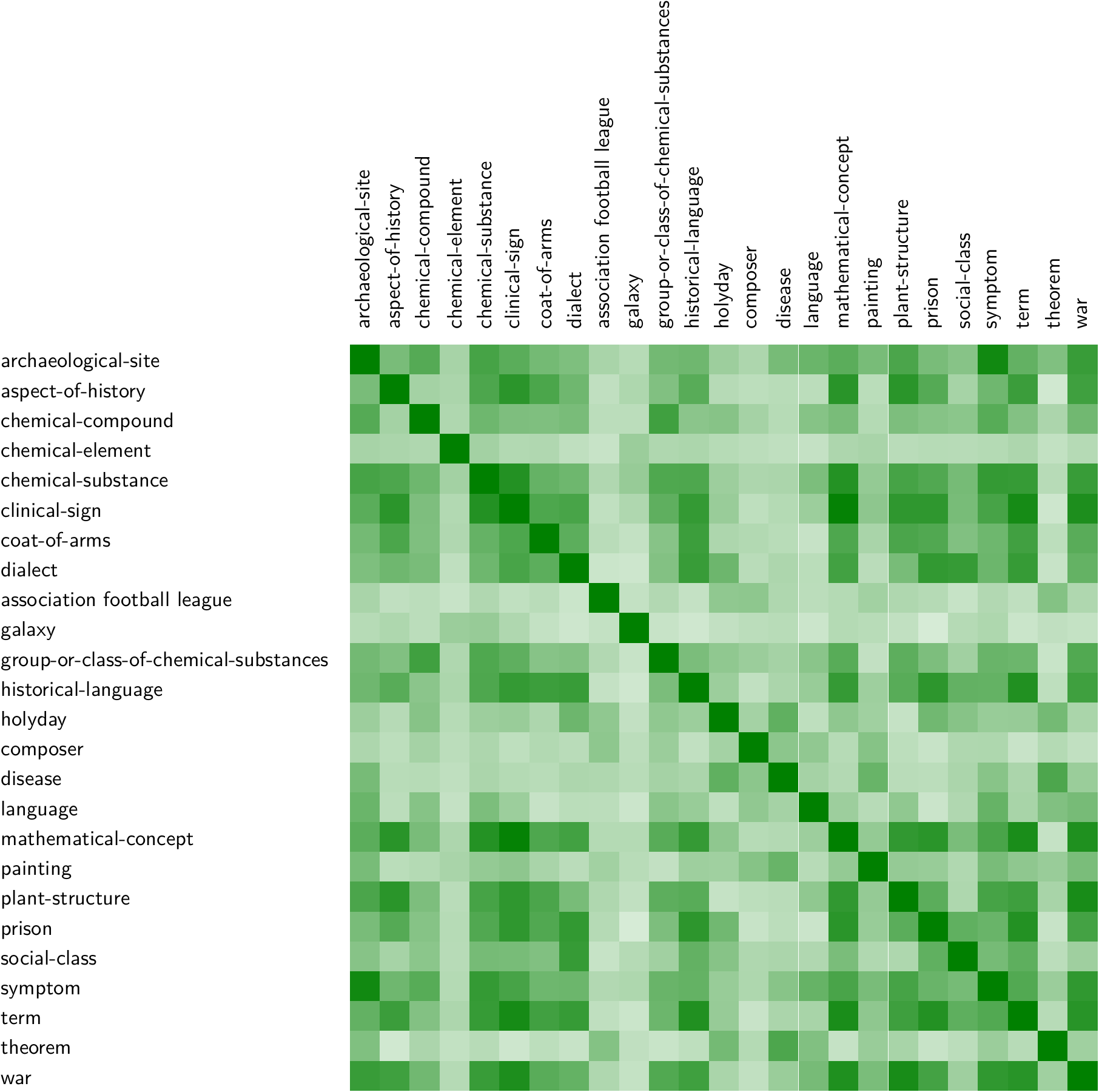}} &
	{\includegraphics[width=0.45\linewidth]{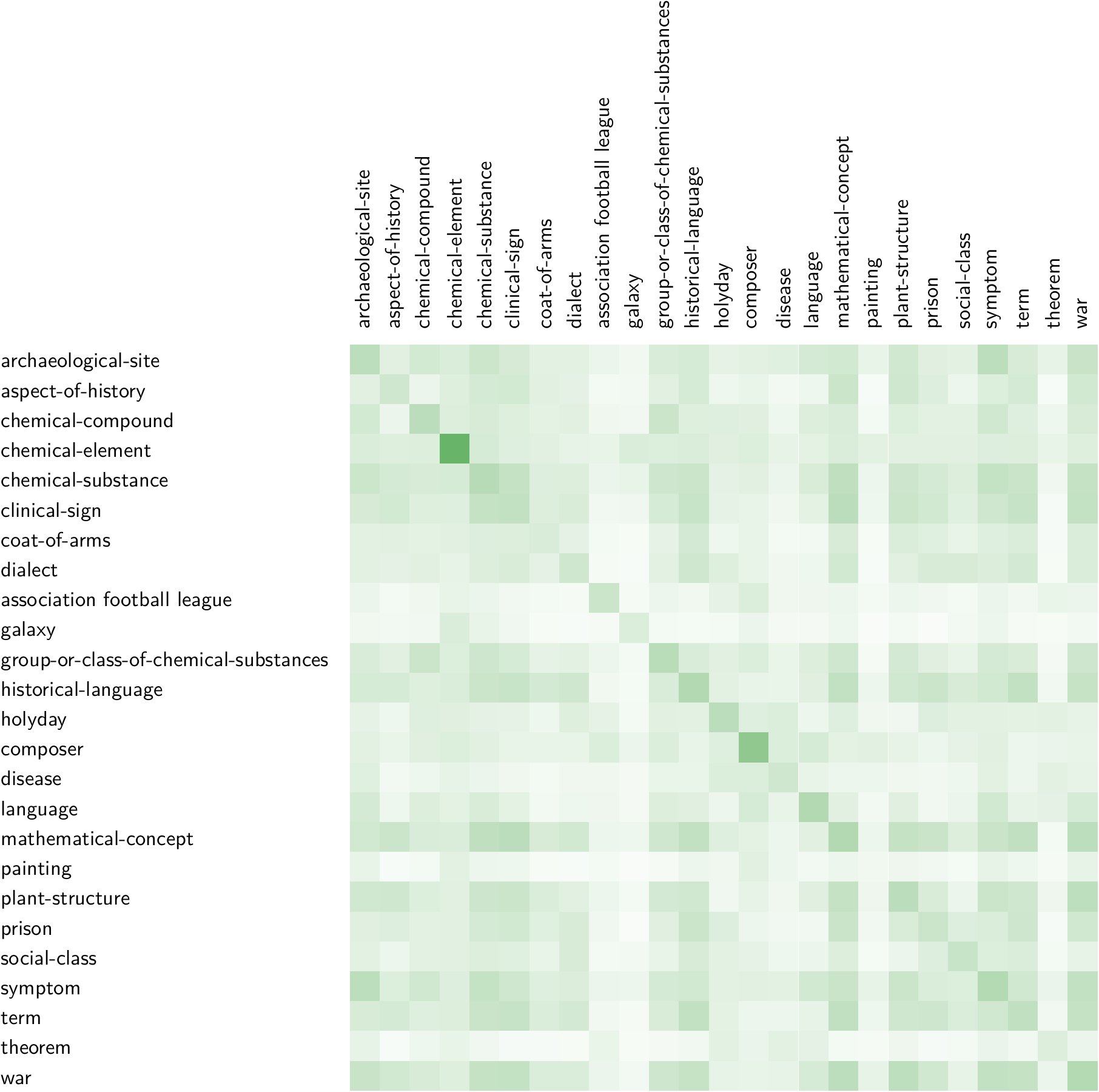}} \\[3pt] \hline
	{\includegraphics[width=0.45\linewidth]{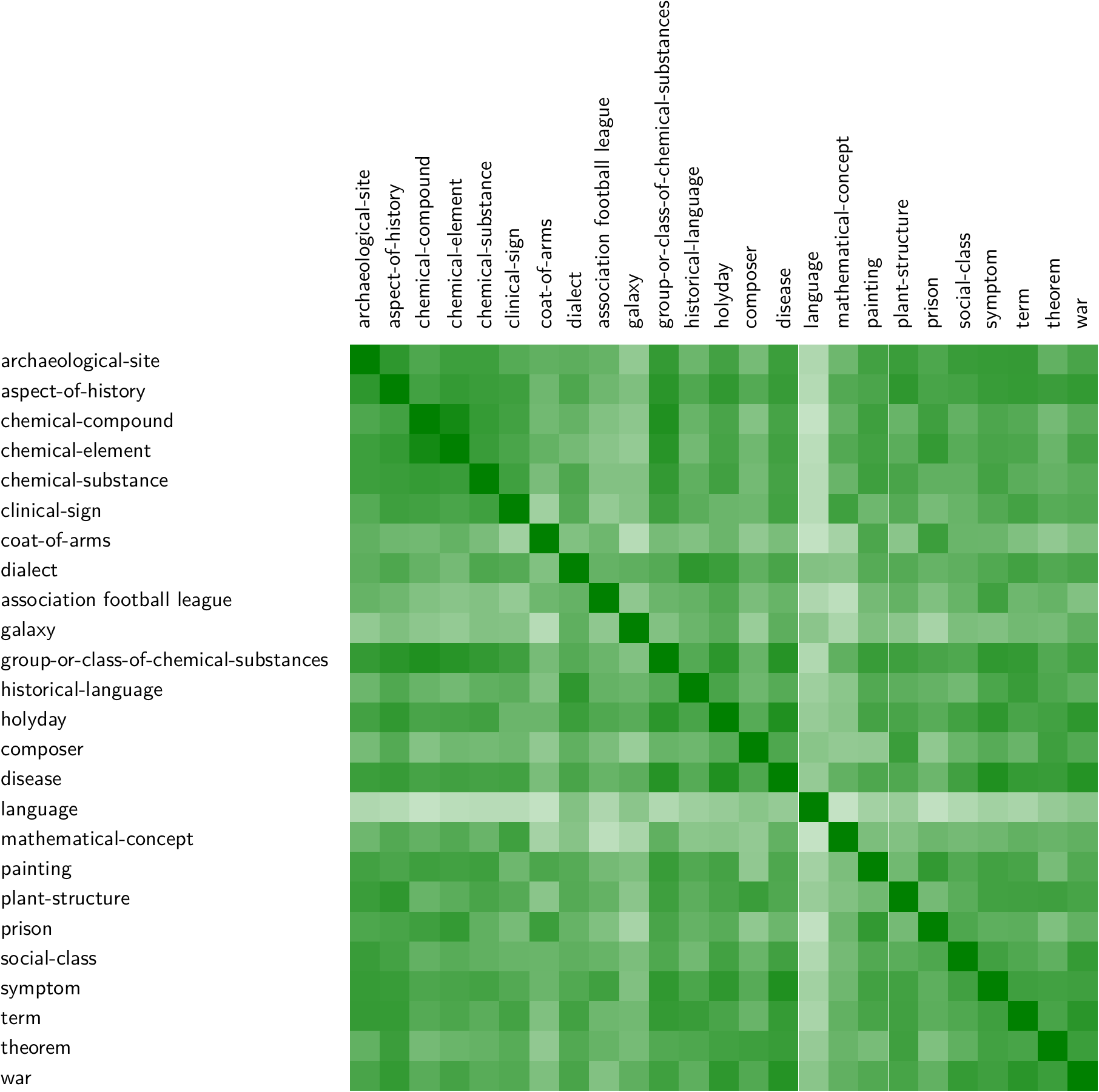}} &
	{\includegraphics[width=0.45\linewidth]{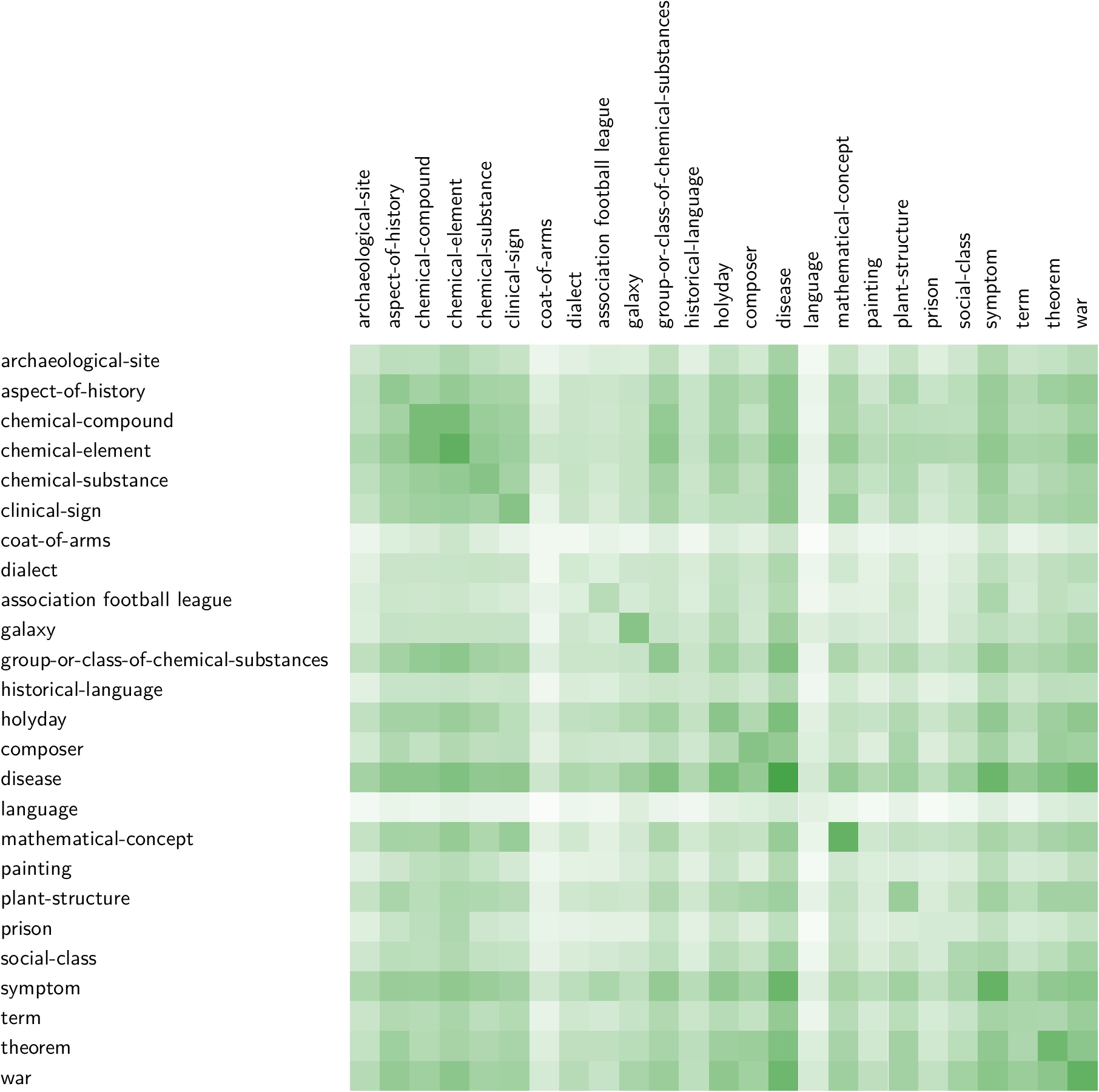}}
\end{tabular}
\caption{From left to right: Unweighted and weighted distance correlations of the centralities of 35 languages in 25 language networks. First row: GES; second row: SV2.}
\label{fig:network_centrality_analysis:2}
\end{figure}

To conclude this analysis, we ask about the similarities of the positions of all 35 languages in the networks derived from the heatmaps of Table \ref{tab:Longtable} and weight them by the ratio of these networks' strength centralities to the respective maximum.
That is, we take the completely connected language similarity network for each of the 25 topics, in which each edge is weighted according to the similarity of the associated languages in the sense of the underlying similarity measure.
%
Regardless of the actual structure of the INs, in the ideal state of equal INs for the different languages about the same topic, it is to be expected that the edges in the language networks are weighted by 1. 
Under this regime, the distance correlations of the strength centralities of languages should be maximal when being compared for each pair of topic-related language networks:\footnote{For the notion of vertex strength (i.e.\ weighted degree values) see \citet{Barrat:Barthelemy:Pastor-Satorras:Vespignani:2004:a}, for computing network centralities based on vertex indices see \citet{Feldman:Sanger:2007}.}
the languages then occupy identical network positions -- independent of the topics.
However, if all INs are maximally dissimilar, we likewise get maximum distance correlations of the languages according to their network positions. 
Thus, to differentiate between these two cases, we weight observed distance correlations by the quotient of observed network strength and maximum possible strength.
For pairs of the fully connected language networks considered here, this maximum is 35 * 34, assuming a maximum edge weight of 1. 
To avoid unrealistically low weights, we multiply this maximum by the actually observed maximum value per similarity measure. 
The resulting heatmaps are shown in Figure \ref{fig:network_centrality_analysis}: 
If INs on the same topic tend to be similar, both heatmaps are expected to be saturated, the weighted and the unweighted. 
Conversely, if these INs tend to be dissimilar, only the unweighted map is likely to be saturated.
In Figure \ref{fig:network_centrality_analysis} we show the corresponding results for the optimistic variant of similarity measurement (DeltaCon) and its pessimistic counterpart (CGS).
Obviously, with CGS we obtain higher distance correlations of the strength centralities of the languages in the 25 language networks than with DeltaCon, but very small weighted variants of these correlations, which indicate that the INs are extremely far from ideal similarities.
The only exception is \textit{chemical element}: 
this topic is described similarly in the different languages and more close to the ideal scope than any other topic.
In the case of DeltaCon the weighted distance correlations are much higher, but still far from ideal.
In Figure \ref{fig:network_centrality_analysis:2} we add the distance correlations generated by GES and SV2: 
for GES the unweighted correlations are similar to those of DeltaCon, for SV2 those of CGS.
The situation is the opposite in the case of the weighted correlations.
The general tendency is that the languages tend to have similar network positions, but at the price of lower similarities of the INs.

\subsection{Discussion}

Given the scope and growth dynamics of Wikipedia, it is not surprising that a reader who reads it in a particular language expects to be sufficiently broadly and deeply informed about the subject of his or her information search, provided it is one of the larger Wikipedias. 
We have shown that such an assumption about Wikipedia as a central building block of the IL does not apply -- at least in the context of the subject areas examined here.
Their different language versions differ so much in their treatment of the same subject area that it is necessary to know which area in which language someone is consulting if one wants to know how much the part of the IL he or she is traversing is biased.
It may be the case that a reader's consultation of Wikipedia is accompanied by the assumption, that it is an open, dynamically growing resource that largely covers relevant fields of knowledge, where one likely finds what one is looking for, and vice versa, that what is irrelevant is excluded. 
One might even assume that if a language version of Wikipedia is only sufficiently large, it will probably show this pattern of coverage -- regardless of the underlying language.
We have shown that at the current stage of development, such assumptions do not apply.

Our analysis has shown that subject areas covered by Wikipedia differ from language to language in a way that is hardly predictable by the size of the networks involved.
Consequently, with regard to question Q1, we have to state that the similarities between the languages vary from subject area to subject area. 
It is therefore necessary to define the thematic reference (subject area) in order to say something about comparable languages. 
This may seem obvious, but it shows that Wikipedia's language editions are designed differently for different research fields and subject areas. 
With few exceptions (e.g.\ \textit{chemical element}), we find that inequalities prevail within the same field. 
This makes it difficult to say which field is the more evenly distributed, language-independent one, which ensures that students can expect nearly equal information coverage regardless of language.
Thus, in relation to question Q2, we conclude that the choice of the subject area has a major influence on the  similarity assessment.
And because of this influence, we do not discover a lingua franca which, due to its size and coverage would serve as a kind of reference for the similarity relations of the different languages, so that although they may be dissimilar among themselves, they would be predominantly similar with respect to this central language.
As things stand, the data do not support the attribution of this central role to English Wikipedia -- at least from the point of view of the fields of knowledge and subject areas considered here (question Q3).
In other words, knowing that Wikipedia is small in your first language and therefore probably does not cover the subject area of your task, it is hardly a way out of this dilemma to recommend reading English Wikipedia instead: 
it would be better to read at least both.
However, we also see that depending on the subject area, other languages could play the role of primary sources (e.g.\ Hungarian in the case of subject area \textit{language} -- see Figure \ref{fig:Networks} -- or French in the case of subject area \textit{composer}) -- of course, this presupposes that one has the language skills for them.

\citet{Samoilenko:et:al:2016} point out the influence of multiple dimensions on the commonalities, similarities and differences of Wikipedias language editions, including language, culture and geographical proximity. 
Apparently, we shed light on this view from the perspective of our four-part similarity analysis, which distinguishes between intratextual (text structural, quantitative, and thematic similarity) and intertextual (hypertextual) aspects, where the latter are simultaneously examined by means of four measures.
The important aspect is that these reference points may be influenced to varying degrees by linguistic, cultural or even geographical factors.
In the case of SV3 (quantitative text profiles), for example, we observed rather evenly distributed similarities of the INs of the different subject areas, at a higher similarity level.
The differences observed in this context may indicate the influence of the underlying languages, which their authors are less or barely able to \enquote{escape} through their own control.
The extent to which SV3 models law-like behavior of texts would then make INs of different or related languages look more similar -- and differences could indicate a strong influence of the respective languages.
SV1 (LDS) shows a similar pattern, although this may be due to the underlying webgenre, its editing rules and the way Wikipedia monitors compliance with them.
However, such a finding could negate the role of this SV.
In any case, one should not underestimate the source of information on which SV1 is based, as it concerns elements of documents that are likely to be the focus of reading: 
tables and especially figures and pictures. 
In this respect, a further development is needed that integrates text analysis with image analysis and related approaches.

SV2 (thematic structure) differs from SV1 and SV3: 
It decreases rapidly and reaches a very low level of similarity (in case of subject area \textit{disease} -- see  Table \ref{fig:fuzzyGraphs}).
%
%
Part of this dynamic is likely caused by the diverging F-values of our topic model (see Table \ref{tab:languages}). 
But let us assume its effectiveness.
What could be the cause then, if we do not look for factors such as size or age of a Wikipedia?
Take the example of the subject area \textit{disease}: 
could it be that it is cultural differences that determine which diseases are described in which language edition by means of which intertextual structures (see the low similarity values measured with DeltaCon in Table \ref{tab:Longtable})? 
Is there, so to speak, a cultural or any comparable disposition for the arrangement of intertextual contexts, while the corresponding intratextual similarities (SV2 in Table \ref{tab:Longtable}) tend to be much higher?
In the case of subject area \textit{holiday} (which according to SV2 and GES occupies a position in the middle, according to DeltaCon in the lower range and according to EJS in the upper range of the similarity spectrum), cultural references are rather likely.
Whether true or not, this gives rise to the question of which intertextual structures, which link-based factors for the production of multiple texts are culturally determined or have a cultural imprint.
Is there, so to speak, a linguistic, cultural or knowledge area-related fingerprint that could be read from intertextual structures, from different parts of the IL, a fingerprint which could help to explain its dynamics beyond what is done by lexical-semantic analyses of link anchors?
Even though our research does not answer this question, it does raise it and thus builds a bridge between the kind of data science we pursue and text-linguistic questions from the field of reading research.

To answer these questions, we need a multiple text model that includes the underlying IL as a limiting factor of what results from reading processes as multiple text; 
a model that considers linguistic, cultural, genre- and register-related \citep{Halliday:Hasan:1989} as well as social factors when asking about the function or meaning of a given or missing link. 
It is obvious that such a model-theoretical extension of reading research benefits considerably from browsing models such as those developed by more recent hypertext research \citep{Lamprecht:Lerman:Helic:Strohmaier:2017,Dimitrov:Singer:Lemmerich:Strohmaier:2017}: 
it may not be surprising that readers are more likely to select links in the initial sections of Wikipedia articles according to text-structural criteria.
However, this research has also highlighted the importance of semantic criteria for link selection. It is now necessary to further explore this system of motives and to extend it to the generation of multiple texts as a whole.
In this context, our approach to DeltaCon is of interest, which evaluates random walks as a source of information to assess vertex affinities.
The reason for this is that it can easily be linked to empirical research on the reading or navigation behavior of users in Wikipedia and comparable resources.
In this way, it would integrate intertextual structural analysis with the pragmatics of real hypertext use. 
By additionally integrating the article-content-related similarity views SV1-3 as developed here, it would open up a very broad spectrum of information sources for the analysis and comparison of multiple texts, namely syntactic, semantic and pragmatic sources.
This could ultimately pave the way to go beyond the detection of limiting factors, as the IL in the form of Wikipedia imposes on reading, to gain models of how this IL is actually represented by its readers and entire reader communities through distributed reading processes, that is, as a distributed cognitive map of the IL.

Irrespective of these findings, conclusions and prospects for future work, a number of boundary conditions must be considered with regard to our research:
\begin{enumerate}
\item \textit{Topic analysis:}
Although we have implemented an extensible procedure for extracting INs as intertextual manifestations of subject areas, which in principle can take into account any of the millions of Wikidata items to conduct cross-linguistic studies, we have only analyzed a subset of 25 such areas.
This approach could be extended by asking about the convergence of similarities/dissimilarities between languages, as a result of studying a much wider range of subject areas.
In this sense, we provided the starting point for a more detailed examination of the thematic biases of Wikipedia, compared to what has been studied so far.
This level of detail originates from our three-level topic model (Figure \ref{fig:ThreeLevelTopicModel}), which should be expanded into a model of thematic-rhematic intertextuality \citep[cf.][]{Mehler:Gleim:Gaitsch:Uslu:Hemati:2019}.

\item \textit{Similarity analysis:}
We implemented a hybrid approach to measure the similarities of INs, the structure-oriented part of which includes four approaches to graph similarity measurement. 
However, the spectrum of relevant measures is much wider \citep{Emmert-Streib:Dehmer:Shi:2016}, so that their expressiveness and significance for intertextuality research should be examined more thoroughly.
One may even think of a multiple source similarity measure that simultaneously maps various structural and other informational sources to assess the similarity of multiple texts and hypertexts.

\item \textit{Network analysis:}
The modeling of reading requires the modeling of cognitive processes, which in the case of distributed reading means the modeling of processes in social networks.
%
%
As explained in our introduction, such an endeavor requires modeling the \enquote{fluent alignment} of two processes:
on the one hand, the multi-authorial writing of Wikipedia and its embedding in the larger IL and, on the other hand, the diversity of reading situations, their task contexts and contextual resources by which they are conditioned.
In this way, our approach could be further developed by integrating models of web-based writing research\footnote{For a recent network theoretical sentiment analysis of online writing see, for example, \citet{Stella:Restocchi:DeDeyne:2020}. For a review of network theoretical approaches to knowledge networks in education science see \citet{Siew:2020}.} and social network analysis.
\end{enumerate}

Such a more comprehensive model of distributed reading and writing is certainly a challenge for future research, especially if it is to be based on thorough linguistic analysis. 
Currently, we see no way out of such a research direction for education science, that is, for studying learning in the age of information.

\subsection{How Does our Approach Relate to Linguistic Relativity?}\label{sec:Linguistic Relativity}

Starting from a selected set of topics, we have shown that different language Wikipedias produce quite different networks for informing about the same topics.
That is, we detected a bias: 
extent and organization of a topic's representation depend on the underlying Wikipedia -- the former are biased by the latter.
Since we related this bias to the languages in which the Wikipedias are written, we spoke of a \textit{linguistic bias}.
This raises the question what our approach contributes to research on what is known as \textit{Linguistic Relativity} (LR) \citep{Lakoff:1987,Lucy:1997} or \textit{Cultural Relativity} (CR)  \citep{Gumperz:Levinson:1991,Gumperz:Levinson:1996} (cf.\ also \citealt{Sharifian:2017}).
Are the differences we observe caused by differences in the underlying languages (LR) or even by cultural differences (CR) between the communities of writers producing these Wikipedias?
Our approach does not allow for a direct answer to this question, as it is not based on the linguistic or social data required for such an undertaking.
Nevertheless, it is worth explaining how it relates to this research.

To clarify this we utilize the distinction of \textit{structure-},  \textit{domain-} and \textit{behavior-centered} approaches according to \citet{Lucy:1997}:
the first start from observed semantic differences between languages to examine their influence on thought, the second from \enquote{domain[s] of experienced reality} \citep[p.298]{Lucy:1997} to ask how languages represent them, the third from language-specific practices of use.
%
%
According to \citet[p.298]{Lucy:1997}, the distinguishing feature of domain-specific approaches is that they characterize domains independently of the target languages. 
This is what we intend to do when deriving topic representations from Wikidata items and their relations (cf.\ \citealt{Mehler:Pustylnikov:Diewald:2010:a}): 
we explore this data to gain access to conceptual representations of parts of experienced reality, ask how they are described in different Wikipedias and whether the networked descriptions based thereon are commensurable or not.
Our first assessment is that most approaches to Wikipedia's LR or CR \citep[cf.][]{Massa:Scrinzi:2012,Laufer:Wagner:Floeck:Strohmaier:2015,Miquel-Ribe:Laniado:2016,Miz:Hanna:Aspert:Ricaud:Vandergheynst:2020} are domain-centered in such a way.
Before assessing what they can say about LR, we go one step further in characterizing our approach, this time with the help of \citet[p.322]{Lakoff:1987} who discusses conceptual organization as a reference point for assessing the commensurability of conceptual systems, with which our approach to graph similarity is apparently compatible.
%
In particular, \citet[p.334]{Lakoff:1987} concludes that organizational differences point to different conceptual systems and thus to LR.
Let us assume that INs \textit{manifest} such conceptual systems hypertextually. 
The differences we find in these INs could then reflect conceptual differences of the underlying languages and, if these differences are culturally determined, CR:
conceptualizations of the subject areas investigated here would then be linguistically relative and ultimately culture-specific.
%
%
If languages encode world views \citep[2]{Gumperz:Levinson:1996}, Wikipedias can then be seen as manifestations of parts of such views, and since languages differ in encoding them, their Wikipedias are consequently non-trivially different.
Moreover, due to the distributed authorship of Wikipedia, there is a direct link to the concept of distributed cognition \citep{Hollan:Hutchins:Kirsh:2000}, to which approaches to CR are connected \citep{Gumperz:Levinson:1991,Sharifian:2017}.
From this point of view, it seems plausible to assume effects of different communities, each of which produces representations of conceptual systems or world views in the form of hypertexts, that are more or less incommensurable.
From this it follows that by examining such differences we should get access to the differences of the underlying worldviews and their encodings.
To sum up:
we use Wikidata to identify cross-linguistic conceptual units (as a bridge to experienced reality), examine their language-specific lexicalizations (article names) and interconnections (hyperlinks) using corresponding excerpts from Wikipedia, interpret their differences as evidence of LR and speculate on CR as its cause (see above). 

Apparently this consideration already brings us to the end of the analogy. 
The reason for this assessment concerns the research objects of the areas compared here.
More precisely, the question is to which entities or systems the observed differences are finally ascribed. 
While research on LR aims to make statements about \textit{language systems} -- beyond the level of lexis -- or \textit{conceptual systems} (by asking whether the same parts of reality (e.g.\ the color spectrum) are conceptualized and coded in a language-specific way), we and our relatives in Wikipedia make statements about  \textit{textual instances} of such systems.
The former deal with differences in language systems, while we study hypertextual differences without directly drawing conclusions about the underlying systems -- beyond the lexical level.
%
%
That is, we do not directly contribute to research on LR at the level of \textit{language as system} but rather at the level of \textit{language as text} \citep{Hjelmslev:1969}.
Moreover, beyond the question of whether or not a concept is manifested and networked in a language, we do not consider cases of conceptual splitting, fusion, etc.
From this point of view, it is certainly no overstatement to claim that we found evidence of the type of linguistic bias described above.
But it would be an overstatement to claim that we thereby measure LR on the level of language systems and the underlying cultures.
Though we can speculate that cultural differences are responsible for the differences we measure, we cannot yet prove this with our apparatus. 
To consider language as a system, linguistic analyses are required such as those provided by comparative linguistics \citep{Bisang:Czerwinski:2019}.
For example, we can ask about differences in the linguistic manifestation of topics (e.g.\ information structure, density and uncertainty, relevance, salience, etc.) and what effects this has on the organization and linking of articles. 
Based on this, we could ask for language-specific text patterns and attribute them to author communities and thus to cultural differences, but we would need independent data to support such conclusions. 
At least the linguistic part of this task now lies within the interdisciplinary reach of comparative and computational linguistics, so that we can put research on \textit{reading/writing multiple texts} on a broader methodological basis. 
This is work for the future.

\section{Conclusion}\label{sec:Conclusion}

We introduced a three-level topic model in combination with a graph-theoretical model for measuring the intra- \textit{and} intertextual similarities of article networks from different language editions of Wikipedia.
%
%
In this way we built a bridge between reading research, educational science and computational linguistics.
To this end, we described a new perspective for reading research that focuses more on the information landscape as a limiting factor of online reading. 
We have continued research showing that Wikipedia exhibits a topical coverage bias.
However, we have done this using a much more elaborate topic and text structure model in conjunction with a quantitative model of hypertext structure, a hybrid model that is more realistic from the perspective of hypertext linguistics \citep{Storrer:2002}.
In future work we will continue elaborating our computational hypertext model. 
This will be done with special attention to hypertext usage to obtain graph models that integrate browsing behavior into graph similarity analysis.






\section*{Conflict of Interest Statement}

The authors declare that the research was conducted in the absence of any commercial or financial relationships that could be construed as a potential conflict of interest.

\section*{Author Contributions}
AM has written Section 1, 2, 4 (together with PW) and 5; he generated Figure 1, 2, 3, 4, 6, 7, 11, 13, 14 and Table 1. 
AM, MK, PW, TU and WH have written Section 3.
WH extracted INs from Wikidata and Wikipedia and generated Figure 8, 9 and 10 as well as Table 3, 4 and 5; together with AM, he generated Figure 5.
TU generated Table 2.
AM (CGS), MK (SV1, SV2), PW (EJS, DeltaCon), TU (SV2) and WH (GES) implemented, computed and described the similarity measures. 
TU  trained and tested text2ddc.
AM, MK, PW and WH proofread and revised the final manuscript.
All authors contributed to the conception of the paper, its algorithmization and the interpretation of the results.


\section*{Funding}

WH is partly financed by the State of Hessen through the LOEWE research focus on  \textit{minority studies: language and identity}.
PW is partially funded by the Federal Ministry of Education and Research of Germany as part of the competence center for machine learning ML2R (01|S18038C) and is partially funded by the Deutsche Forschungsgemeinschaft (DFG, German Research Foundation) under Germany’s Excellence Strategy -- EXC 2070 -- 390732324.



\section*{Data Availability Statement}
Code and datasets used for this study can be found on github:  \url{https://github.com/texttechnologylab/WikiSim}.

\color{black}
\bibliographystyle{frontiersinSCNS_ENG_HUMS} 
\bibliography{InfoLandscape.bib}

\end{document}

%% file: Tabellen/Subject-Areas.tex
	\begin{tikzpicture}
		[Kasten/.style={rectangle, minimum height=0.40cm, anchor=west}, Path/.style={font={\small\vphantom{Ag}}, midway, above},]
		\matrix [
			matrix of nodes, 
			nodes=Kasten, 
			row sep=0.03cm, 
			column sep=0.03cm, 
			font={\scriptsize\vphantom{Ag}}, 
		column 1/.append style={nodes={text width=0.35cm, align=center, fill=SeminarSehrSehrHellGrau}}, 
		column 2/.append style={nodes={text width=3.6cm, align=left, fill=SeminarSehrSehrHellGrau}},
		column 3/.append style={nodes={text width=4.3cm, align=left, fill=SeminarSehrSehrHellGrau}},
		column 4/.append style={nodes={text width=0.35cm, align=center, fill=SeminarSehrSehrHellGrau}},
		row 1/.append style={nodes={fill=SeminarSehrHellBlau}},
		row 6/.append style={nodes={minimum height=1.0cm}},
		row 15/.append style={nodes={minimum height=1.0cm}},
		row 16/.append style={nodes={minimum height=1.0cm}},
		row 26/.append style={nodes={minimum height=1.0cm}},
		] (Link) {
			& Subject Area & OECD category & \# \\
1 & disease & Health sciences & 35 \\
2 & chemical-compound & Chemical sciences & 35 \\
3 & archaeological-site & History and Archaeology & 35 \\ 
4 & war & History and Archaeology & 35 \\
5 & association football league & Economics and Business & 31 \\ 
6 & aspect-of-history & History and Archaeology & 35 \\ 
7 & painting & Arts & 35 \\
8 & chemical-element & Chemical sciences & 35 \\
9 & theorem & Mathematics & 33 \\
10 & galaxy & Physical sciences & 29 \\
11 & dialect & Languages and Literature & 35 \\ 
12 & term & Languages and Literature & 35 \\ 
13 & coat-of-arms & History and Archaeology & 28 \\
14 & group-or-class-of-chemical-substances & Chemical sciences & 35 \\ 
15 & holiday & Philosophy, Ethics and Religion & 35 \\
16 & symptom & Basic medicine & 35 \\
17 & language & Languages and Literature & 34 \\
18 & chemical-substance & Chemical sciences & 34 \\
19 & social-class & Sociology & 32 \\
20 & plant-structure & Biological sciences & 33 \\
21 & prison & Sociology & 34 \\ 
22 & clinical-sign & Basic medicine & 34 \\
23 & historical-language & Languages and Literature & 32 \\
24 & composer & Arts & 35 \\
25 & mathematical-concept & Mathematics & 34 \\ 
		};
	\end{tikzpicture}

%% file: Tabellen/Languages1.tex
	\begin{tikzpicture}
		[Kasten/.style={rectangle, minimum height=0.40cm, anchor=west}, Path/.style={font={\small\vphantom{Ag}}, midway, above},]
		\matrix [matrix of nodes, nodes=Kasten, row sep=0.03cm, column sep=0.03cm, font={\scriptsize\vphantom{Ag}}, 
		row 1/.append style={nodes={fill=SeminarSehrHellBlau, align=left}}, 
		column 1/.append style={nodes={text width=0.5cm, align=center, fill=SeminarSehrSehrHellGrau}}, 
		column 2/.append style={nodes={text width=1.0cm, align=left, fill=SeminarSehrSehrHellGrau}},
		column 3/.append style={nodes={text width=2.5cm, align=left, fill=SeminarSehrSehrHellGrau}},
		column 4/.append style={nodes={text width=1.2cm, align=right, fill=SeminarSehrSehrHellGrau}},
		column 5/.append style={nodes={text width=1.2cm, align=right, fill=SeminarSehrSehrHellGrau}},
		column 6/.append style={nodes={text width=1.2cm, align=right, fill=SeminarSehrSehrHellGrau}},
		] (Link) {
			& Code & Language & Topics & Train & F-score \\
			1 & ar & Arabic &  96 & \numprint{14991} & \nprounddigits{2}\numprint{ 0.798000} & \\
			2 & bs & Bosnian &  87 & \numprint{5599} & \nprounddigits{2}\numprint{0.834842} \\
			3 & ceb & Cebuano &  68 & \numprint{2069} & \nprounddigits{2}\numprint{0.8734940000000001} \\
			4 & ckb & Sorani &  82 & \numprint{3090} & \nprounddigits{2}\numprint{0.752336} \\
			5 & da & Danish &  94 & \numprint{12761} & \nprounddigits{2}\numprint{0.777692} \\
			6 & de & German &  98 & \numprint{21933} & \nprounddigits{2}\numprint{0.874} \\
			7 & el & Greek &  92 & \numprint{8395} & \nprounddigits{2}\numprint{0.764016} \\
			8 & en & English &  97 & \numprint{19772} & \nprounddigits{2}\numprint{0.8543780000000001} \\
			9 & es & Spanish &  95 & \numprint{16951} & \nprounddigits{2}\numprint{0.85} \\
			10 & fr & French &  94 & \numprint{17313} & \nprounddigits{2}\numprint{0.857592} \\
			11 & he & Hebrew &  94 & \numprint{12262} & \nprounddigits{2}\numprint{0.829365} \\
			12 & hi & Hindi &  85 & \numprint{7271} & \nprounddigits{2}\numprint{0.739771} \\
			13 & hu & Hungarian &  91 & \numprint{10854} & \nprounddigits{2}\numprint{0.848458} \\
			14 & id & Indonesian &  93 & \numprint{11265} & \nprounddigits{2}\numprint{0.8147479999999999} \\
			15 & it & Italian &  94 & \numprint{15894} & \nprounddigits{2}\numprint{0.8536309999999999} \\
			16 & ja & Japanese &  {93} & \numprint{16390} & \nprounddigits{2}\numprint{0.841306} \\
			17 & ko & Korean &  92 & \numprint{13557} & \nprounddigits{2}\numprint{0.7584310000000001} \\
			18 & lv & Latvian &  89 & \numprint{7572} & \nprounddigits{2}\numprint{0.827778} \\
		};
	\end{tikzpicture}

%% file: Tabellen/Languages2.tex
	\begin{tikzpicture}
		[Kasten/.style={rectangle, minimum height=0.40cm, anchor=west}, Path/.style={font={\small\vphantom{Ag}}, midway, above},]
		\matrix [matrix of nodes, nodes=Kasten, row sep=0.03cm, column sep=0.03cm, font={\scriptsize\vphantom{Ag}}, 
		row 1/.append style={nodes={fill=SeminarSehrHellBlau, align=left}}, 
column 1/.append style={nodes={text width=0.5cm, align=center, fill=SeminarSehrSehrHellGrau}}, 
column 2/.append style={nodes={text width=1.0cm, align=left, fill=SeminarSehrSehrHellGrau}},
column 3/.append style={nodes={text width=2.5cm, align=left, fill=SeminarSehrSehrHellGrau}},
column 4/.append style={nodes={text width=1.2cm, align=right, fill=SeminarSehrSehrHellGrau}},
column 5/.append style={nodes={text width=1.2cm, align=right, fill=SeminarSehrSehrHellGrau}},
column 6/.append style={nodes={text width=1.2cm, align=right, fill=SeminarSehrSehrHellGrau}},
] (Link) {
	& Code & Language & Topics & Train & F-score \\
			19 & mk & Macedonian &  88 & \numprint{5750} & \nprounddigits{2}\numprint{0.764925} \\
			20 & ml & Malayalam &  84 & \numprint{5465} & \nprounddigits{2}\numprint{0.8539329999999999} \\
			21 & mr & Marathi &  83 & \numprint{3061} & \nprounddigits{2}\numprint{0.8457450000000001} \\
			22 & nl & Dutch &  97 & \numprint{15507} & \nprounddigits{2}\numprint{0.848411} \\
			23 & pl & Polish &  96 & \numprint{16356} & \nprounddigits{2}\numprint{0.837249} \\
			24 & pt & Portuguese &  93 & \numprint{15663} & \nprounddigits{2}\numprint{0.840432} \\
			25 & ro & Romanian &  93 & \numprint{10690} & \nprounddigits{2}\numprint{0.772856} \\
			26 & ru & Russian &  97 & \numprint{17302} & \nprounddigits{2}\numprint{0.8538959999999999} \\
			27 & sh & Serbo-cro. &  94 & \numprint{9536} & \nprounddigits{2}\numprint{0.820896} \\
			28 & si & Sinhala &  81 & \numprint{2521} & \nprounddigits{2}\numprint{0.8342860000000001} \\
			29 & simple & Simple English &  93 & \numprint{10882} & \nprounddigits{2}\numprint{0.8273450000000001} \\
			30 & sr & Serbian &  91 & \numprint{10607} & \nprounddigits{2}\numprint{0.818357} \\
			31 & sv & Swedish &  95 & \numprint{16458} & \nprounddigits{2}\numprint{0.802768} \\
			32 & te & Telugu &  80 & \numprint{3916} & \nprounddigits{2}\numprint{0.835664} \\
			33 & vi & Vietnamese &  88 & \numprint{10279} & \nprounddigits{2}\numprint{0.8283510000000001} \\
			34 & war & Waray &  85 & \numprint{4282} & \nprounddigits{2}\numprint{0.7477199999999999} \\
			35 & zh & Chinese &  {92} & \numprint{15595} & \nprounddigits{2}\numprint{0.839856} \\
		};
	\end{tikzpicture}

%% file: InfoLandscape.bbl
\begin{thebibliography}{102}
\providecommand{\natexlab}[1]{#1}
\expandafter\ifx\csname urlstyle\endcsname\relax
  \providecommand{\doi}[1]{doi:\discretionary{}{}{}#1}\else
  \providecommand{\doi}{doi:\discretionary{}{}{}\begingroup
  \urlstyle{rm}\Url}\fi
\providecommand{\selectlanguage}[1]{\relax}
\providecommand{\bibAnnoteFile}[1]{%
  \IfFileExists{#1}{\begin{quotation}\noindent\textsc{Key:} #1\\
  \textsc{Annotation:}\ \input{#1}\end{quotation}}{}}
\providecommand{\bibAnnote}[2]{%
  \begin{quotation}\noindent\textsc{Key:} #1\\
  \textsc{Annotation:}\ #2\end{quotation}}

\bibitem[{{\'A}lvarez et~al.(2020){\'A}lvarez, Oeberst, Cress, and
  Ferrari}]{Alvarez:et:al:2020}
{\'A}lvarez, G., Oeberst, A., Cress, U., and Ferrari, L. (2020).
\newblock Linguistic evidence of in-group bias in english and spanish
  {Wikipedia} articles about international conflicts.
\newblock \emph{Discourse, Context \& Media} 35, 100391
\bibAnnoteFile{Alvarez:et:al:2020}

\bibitem[{Bahdanau et~al.(2015)Bahdanau, Cho, and Bengio}]{Bahdanau:et:al:2014}
Bahdanau, D., Cho, K., and Bengio, Y. (2015).
\newblock Neural machine translation by jointly learning to align and
  translate.
\newblock In \emph{Proceedings of the International Conference on Learning
  Representations}. 1--15
\bibAnnoteFile{Bahdanau:et:al:2014}

\bibitem[{Bao et~al.(2012)Bao, Hecht, Carton, Quaderi, Horn, and
  Gergle}]{Bao:et:al:2012}
Bao, P., Hecht, B., Carton, S., Quaderi, M., Horn, M., and Gergle, D. (2012).
\newblock {Omnipedia}: bridging the wikipedia language gap.
\newblock In \emph{Proceedings of the SIGCHI Conference on Human Factors in
  Computing Systems}. 1075--1084
\bibAnnoteFile{Bao:et:al:2012}

\bibitem[{Barrat et~al.(2004)Barrat, Barth{\'e}lemy, Pastor-Satorras, and
  Vespignani}]{Barrat:Barthelemy:Pastor-Satorras:Vespignani:2004:a}
Barrat, A., Barth{\'e}lemy, M., Pastor-Satorras, R., and Vespignani, A. (2004).
\newblock The architecture of complex weighted networks.
\newblock \emph{Proc. Nat. Acad. Sci. USA} 101, 3747--3752
\bibAnnoteFile{Barrat:Barthelemy:Pastor-Satorras:Vespignani:2004:a}

\bibitem[{Barzilai and Zohar(2012)}]{Barzilai:Zohar:2012}
Barzilai, S. and Zohar, A. (2012).
\newblock Epistemic thinking in action: Evaluating and integrating online
  sources.
\newblock \emph{Cognition and Instruction} 30, 39--85
\bibAnnoteFile{Barzilai:Zohar:2012}

\bibitem[{Bisang and Czerwinski(2019)}]{Bisang:Czerwinski:2019}
Bisang, W. and Czerwinski, P. (2019).
\newblock Performance in knowledge assessment tests from the perspective of
  linguistic typology.
\newblock In \emph{Frontiers and Advances in Positive Learning in the Age of
  InformaTiOn (PLATO)} (Springer). 207--235
\bibAnnoteFile{Bisang:Czerwinski:2019}

\bibitem[{Braasch et~al.(2018{\natexlab{a}})Braasch, Br{\aa}ten, and
  McCrudden}]{Braasch:Braaten:McCrudden:2018}
Braasch, J.~L., Br{\aa}ten, I., and McCrudden, M.~T. (eds.)
  (2018{\natexlab{a}}).
\newblock \emph{Handbook of multiple source use} (New York \& London:
  Routledge)
\bibAnnoteFile{Braasch:Braaten:McCrudden:2018}

\bibitem[{Braasch et~al.(2018{\natexlab{b}})Braasch, McCrudden, and
  Bråten}]{Braasch:McCrudden:Braten:2018}
Braasch, J. L.~G., McCrudden, M.~T., and Bråten, I. (2018{\natexlab{b}}).
\newblock Reflections and future directions.
\newblock In \emph{Handbook of Multiple Source Use}, eds. J.~L. Braasch,
  I.~Br{\aa}ten, and M.~T. McCrudden (New York \& London: Routledge), chap.~29.
  527--537.
\newblock \doi{10.4324/9781315627496-29}
\bibAnnoteFile{Braasch:McCrudden:Braten:2018}

\bibitem[{Britt et~al.(2012)Britt, Rouet, and
  Braasch}]{Britt:Rouet:Braasch:2012}
Britt, M.~A., Rouet, J.-F., and Braasch, J.~L. (2012).
\newblock Documents as entities: Extending the situation model theory of
  comprehension.
\newblock In \emph{Reading-from words to multiple texts}, eds. M.~A. Britt,
  S.~R. Goldmann, and J.-F. Rouet (Routledge). 161--179
\bibAnnoteFile{Britt:Rouet:Braasch:2012}

\bibitem[{Britt et~al.(2018)Britt, Rouet, and Durik}]{Britt:Rouet:Durik:2018}
Britt, M.~A., Rouet, J.-F., and Durik, A.~M. (2018).
\newblock \emph{Literacy beyond text comprehension: A theory of purposeful
  reading} (New York: Routledge)
\bibAnnoteFile{Britt:Rouet:Durik:2018}

\bibitem[{Callahan and Herring(2011)}]{Callahan:Herring:2011}
Callahan, E.~S. and Herring, S.~C. (2011).
\newblock Cultural bias in {Wikipedia} content on famous persons.
\newblock \emph{Journal of the American Society for Information Science and
  Technology} 62, 1899--1915.
\newblock \doi{10.1002/asi.21577}
\bibAnnoteFile{Callahan:Herring:2011}

\bibitem[{Cheng et~al.(2016)Cheng, Dong, and Lapata}]{Cheng:Dong:Lapata:2016}
Cheng, J., Dong, L., and Lapata, M. (2016).
\newblock Long short-term memory-networks for machine reading.
\newblock In \emph{Proceedings of the 2016 Conference on Empirical Methods in
  Natural Language Processing} (Austin, Texas: Association for Computational
  Linguistics), 551--561.
\newblock \doi{10.18653/v1/D16-1053}
\bibAnnoteFile{Cheng:Dong:Lapata:2016}

\bibitem[{Cho and Afflerbach(2015)}]{Cho:Afflerbach:2015}
Cho, B.-Y. and Afflerbach, P. (2015).
\newblock Reading on the internet: Realizing and constructing potential texts.
\newblock \emph{Journal of Adolescent \& Adult Literacy} 58, 504--517
\bibAnnoteFile{Cho:Afflerbach:2015}

\bibitem[{Coiro et~al.(2018)Coiro, Sparks, and
  Kulikowich}]{Coiro:Sparks:Kulikowich:2018}
Coiro, J., Sparks, J.~R., and Kulikowich, J.~M. (2018).
\newblock Assessing online collaborative inquiry and social deliberation skills
  as learners navigate multiple sources and perspectives.
\newblock In \emph{Handbook of Multiple Source Use}, eds. J.~L. Braasch,
  I.~Br{\aa}ten, and M.~T. McCrudden (New York \& London: Routledge New York,
  NY). 485--501
\bibAnnoteFile{Coiro:Sparks:Kulikowich:2018}

\bibitem[{Conde et~al.(2020)Conde, Arruarte, Larra{\~n}aga, and
  Elorriaga}]{Conde:et:al:2020}
Conde, A., Arruarte, A., Larra{\~n}aga, M., and Elorriaga, J.~A. (2020).
\newblock How can {Wikipedia} be used to support the process of automatically
  building multilingual domain modules? {A} case study.
\newblock \emph{Information Processing \& Management} 57, 102232
\bibAnnoteFile{Conde:et:al:2020}

\bibitem[{Denning et~al.(2005)Denning, Horning, Parnas, and
  Weinstein}]{Denning:Horning:Parnas:Weinstein:2005}
Denning, P., Horning, J., Parnas, D., and Weinstein, L. (2005).
\newblock {Wikipedia} risks.
\newblock \emph{Communications of the ACM} 48, 152--152
\bibAnnoteFile{Denning:Horning:Parnas:Weinstein:2005}

\bibitem[{DeStefano and LeFevre(2007)}]{DeStefano:LeFevre:2007}
DeStefano, D. and LeFevre, J.-A. (2007).
\newblock Cognitive load in hypertext reading: A review.
\newblock \emph{Computers in human behavior} 23, 1616--1641
\bibAnnoteFile{DeStefano:LeFevre:2007}

\bibitem[{Devlin et~al.(2019)Devlin, Chang, Lee, and
  Toutanova}]{Devlin:et:al:2018}
Devlin, J., Chang, M., Lee, K., and Toutanova, K. (2019).
\newblock {BERT:} pre-training of deep bidirectional transformers for language
  understanding.
\newblock In \emph{Proceedings of the Conference of the North American Chapter
  of the Association for Computational Linguistics: Human Language
  Technologies}. 4171--4186.
\newblock \doi{10.18653/v1/n19-1423}
\bibAnnoteFile{Devlin:et:al:2018}

\bibitem[{Dimitrov et~al.(2017)Dimitrov, Singer, Lemmerich, and
  Strohmaier}]{Dimitrov:Singer:Lemmerich:Strohmaier:2017}
Dimitrov, D., Singer, P., Lemmerich, F., and Strohmaier, M. (2017).
\newblock What makes a link successful on wikipedia?
\newblock In \emph{Proceedings of the 26th International Conference on World
  Wide Web} (Republic and Canton of Geneva, CHE: International World Wide Web
  Conferences Steering Committee), WWW ’17, 917–926.
\newblock \doi{10.1145/3038912.3052613}
\bibAnnoteFile{Dimitrov:Singer:Lemmerich:Strohmaier:2017}

\bibitem[{Downs and Stea(1977)}]{Downs:Stea:1977}
Downs, R.~M. and Stea, D. (1977).
\newblock \emph{Maps in minds: Reflections on cognitive mapping} (New York:
  Harper \& Row)
\bibAnnoteFile{Downs:Stea:1977}

\bibitem[{Emmert-Streib et~al.(2016)Emmert-Streib, Dehmer, and
  Shi}]{Emmert-Streib:Dehmer:Shi:2016}
Emmert-Streib, F., Dehmer, M., and Shi, Y. (2016).
\newblock Fifty years of graph matching, network alignment and network
  comparison.
\newblock \emph{Information Sciences} 346-347, 180 -- 197.
\newblock \doi{https://doi.org/10.1016/j.ins.2016.01.074}
\bibAnnoteFile{Emmert-Streib:Dehmer:Shi:2016}

\bibitem[{Erd\H{o}s and R{\'e}nyi(1959)}]{Erdoes:Renyi:1959}
Erd\H{o}s, P. and R{\'e}nyi, A. (1959).
\newblock On random graphs.
\newblock \emph{Publicationes Mathematicae} 6, 290--297
\bibAnnoteFile{Erdoes:Renyi:1959}

\bibitem[{Feldman and Sanger(2007)}]{Feldman:Sanger:2007}
Feldman, R. and Sanger, J. (2007).
\newblock \emph{The Text Mining Handbook. Advanced Approaches in Analyzing
  Unstructured Data} (Cambridge: Cambridge University Press)
\bibAnnoteFile{Feldman:Sanger:2007}

\bibitem[{Goldman et~al.(2012)Goldman, Braasch, Wiley, Graesser, and
  Brodowinska}]{Goldman:Braasch:Wiley:Graesser:Brodowinska:2012}
Goldman, S.~R., Braasch, J.~L., Wiley, J., Graesser, A.~C., and Brodowinska, K.
  (2012).
\newblock Comprehending and learning from internet sources: Processing patterns
  of better and poorer learners.
\newblock \emph{Reading Research Quarterly} 47, 356--381
\bibAnnoteFile{Goldman:Braasch:Wiley:Graesser:Brodowinska:2012}

\bibitem[{Graham et~al.(2015)Graham, Straumann, and
  Hogan}]{Graham:Straumann:Hogan:2015}
Graham, M., Straumann, R.~K., and Hogan, B. (2015).
\newblock Digital divisions of labor and informational magnetism: Mapping
  participation in {Wikipedia}.
\newblock \emph{Annals of the Association of American Geographers} 105,
  1158--1178
\bibAnnoteFile{Graham:Straumann:Hogan:2015}

\bibitem[{Gumperz and Levinson(1991)}]{Gumperz:Levinson:1991}
Gumperz, J.~J. and Levinson, S.~C. (1991).
\newblock Rethinking linguistic relativity.
\newblock \emph{Current Anthropology} 32, 613--623
\bibAnnoteFile{Gumperz:Levinson:1991}

\bibitem[{Gumperz and Levinson(1996)}]{Gumperz:Levinson:1996}
Gumperz, J.~J. and Levinson, S.~C. (1996).
\newblock Introduction: Linguistic relativity re-examined.
\newblock In \emph{Rethinking linguistic relativity} (Cambridge University
  Press). 1--18
\bibAnnoteFile{Gumperz:Levinson:1996}

\bibitem[{Halavais and Lackaff(2008)}]{Halavais:Lackaff:2008}
Halavais, A. and Lackaff, D. (2008).
\newblock An analysis of topical coverage of {Wikipedia}.
\newblock \emph{Journal of computer-mediated communication} 13, 429--440
\bibAnnoteFile{Halavais:Lackaff:2008}

\bibitem[{{Halliday} and Hasan(1989)}]{Halliday:Hasan:1989}
{Halliday}, M. A.~K. and Hasan, R. (1989).
\newblock \emph{Language, Context, and Text: Aspects of Language in a
  Socialsemiotic Perspective} (Oxford: Oxford University Press)
\bibAnnoteFile{Halliday:Hasan:1989}

\bibitem[{Hargittai and Dobransky(2017)}]{Hargittai:Dobransky:2017}
Hargittai, E. and Dobransky, K. (2017).
\newblock Old dogs, new clicks: Digital inequality in skills and uses among
  older adults.
\newblock \emph{Canadian Journal of Communication} 42
\bibAnnoteFile{Hargittai:Dobransky:2017}

\bibitem[{Hartman et~al.(2018)Hartman, Hagerman, and
  Leu}]{Hartman:Hagerman:Leu:2018}
Hartman, D.~K., Hagerman, M.~S., and Leu, D.~J. (2018).
\newblock Toward a new literacies perspective of synthesis: Multiple source
  meaning construction.
\newblock In \emph{Handbook of Multiple Source Use}, eds. J.~L. Braasch,
  I.~Br{\aa}ten, and M.~T. McCrudden (New York \& London: Routledge), chap.~4.
  55--78
\bibAnnoteFile{Hartman:Hagerman:Leu:2018}

\bibitem[{Head(2013)}]{Head:2013}
Head, A. (2013).
\newblock Project information literacy: What can be learned about the
  information-seeking behavior of today's college students?
\newblock \emph{SSRN Electronic Journal} , 1--11
\bibAnnoteFile{Head:2013}

\bibitem[{Hecht and Gergle(2009)}]{Hecht:Gergle:2009}
Hecht, B. and Gergle, D. (2009).
\newblock Measuring self-focus bias in community-maintained knowledge
  repositories.
\newblock In \emph{Proceedings of the Fourth International Conference on
  Communities and Technologies}. 11--20.
\newblock \doi{10.1145/1556460.1556463}
\bibAnnoteFile{Hecht:Gergle:2009}

\bibitem[{Hecht and Gergle(2010{\natexlab{a}})}]{Hecht:Gergle:2010:a}
Hecht, B. and Gergle, D. (2010{\natexlab{a}}).
\newblock The {Tower of Babel} meets {Web 2.0}: User-generated content and its
  applications in a multilingual context.
\newblock In \emph{Proceedings of the SIGCHI Conference on Human Factors in
  Computing Systems}. CHI '10, 291--300.
\newblock \doi{10.1145/1753326.1753370}
\bibAnnoteFile{Hecht:Gergle:2010:a}

\bibitem[{Hecht and Gergle(2010{\natexlab{b}})}]{Hecht:Gergle:2010:b}
Hecht, B.~J. and Gergle, D. (2010{\natexlab{b}}).
\newblock On the "localness" of user-generated content.
\newblock In \emph{Proceedings of the 2010 ACM Conference on Computer Supported
  Cooperative Work} (New York, NY, USA: ACM), CSCW '10, 229--232.
\newblock \doi{10.1145/1718918.1718962}
\bibAnnoteFile{Hecht:Gergle:2010:b}

\bibitem[{Hemati et~al.(2016)Hemati, Uslu, and
  Mehler}]{Hemati:Uslu:Mehler:2016}
Hemati, W., Uslu, T., and Mehler, A. (2016).
\newblock {TextImager}: a distributed {UIMA}-based system for {NLP}.
\newblock In \emph{Proc.\ of COLING 2016: System Demonstrations}. 59--63
\bibAnnoteFile{Hemati:Uslu:Mehler:2016}

\bibitem[{Hjelmslev(1969)}]{Hjelmslev:1969}
Hjelmslev, L. (1969).
\newblock \emph{Prolegomena to a Theory of Language} (Madison: University of
  Wisconsin Press)
\bibAnnoteFile{Hjelmslev:1969}

\bibitem[{Hollan et~al.(2000)Hollan, Hutchins, and
  Kirsh}]{Hollan:Hutchins:Kirsh:2000}
Hollan, J., Hutchins, E., and Kirsh, D. (2000).
\newblock {Distributed Cognition: Toward a New Foundation for Human-Computer
  Interaction Research}.
\newblock \emph{ACM Transaction on Computer-Human Interaction} 7, 174--196
\bibAnnoteFile{Hollan:Hutchins:Kirsh:2000}

\bibitem[{Holloway et~al.(2007)Holloway, Bozicevic, and
  B\"{o}rner}]{Holloway:Bozicevic:Boerner:2007}
Holloway, T., Bozicevic, M., and B\"{o}rner, K. (2007).
\newblock Analyzing and visualizing the semantic coverage of {Wikipedia} and
  its authors: Research articles.
\newblock \emph{Complexity} 12, 30--40
\bibAnnoteFile{Holloway:Bozicevic:Boerner:2007}

\bibitem[{Hsieh(2012)}]{Hsieh:2012}
Hsieh, Y.~P. (2012).
\newblock Online social networking skills: The social affordances approach to
  digital inequality.
\newblock \emph{First Monday} 17
\bibAnnoteFile{Hsieh:2012}

\bibitem[{Jiang et~al.(2017)Jiang, Bai, Zhang, and Hu}]{Jiang:et:al:2017}
Jiang, Y., Bai, W., Zhang, X., and Hu, J. (2017).
\newblock {Wikipedia}-based information content and semantic similarity
  computation.
\newblock \emph{Information Processing \& Management} 53, 248--265
\bibAnnoteFile{Jiang:et:al:2017}

\bibitem[{Joulin et~al.(2017)Joulin, Grave, Bojanowski, and
  Mikolov}]{Joulin:Grave:Bojanowski:Mikolov:2016}
Joulin, A., Grave, E., Bojanowski, P., and Mikolov, T. (2017).
\newblock Bag of tricks for efficient text classification.
\newblock In \emph{Proceedings of the 15th Conference of the EACL: Volume 2,
  Short Papers} (Valencia, Spain: Association for Computational Linguistics),
  427--431
\bibAnnoteFile{Joulin:Grave:Bojanowski:Mikolov:2016}

\bibitem[{Karimi et~al.(2015)Karimi, Bohlin, Samoilenko, Rosvall, and
  Lancichinetti}]{Karimi:et:al:2015}
Karimi, F., Bohlin, L., Samoilenko, A., Rosvall, M., and Lancichinetti, A.
  (2015).
\newblock Mapping bilateral information interests using the activity of
  {Wikipedia} editors.
\newblock \emph{Palgrave Communications} 1, 1--7
\bibAnnoteFile{Karimi:et:al:2015}

\bibitem[{Kendeou et~al.(2019)Kendeou, Robinson, and
  McCrudden}]{Kendeou:Robinson:McCrudden:2019}
Kendeou, P., Robinson, D.~H., and McCrudden, M.~T. (2019).
\newblock Modeling the dissemination of misinformation through discourse
  dynamics.
\newblock In \emph{Misinformation and Disinformation in Education: an
  Introduction} (IAP). 1--4
\bibAnnoteFile{Kendeou:Robinson:McCrudden:2019}

\bibitem[{Kintsch(1998)}]{Kintsch:1998}
Kintsch, W. (1998).
\newblock \emph{Comprehension. A Paradigm for Cognition} (Cambridge: Cambridge
  University Press)
\bibAnnoteFile{Kintsch:1998}

\bibitem[{Kittur et~al.(2009)Kittur, Chi, and Suh}]{Kittur:Chi:Suh:2009}
Kittur, A., Chi, E.~H., and Suh, B. (2009).
\newblock What's in {Wikipedia}?: mapping topics and conflict using socially
  annotated category structure.
\newblock In \emph{Proceedings of the 27th international conference on Human
  factors in computing systems} (New York, NY, USA: ACM), CHI '09, 1509--1512
\bibAnnoteFile{Kittur:Chi:Suh:2009}

\bibitem[{K{\"o}hler et~al.(2005)K{\"o}hler, Altmann, and
  Piotrowski}]{Koehler:Altmann:Piotrowski:2005}
K{\"o}hler, R., Altmann, G., and Piotrowski, R.~G. (eds.) (2005).
\newblock \emph{Quantitative Linguistics. {A}n International Handbook}
  (Berlin/New York: Mouton de Gruyter)
\bibAnnoteFile{Koehler:Altmann:Piotrowski:2005}

\bibitem[{Konca et~al.(2020)Konca, Mehler, Baumartz, and
  Hemati}]{Konca:Mehler:Baumartz:Hemati:2020}
Konca, M., Mehler, A., Baumartz, D., and Hemati, W. (2020).
\newblock From distinguishability to informativity: A quantitative text model
  for detecting random texts.
\newblock \emph{Manuscript accepted for publication}
\bibAnnoteFile{Konca:Mehler:Baumartz:Hemati:2020}

\bibitem[{Konieczny(2016)}]{Konieczny:2016}
Konieczny, P. (2016).
\newblock Teaching with {Wikipedia} in a 21st-century classroom: Perceptions of
  {Wikipedia} and its educational benefits.
\newblock \emph{Journal of the Association for Information Science and
  Technology} 67, 1523--1534
\bibAnnoteFile{Konieczny:2016}

\bibitem[{Koutra et~al.(2016)Koutra, Shah, Vogelstein, Gallagher, and
  Faloutsos}]{Koutra:et:al:2016}
Koutra, D., Shah, N., Vogelstein, J.~T., Gallagher, B., and Faloutsos, C.
  (2016).
\newblock {DeltaCon}: Principled massive-graph similarity function with
  attribution.
\newblock \emph{{ACM} Transactions on Knowledge Discovery from Data} 10,
  28:1--28:43.
\newblock \doi{10.1145/2824443}
\bibAnnoteFile{Koutra:et:al:2016}

\bibitem[{Lakoff(1987)}]{Lakoff:1987}
Lakoff, G. (1987).
\newblock \emph{Women, Fire, and Dangerous Things: What Categories Reveal about
  the Mind} (Chicago: University of Chicago Press)
\bibAnnoteFile{Lakoff:1987}

\bibitem[{Lamprecht et~al.(2017)Lamprecht, Lerman, Helic, and
  Strohmaier}]{Lamprecht:Lerman:Helic:Strohmaier:2017}
Lamprecht, D., Lerman, K., Helic, D., and Strohmaier, M. (2017).
\newblock How the structure of wikipedia articles influences user navigation.
\newblock \emph{New Review of Hypermedia and Multimedia} 23, 29--50
\bibAnnoteFile{Lamprecht:Lerman:Helic:Strohmaier:2017}

\bibitem[{Laufer et~al.(2015)Laufer, Wagner, Fl{\"o}ck, and
  Strohmaier}]{Laufer:Wagner:Floeck:Strohmaier:2015}
Laufer, P., Wagner, C., Fl{\"o}ck, F., and Strohmaier, M. (2015).
\newblock Mining cross-cultural relations from {Wikipedia}: a study of 31
  {European} food cultures.
\newblock In \emph{Proceedings of the ACM Web Science Conference}. 1--10
\bibAnnoteFile{Laufer:Wagner:Floeck:Strohmaier:2015}

\bibitem[{Lemmerich et~al.(2019)Lemmerich, S{\'a}ez-Trumper, West, and
  Zia}]{Lemmerich:et:al:2019}
Lemmerich, F., S{\'a}ez-Trumper, D., West, R., and Zia, L. (2019).
\newblock Why the world reads {Wikipedia}: Beyond english speakers.
\newblock In \emph{Proceedings of the Twelfth ACM International Conference on
  Web Search and Data Mining}. 618--626
\bibAnnoteFile{Lemmerich:et:al:2019}

\bibitem[{List and Alexander(2019)}]{List:Alexander:2019}
List, A. and Alexander, P.~A. (2019).
\newblock Toward an integrated framework of multiple text use.
\newblock \emph{Educational Psychologist} 54, 20--39
\bibAnnoteFile{List:Alexander:2019}

\bibitem[{Loh and Kanai(2016)}]{Loh:Kanai:2016}
Loh, K.~K. and Kanai, R. (2016).
\newblock How has the internet reshaped human cognition?
\newblock \emph{The Neuroscientist} 22, 506--520
\bibAnnoteFile{Loh:Kanai:2016}

\bibitem[{Lorini et~al.(2020)Lorini, Rando, Saez-Trumper, and
  Castillo}]{Lorini:et:al:2020}
Lorini, V., Rando, J., Saez-Trumper, D., and Castillo, C. (2020).
\newblock Uneven coverage of natural disasters in {Wikipedia}: the case of
  flood.
\newblock \emph{arXiv preprint arXiv:2001.08810}
\bibAnnoteFile{Lorini:et:al:2020}

\bibitem[{Lucassen and Schraagen(2010)}]{Lucassen:Schraagen:2010}
Lucassen, T. and Schraagen, J.~M. (2010).
\newblock Trust in {Wikipedia}: How users trust information from an unknown
  source.
\newblock In \emph{Proceedings of the 4th Workshop on Information Credibility}.
  WICOW '10, 19--26.
\newblock \doi{10.1145/1772938.1772944}
\bibAnnoteFile{Lucassen:Schraagen:2010}

\bibitem[{Lucy(1997)}]{Lucy:1997}
Lucy, J.~A. (1997).
\newblock Linguistic relativity.
\newblock \emph{Annual Review of Anthropology} 26, 291--312
\bibAnnoteFile{Lucy:1997}

\bibitem[{Massa and Scrinzi(2012)}]{Massa:Scrinzi:2012}
Massa, P. and Scrinzi, F. (2012).
\newblock {Manypedia}: Comparing language points of view of wikipedia
  communities.
\newblock In \emph{Proceedings of the Eighth Annual International Symposium on
  Wikis and Open Collaboration} (New York, NY, USA: Association for Computing
  Machinery), WikiSym ’12, 1--9.
\newblock \doi{10.1145/2462932.2462960}
\bibAnnoteFile{Massa:Scrinzi:2012}

\bibitem[{McMahon et~al.(2017)McMahon, Johnson, and
  Hecht}]{McMahon:Johnson:Hecht:2017}
McMahon, C., Johnson, I., and Hecht, B. (2017).
\newblock The substantial interdependence of {Wikipedia} and {Google}: A case
  study on the relationship between peer production communities and information
  technologies.
\newblock In \emph{Eleventh International AAAI Conference on Web and Social
  Media}. 142--151
\bibAnnoteFile{McMahon:Johnson:Hecht:2017}

\bibitem[{Mehler et~al.(2019)Mehler, Gleim, Gaitsch, Uslu, and
  Hemati}]{Mehler:Gleim:Gaitsch:Uslu:Hemati:2019}
Mehler, A., Gleim, R., Gaitsch, R., Uslu, T., and Hemati, W. (2019).
\newblock From topic networks to distributed cognitive maps: {Zipfian} topic
  universes in the area of volunteered geographic information.
\newblock \emph{Complexity} Accepted
\bibAnnoteFile{Mehler:Gleim:Gaitsch:Uslu:Hemati:2019}

\bibitem[{Mehler et~al.(2018)Mehler, Hemati, Uslu, and
  L\"{u}cking}]{Mehler:Hemati:Uslu:Luecking:2018}
Mehler, A., Hemati, W., Uslu, T., and L\"{u}cking, A. (2018).
\newblock A multidimensional model of syntactic dependency trees for authorship
  attribution.
\newblock In \emph{Quantitative analysis of dependency structures}, eds.
  J.~Jiang and H.~Liu (Berlin/New York: De Gruyter). 315--348
\bibAnnoteFile{Mehler:Hemati:Uslu:Luecking:2018}

\bibitem[{Mehler et~al.(2011)Mehler, Pustylnikov, and
  Diewald}]{Mehler:Pustylnikov:Diewald:2010:a}
Mehler, A., Pustylnikov, O., and Diewald, N. (2011).
\newblock Geography of social ontologies: Testing a variant of the {Sapir-Whorf
  Hypothesis} in the context of {Wikipedia}.
\newblock \emph{Computer Speech and Language} 25, 716--740.
\newblock \doi{http://dx.doi.org/10.1016/j.csl.2010.05.006}
\bibAnnoteFile{Mehler:Pustylnikov:Diewald:2010:a}

\bibitem[{Mehler and Ramesh(2019)}]{Mehler:Ramesh:2019}
Mehler, A. and Ramesh, V. (2019).
\newblock {TextInContext}: On the way to a framework for measuring the
  context-sensitive complexity of educationally relevant texts---a combined
  cognitive and computational linguistic approach.
\newblock In \emph{Frontiers and Advances in Positive Learning in the Age of
  InformaTiOn (PLATO)}, ed. O.~Zlatkin-Troitschanskaia (Cham: Springer
  International Publishing). 167--195.
\newblock \doi{10.1007/978-3-030-26578-6_14}
\bibAnnoteFile{Mehler:Ramesh:2019}

\bibitem[{Mehler et~al.(2010)Mehler, Sharoff, and
  Santini}]{Mehler:Sharoff:Santini:2009:a}
Mehler, A., Sharoff, S., and Santini, M. (eds.) (2010).
\newblock \emph{Genres on the Web: Computational Models and Empirical Studies}
  (Dordrecht: Springer)
\bibAnnoteFile{Mehler:Sharoff:Santini:2009:a}

\bibitem[{Mesgari et~al.(2015)Mesgari, Okoli, Mehdi, Nielsen, and
  Lanam{\"a}ki}]{Mesgari:Okoli:Mehdi:Nielsen:Lanamaki:2015}
Mesgari, M., Okoli, C., Mehdi, M., Nielsen, F.~{\AA}., and Lanam{\"a}ki, A.
  (2015).
\newblock \enquote{The sum of all human knowledge}: A systematic review of
  scholarly research on the content of {Wikipedia}.
\newblock \emph{Journal of the Association for Information Science and
  Technology} 66, 219--245
\bibAnnoteFile{Mesgari:Okoli:Mehdi:Nielsen:Lanamaki:2015}

\bibitem[{Miquel-Rib{\'e} and Laniado(2016)}]{Miquel-Ribe:Laniado:2016}
Miquel-Rib{\'e}, M. and Laniado, D. (2016).
\newblock Cultural identities in wikipedias.
\newblock In \emph{Proceedings of the 7th 2016 International Conference on
  Social Media \& Society}. 1--10
\bibAnnoteFile{Miquel-Ribe:Laniado:2016}

\bibitem[{Miz et~al.(2020)Miz, Hanna, Aspert, Ricaud, and
  Vandergheynst}]{Miz:Hanna:Aspert:Ricaud:Vandergheynst:2020}
Miz, V., Hanna, J., Aspert, N., Ricaud, B., and Vandergheynst, P. (2020).
\newblock What is trending on {Wikipedia}? capturing trends and language biases
  across {Wikipedia} editions.
\newblock In \emph{Companion Proceedings of the Web Conference 2020}. 794--801
\bibAnnoteFile{Miz:Hanna:Aspert:Ricaud:Vandergheynst:2020}

\bibitem[{Molerov et~al.(2020)Molerov, Zlatkin-Troitschanskaia, Nagel, Schmidt,
  Solano-Flores, and G.}]{Molerov:et:al:2020}
Molerov, D., Zlatkin-Troitschanskaia, O., Nagel, M.-T., Schmidt, S.,
  Solano-Flores, and G., S.~R. (2020).
\newblock `assessing university students' ability to judge online information
  and the impact of individual characteristics.
\newblock \emph{Frontiers in Education. Submitted to this issue special issue:
  Assessing Information Processing and Online Reasoning as a Prerequisite for
  Learning in Higher Education}
\bibAnnoteFile{Molerov:et:al:2020}

\bibitem[{Oeberst et~al.(2018)Oeberst, von~der Beck, Back, Cress, and
  Nestler}]{Oeberst:et:al:2018}
Oeberst, A., von~der Beck, I., Back, M.~D., Cress, U., and Nestler, S. (2018).
\newblock Biases in the production and reception of collective knowledge: the
  case of hindsight bias in {Wikipedia}.
\newblock \emph{Psychological research} 82, 1010--1026
\bibAnnoteFile{Oeberst:et:al:2018}

\bibitem[{Oeberst et~al.(2019)Oeberst, von~der Beck, Matschke, Ihme, and
  Cress}]{Oeberst:vonderBeck:Matschke:Ihme:Cress:2019}
Oeberst, A., von~der Beck, I., Matschke, C., Ihme, T.~A., and Cress, U. (2019).
\newblock Collectively biased representations of the past: Ingroup bias in
  {Wikipedia} articles about intergroup conflicts.
\newblock \emph{British Journal of Social Psychology}
\bibAnnoteFile{Oeberst:vonderBeck:Matschke:Ihme:Cress:2019}

\bibitem[{OECD(2007)}]{OECD:2007}
[Dataset] OECD (2007).
\newblock {Revised Field of Science and Technology} ({FOS}).
\newblock \url{www.oecd.org/science/inno/38235147.pdf}
\bibAnnoteFile{OECD:2007}

\bibitem[{Okoli et~al.(2012)Okoli, Mehdi, Mesgari, Nielsen, and
  Lanam{\"a}ki}]{Okoli:et:al:2012}
Okoli, C., Mehdi, M., Mesgari, M., Nielsen, F.~{\AA}., and Lanam{\"a}ki, A.
  (2012).
\newblock The people's encyclopedia under the gaze of the sages: {A} systematic
  review of scholarly research on {Wikipedia}.
\newblock \emph{SSRN Electronic Journal} , 1--138\doi{10.2139/ssrn.2021326}
\bibAnnoteFile{Okoli:et:al:2012}

\bibitem[{Okoli et~al.(2014)Okoli, Mehdi, Mesgari, Nielsen, and
  Lanam{\"a}ki}]{Okoli:Mehdi:Mesgari:Nielsen:Lanamaki:2014}
Okoli, C., Mehdi, M., Mesgari, M., Nielsen, F.~{\AA}., and Lanam{\"a}ki, A.
  (2014).
\newblock {Wikipedia} in the eyes of its beholders: A systematic review of
  scholarly research on wikipedia readers and readership.
\newblock \emph{Journal of the Association for Information Science and
  Technology} 65, 2381--2403
\bibAnnoteFile{Okoli:Mehdi:Mesgari:Nielsen:Lanamaki:2014}

\bibitem[{Pentzold et~al.(2017)Pentzold, Weltevrede, Mauri, Laniado,
  Kaltenbrunner, and Borra}]{Pentzold:et:al:2017}
Pentzold, C., Weltevrede, E., Mauri, M., Laniado, D., Kaltenbrunner, A., and
  Borra, E. (2017).
\newblock Digging wikipedia: the online encyclopedia as a digital cultural
  heritage gateway and site.
\newblock \emph{Journal on Computing and Cultural Heritage (JOCCH)} 10, 1--19
\bibAnnoteFile{Pentzold:et:al:2017}

\bibitem[{Perfetti et~al.(1999)Perfetti, Rouet, and
  Britt}]{Perfetti:Rouet:Britt:1999}
Perfetti, C.~A., Rouet, J.-F., and Britt, M.~A. (1999).
\newblock Toward a theory of documents representation.
\newblock In \emph{The construction of mental representations during reading},
  eds. H.~van Oostendorp and S.~R. Goldman (Mahwah, NJ: Erlbaum). 99--122
\bibAnnoteFile{Perfetti:Rouet:Britt:1999}

\bibitem[{Pickering and Garrod(2004)}]{Pickering:Garrod:2004}
Pickering, M.~J. and Garrod, S. (2004).
\newblock Toward a mechanistic psychology of dialogue.
\newblock \emph{Behavioral and Brain Sciences} 27, 169--226
\bibAnnoteFile{Pickering:Garrod:2004}

\bibitem[{Power et~al.(2003)Power, Scott, and
  Bouayad-Agha}]{Power:Scott:BouayadAgha:2003}
Power, R., Scott, D., and Bouayad-Agha, N. (2003).
\newblock Document structure.
\newblock \emph{Computational Linguistics} 29, 211--260
\bibAnnoteFile{Power:Scott:BouayadAgha:2003}

\bibitem[{Primor and Katzir(2018)}]{Primor:Katzir:2018}
Primor, L. and Katzir, T. (2018).
\newblock Measuring multiple text integration: a review.
\newblock \emph{Frontiers in Psychology} 9, 2294
\bibAnnoteFile{Primor:Katzir:2018}

\bibitem[{Salmer{\'o}n et~al.(2018)Salmer{\'o}n, Kammerer, and
  Delgado}]{Salmeron:Kammerer:Delgado:2018}
Salmer{\'o}n, L., Kammerer, Y., and Delgado, P. (2018).
\newblock Non-academic multiple source use on the internet.
\newblock In \emph{Handbook of Multiple Source Use}, eds. J.~L. Braasch,
  I.~Br{\aa}ten, and M.~T. McCrudden (New York \& London: Routledge). 285--302
\bibAnnoteFile{Salmeron:Kammerer:Delgado:2018}

\bibitem[{Samoilenko et~al.(2016)Samoilenko, Karimi, Edler, Kunegis, and
  Strohmaier}]{Samoilenko:et:al:2016}
Samoilenko, A., Karimi, F., Edler, D., Kunegis, J., and Strohmaier, M. (2016).
\newblock Linguistic neighbourhoods: explaining cultural borders on {Wikipedia}
  through multilingual co-editing activity.
\newblock \emph{EPJ data science} 5, 9
\bibAnnoteFile{Samoilenko:et:al:2016}

\bibitem[{Samoilenko et~al.(2017)Samoilenko, Lemmerich, Weller, Zens, and
  Strohmaier}]{Samoilenko:et:al:2017}
Samoilenko, A., Lemmerich, F., Weller, K., Zens, M., and Strohmaier, M. (2017).
\newblock Analysing timelines of national histories across {Wikipedia}
  editions: A comparative computational approach.
\newblock In \emph{Eleventh International AAAI Conference on Web and Social
  Media}. 210--219
\bibAnnoteFile{Samoilenko:et:al:2017}

\bibitem[{Scaffidi et~al.(2017)Scaffidi, Khan, Wang, Keren, Tsui, Garg
  et~al.}]{Scaffidi:et:al:2017}
Scaffidi, M.~A., Khan, R., Wang, C., Keren, D., Tsui, C., Garg, A., et~al.
  (2017).
\newblock Comparison of the impact of wikipedia, {UpToDate}, and a digital
  textbook on short-term knowledge acquisition among medical students.
\newblock \emph{JMIR medical education} 3, e20
\bibAnnoteFile{Scaffidi:et:al:2017}

\bibitem[{Sharifian(2017)}]{Sharifian:2017}
Sharifian, F. (2017).
\newblock Cultural linguistics and linguistic relativity.
\newblock \emph{Language Sciences} 59, 83--92
\bibAnnoteFile{Sharifian:2017}

\bibitem[{Siew(2020)}]{Siew:2020}
Siew, C. S.~Q. (2020).
\newblock Applications of network science to education research: Quantifying
  knowledge and the development of expertise through network analysis.
\newblock \emph{Education Sciences} 10, 101
\bibAnnoteFile{Siew:2020}

\bibitem[{Singer et~al.(2017)Singer, Lemmerich, West, Zia, Wulczyn, Strohmaier
  et~al.}]{Singer:et:al:2017}
Singer, P., Lemmerich, F., West, R., Zia, L., Wulczyn, E., Strohmaier, M.,
  et~al. (2017).
\newblock Why we read {Wikipedia}.
\newblock In \emph{Proceedings of the 26th International Conference on World
  Wide Web}. 1591--1600
\bibAnnoteFile{Singer:et:al:2017}

\bibitem[{Smith(2020)}]{Smith:2020}
Smith, D.~A. (2020).
\newblock Situating {Wikipedia} as a health information resource in various
  contexts: A scoping review.
\newblock \emph{PloS one} 15, e0228786
\bibAnnoteFile{Smith:2020}

\bibitem[{Stella et~al.(2020)Stella, Restocchi, and
  Deyne}]{Stella:Restocchi:DeDeyne:2020}
Stella, M., Restocchi, V., and Deyne, S.~D. (2020).
\newblock {\#}lockdown: Network-enhanced emotional profiling in the time of
  {COVID-19}.
\newblock \emph{Big Data Cogn. Comput.} 4, 14.
\newblock \doi{10.3390/bdcc4020014}
\bibAnnoteFile{Stella:Restocchi:DeDeyne:2020}

\bibitem[{Storrer(2002)}]{Storrer:2002}
Storrer, A. (2002).
\newblock Coherence in text and hypertext.
\newblock \emph{Document Design} 3, 156--168
\bibAnnoteFile{Storrer:2002}

\bibitem[{Sweller(1994)}]{Sweller:1994}
Sweller, J. (1994).
\newblock Cognitive load theory, learning difficulty, and instructional design.
\newblock \emph{Learning and Instruction} 4, 295 -- 312.
\newblock \doi{https://doi.org/10.1016/0959-4752(94)90003-5}
\bibAnnoteFile{Sweller:1994}

\bibitem[{Tobler(1970)}]{Tobler:1970}
Tobler, W.~R. (1970).
\newblock A computer movie simulating urban growth in the {Detroit} region.
\newblock \emph{Economic Geography} 46, 234--240
\bibAnnoteFile{Tobler:1970}

\bibitem[{Uslu et~al.(2019)Uslu, Mehler, and
  Baumartz}]{Uslu:Mehler:Baumartz:2019}
Uslu, T., Mehler, A., and Baumartz, D. (2019).
\newblock {Computing Classifier-based Embeddings with the Help of text2ddc}.
\newblock In \emph{{Proceedings of the 20th International Conference on
  Computational Linguistics and Intelligent Text Processing, (CICLing 2019)}}.
  {CICLing 2019}
\bibAnnoteFile{Uslu:Mehler:Baumartz:2019}

\bibitem[{van Dijk(1980)}]{Dijk:1980:a}
van Dijk, T.~A. (1980).
\newblock \emph{Macrostructures. {A}n interdisciplinary Study of Global
  Structures in Discourse, Interaction, and Cognition} (Hillsdale: Erlbaum)
\bibAnnoteFile{Dijk:1980:a}

\bibitem[{van Dijk and Kintsch(1983)}]{Dijk:Kintsch:1983}
van Dijk, T.~A. and Kintsch, W. (1983).
\newblock \emph{Strategies of Discourse Comprehension} (New York: Academic
  Press)
\bibAnnoteFile{Dijk:Kintsch:1983}

\bibitem[{Vaswani et~al.(2017)Vaswani, Shazeer, Parmar, Uszkoreit, Jones, Gomez
  et~al.}]{Vaswani:et:al:2017}
Vaswani, A., Shazeer, N., Parmar, N., Uszkoreit, J., Jones, L., Gomez, A.~N.,
  et~al. (2017).
\newblock Attention is all you need.
\newblock In \emph{Advances in neural information processing systems}.
  5998--6008
\bibAnnoteFile{Vaswani:et:al:2017}

\bibitem[{Wagner et~al.(2016)Wagner, Graells-Garrido, Garcia, and
  Menczer}]{Wagner:et:al:2016}
Wagner, C., Graells-Garrido, E., Garcia, D., and Menczer, F. (2016).
\newblock Women through the glass ceiling: gender asymmetries in {Wikipedia}.
\newblock \emph{EPJ Data Science} 5, 5
\bibAnnoteFile{Wagner:et:al:2016}

\bibitem[{Warncke-Wang et~al.(2012)Warncke-Wang, Uduwage, Dong, and
  Riedl}]{Warncke-Wang:et:al:2012}
Warncke-Wang, M., Uduwage, A., Dong, Z., and Riedl, J. (2012).
\newblock In search of the ur-wikipedia: universality, similarity, and
  translation in the {Wikipedia} inter-language link network.
\newblock In \emph{Proceedings of the Eighth Annual International Symposium on
  Wikis and Open Collaboration}. 1--10
\bibAnnoteFile{Warncke-Wang:et:al:2012}

\bibitem[{Wolf(2018)}]{Wolf:2018}
Wolf, M. (2018).
\newblock \emph{Reader, come home: The reading brain in a digital world}
  (Harper New York, NY)
\bibAnnoteFile{Wolf:2018}

\bibitem[{Zlatkin-Troitschanskaia
  et~al.(2019{\natexlab{a}})Zlatkin-Troitschanskaia, Bisang, Mehler, Banerjee,
  and Roeper}]{Zlatkin-Troitschanskaia:et:al:2019}
Zlatkin-Troitschanskaia, O., Bisang, W., Mehler, A., Banerjee, M., and Roeper,
  J. (2019{\natexlab{a}}).
\newblock Positive learning in the internet age: Developments and perspectives
  in the plato program.
\newblock In \emph{Frontiers and Advances in {Positive Learning in the Age of
  InformaTiOn} (PLATO)}, ed. O.~Zlatkin-Troitschanskaia (Cham: Springer
  International Publishing). 1--5.
\newblock \doi{10.1007/978-3-030-26578-6_1}
\bibAnnoteFile{Zlatkin-Troitschanskaia:et:al:2019}

\bibitem[{Zlatkin-Troitschanskaia
  et~al.(2019{\natexlab{b}})Zlatkin-Troitschanskaia, Br{\"u}ckner, Molerov, and
  Bisang}]{Zlatkin-Troitschanskaia:Bruckner:Molerov:Bisang:2019}
Zlatkin-Troitschanskaia, O., Br{\"u}ckner, S., Molerov, D., and Bisang, W.
  (2019{\natexlab{b}}).
\newblock What can we learn from theoretical considerations and empirical
  evidence on learning in higher education? implications for an
  interdisciplinary research framework.
\newblock In \emph{Frontiers and Advances in Positive Learning in the Age of
  InformaTiOn (PLATO)} (Springer). 287--309
\bibAnnoteFile{Zlatkin-Troitschanskaia:Bruckner:Molerov:Bisang:2019}

\bibitem[{Zlatkin-Troitschanskaia et~al.(2018)Zlatkin-Troitschanskaia, Schmidt,
  Molerov, Shavelson, and
  Berliner}]{Zlatkin-Troitschanskaia:Schmidt:Molerov:Shavelson:Berliner:2017}
Zlatkin-Troitschanskaia, O., Schmidt, S., Molerov, D., Shavelson, R.~J., and
  Berliner, D. (2018).
\newblock Conceptual fundamentals for a theoretical and empirical framework of
  positive learning.
\newblock In \emph{Positive Learning in the Age of Information ({PLATO}) -- A
  blessing or a curse?}, eds. O.~Zlatkin-Troitschanskaia, G.~Wittum, and
  A.~Dengel (Springer). 29--50
\bibAnnoteFile{Zlatkin-Troitschanskaia:Schmidt:Molerov:Shavelson:Berliner:2017}

\end{thebibliography}
